\documentclass[twoside,11pt]{article}

\usepackage{bbm}
\usepackage{algorithmicx}
\usepackage{jmlr2e}
\usepackage{amsmath}
\usepackage{xcolor}
\usepackage{bm}
\usepackage{tikz}
\usetikzlibrary{bayesnet}
\usepackage{subcaption}
\usepackage{float}
\usepackage{stackengine}
\usepackage{multirow}
\usepackage{stmaryrd}
\usepackage{pgfplots}
\usepgfplotslibrary{fillbetween}
\usetikzlibrary{patterns}
\usetikzlibrary{shapes.geometric}
\usepackage[ruled,vlined]{algorithm2e}
\usepackage{xstring}
\usepackage[margin=2cm,bottom=1in]{geometry}
\usepackage{xfp}
\usepackage{afterpage}

\definecolor{BlueCode}{RGB}{0,0,250}
\definecolor{GreenCode}{RGB}{0,125,0}

\let\oldmathcal\mathcal
\DeclareMathAlphabet{\mathdutchcal}{U}{dutchcal}{m}{n}

\renewcommand{\mathcal}[1]{%
  \IfSubStr{ABCDEFGHIJKLMNOPQRSTUVWXYZ}{#1}{\oldmathcal{#1}}{\mathdutchcal{#1}}
}
\makeatother

\pgfmathdeclarefunction{gauss}{3}{%
  \pgfmathparse{1/(#3*sqrt(2*pi))*exp(-((#1-#2)^2)/(2*#3^2))}%
}
\pgfmathdeclarefunction{sum_gauss}{5}{%
  \pgfmathparse{1/(3*#3*sqrt(2*pi))*exp(-((#1-#2)^2)/(2*#3^2)) + 2/(3*#5*sqrt(2*pi))*exp(-((#1-#4)^2)/(2*#5^2))}%
}

\newcommand{\kl}[2]{D_{\mathrm{KL}} \left[ \left. \left. #1 \right|\right| #2 \right] }
\DeclareMathOperator*{\argmax}{arg\,max}
\DeclareMathOperator*{\argmin}{arg\,min}
\definecolor{blue-violet}{rgb}{0.54, 0.17, 0.89}
\definecolor{bluepigment}{rgb}{0.2, 0.2, 0.6}
\makeatletter
\newcommand*\bigcdot{\mathpalette\bigcdot@{.5}}
\newcommand*\bigcdot@[2]{\mathbin{\vcenter{\hbox{\scalebox{#2}{$\m@th#1\bullet$}}}}}
\makeatother
\def\delequal{\mathrel{\ensurestackMath{\stackon[1pt]{=}{\scriptstyle\Delta}}}}
\definecolor{Yellow}{RGB}{211, 176, 15}
\definecolor{Green}{rgb}{0.01, 0.75, 0.24}
\colorlet{Red}{red!50!black}
\colorlet{Blue}{blue!50!black}
\definecolor{Violet}{rgb}{0.56,0.14,0.56}
\newcommand{\indep}{\perp \! \! \! \perp}
\newcommand{\MultiBernoulli}{\mathcal{B}\hspace{-0.02cm}\text{ernoulli}}

\makeatletter
\newcounter{subsubsubsection}[subsubsection]
\def\subsubsubsectionmark#1{}

\def\subsubsubsection{\@startsection
     {subsubsubsection}{4}{\z@} {-3.25ex plus -1
     ex minus -.2ex}{1.5ex plus .2ex}{\normalsize\bf}}
\def\l@subsubsubsection{\@dottedtocline{4}{4.8em}
     {4.2em}}
\makeatother

\tikzset{
    pics/vae/.style args={#1/#2/#3/#4/#5}{
    code = {
		\draw[fill=blue!20!white] (-1,-12) rectangle (1,-8);
	    \node at (0, -10) {#1};

		\coordinate (A) at (1.5,-12);
		\coordinate (B) at (4,-11);
		\coordinate (C) at (4,-9);
		\coordinate (D) at (1.5,-8);
		\draw[fill=green!20!white] (A) -- coordinate[pos=.6] (AB) (B)--(C)
         --coordinate[pos=.45] (CD) (D)
         --coordinate[pos=.55] (DA) cycle;
	    \node at (2.75, -10) {Encoder};

		\draw[fill=red!20!white] (4.5,-11) rectangle (5.5,-10.1);
	    \node at (5, -10.5) {#3};
		\draw[fill=red!20!white] (4.5,-9.9) rectangle (5.5,-9);
	    \node at (5, -9.5) {#2};

		\node[latent] at (5, -8.4) (epsilon) {$\epsilon$};

		\coordinate (A) at (7.5,-11);
		\coordinate (B) at (10,-12);
		\coordinate (C) at (10,-8);
		\coordinate (D) at (7.5,-9);
		\draw[fill=green!20!white] (A) -- coordinate[pos=.6] (AB) (B)--(C)
         --coordinate[pos=.45] (CD) (D)
         --coordinate[pos=.55] (DA) cycle;
	    \node at (8.75, -10) {Decoder};

		\draw[fill=gray!20!white] (6,-11) rectangle (7,-9);
	    \node at (6.5,-10) {#4};

		\draw[fill=blue!20!white] (10.5,-12) rectangle (12.5,-8);
	    \node at (11.5, -10) {#5};

		\draw[-latex] (4,-10.5) -- (4.5,-10.5);
		\draw[-latex] (4,-9.5) -- (4.5,-9.5);
		\draw[-latex] (1,-10) -- (1.5,-10);
		\draw[-latex] (5.36,-8.4) -- (5.75,-8.4) -- (5.75,-10) -- (6,-10);
		\draw[-latex] (5.5,-10.5) -- (5.75,-10.5) -- (5.75,-10) -- (6,-10);
		\draw[-latex] (5.5,-9.5) -- (5.75,-9.5) -- (5.75,-10) -- (6,-10);

		\draw[-latex] (7,-10) -- (7.5,-10);
		\draw[-latex] (10,-10) -- (10.5,-10);
    }}
}
    
\tikzset{
    pics/hmm/.code = {
    		\pic[rotate=90]{vae=$o_t$/$\mu$/$\ln \sigma$/$\hat{s}_t$/$\hat{o}_t$};

		\draw[fill=blue!20!white] (12,3.5) rectangle (13,5.5);
	    \node at (12.5, 4.5) {$a_t$};

		\coordinate (A) at (13.5,3.5);
		\coordinate (B) at (16,4.5);
		\coordinate (C) at (16,6.5);
		\coordinate (D) at (13.5,7.5);
		\draw[fill=orange!20!white] (A) -- coordinate[pos=.6] (AB) (B)--(C)
         --coordinate[pos=.45] (CD) (D)
         --coordinate[pos=.55] (DA) cycle;
	    \node at (14.75, 5.5) {Transition};

		\draw[fill=red!20!white] (16.5,4.5) rectangle (17.5,5.4);
	    \node at (17, 5) {$\ln \mathring{\sigma}$};
		\draw[fill=red!20!white] (16.5,5.6) rectangle (17.5,6.5);
	    \node at (17, 6) {$\mathring{\mu}$};
		\node[latent] at (17, 7.1) (epsilon) {$\epsilon$};

		\draw[fill=gray!20!white] (18,4.5) rectangle (19,6.5);
	    \node at (18.5, 5.5) {$\mathring{s}_{t+1}$};

		\draw[-latex] (13,4.5) -- (13.5,4.5);
		\draw[-latex] (11,6.5) -- (13.5,6.5);

		\draw[-latex] (16,5) -- (16.5,5);
		\draw[-latex] (16,6) -- (16.5,6);

		\draw[-latex] (17.5,6) -- (17.75,6) -- (17.75,5.5) -- (18,5.5);
		\draw[-latex] (17.5,5) -- (17.75,5) -- (17.75,5.5) -- (18,5.5);
		\draw[-latex] (17.36,7.1) -- (17.75,7.1) -- (17.75,5.5) -- (18,5.5);

		\pic[xshift=12cm,rotate=90]{vae=$o_{t+1}$/$\hat{\mu}$/$\ln \hat{\sigma}$/$\hat{s}_{t+1}$/$\hat{o}_{t+1}$};
    }
}

\tikzset{
    pics/fountas/.style args={#1/#2/#3/#4/#5}{
    code = {
		\draw[fill=blue!20!white] (-1,-12) rectangle (0,-8);
	    \node at (-0.5, -10) {#1};

		\coordinate (A) at (0.5,-12);
		\coordinate (B) at (3,-11);
		\coordinate (C) at (3,-9);
		\coordinate (D) at (0.5,-8);
		\draw[fill=green!20!white] (A) -- coordinate[pos=.6] (AB) (B)--(C)
         --coordinate[pos=.45] (CD) (D)
         --coordinate[pos=.55] (DA) cycle;
	    \node at (1.75, -10) {Encoder};

		\draw[fill=red!20!white] (3.5,-11) rectangle (4.5,-10.1);
	    \node at (4, -10.5) {#3};
		\draw[fill=red!20!white] (3.5,-9.9) rectangle (4.5,-9);
	    \node at (4, -9.5) {#2};

		\node[latent] at (4, -8.4) (epsilon) {$\epsilon$};

		\coordinate (A) at (8,-4);
		\coordinate (B) at (10.5,-5);
		\coordinate (C) at (10.5,-1);
		\coordinate (D) at (8,-2);
		\draw[fill=green!20!white] (A) -- coordinate[pos=.6] (AB) (B)--(C)
         --coordinate[pos=.45] (CD) (D)
         --coordinate[pos=.55] (DA) cycle;
	    \node at (9.25, -3) {Decoder};

		\draw[fill=gray!20!white] (5,-11) rectangle (6,-9);
	    \node at (5.5,-10) {#4};

		\draw[fill=blue!20!white] (7,-11) rectangle (8,-9);
	    \node at (7.5, -10) {$a_t$};

		\draw[fill=blue!20!white] (11,-5) rectangle (12,-1);
	    \node at (11.5, -3) {#5};

		\draw[-latex] (7.5,-3) -- (8,-3);
		\draw[-latex] (10.5,-3) -- (11,-3);

		\draw[-latex] (3,-10.5) -- (3.5,-10.5);
		\draw[-latex] (3,-9.5) -- (3.5,-9.5);
		\draw[-latex] (0,-10) -- (0.5,-10);
		\draw[-latex] (4.36,-8.4) -- (4.75,-8.4) -- (4.75,-10) -- (5,-10);
		\draw[-latex] (4.5,-10.5) -- (4.75,-10.5) -- (4.75,-10) -- (5,-10);
		\draw[-latex] (4.5,-9.5) -- (4.75,-9.5) -- (4.75,-10) -- (5,-10);
		
		\coordinate (A) at (4.5,-8);
		\coordinate (B) at (5.5,-5.5);
		\coordinate (C) at (7.5,-5.5);
		\coordinate (D) at (8.5,-8);
		\draw[fill=orange!20!white] (A) -- coordinate[pos=.6] (AB) (B)--(C)
         --coordinate[pos=.45] (CD) (D)
         --coordinate[pos=.55] (DA) cycle;
	    \node at (6.5, -6.75) {Transition};
		
		\draw[-latex] (5.95,-5.5) -- (5.95,-5);
		\draw[-latex] (7.05,-5.5) -- (7.05,-5);

		\draw[-latex] (5.95,-4) -- (5.95,-3.75) -- (6.5,-3.75) -- (6.5,-3.5);
		\draw[-latex] (7.05,-4) -- (7.05,-3.75) -- (6.5,-3.75) -- (6.5,-3.5);
		\draw[-latex] (4.9,-4.1) -- (4.9,-3.75) -- (6.5,-3.75) -- (6.5,-3.5);

		\draw[-latex] (5.5,-9) -- (5.5,-8);
		\draw[-latex] (7.5,-9) -- (7.5,-8);

		\draw[fill=red!20!white] (5.5,-5) rectangle (6.4,-4);
	    \node at (5.95,-4.5) {$\mathring{\mu}$};
		\draw[fill=red!20!white] (6.6,-5) rectangle (7.5,-4);
	    \node at (7.05,-4.5) {$\ln \mathring{\sigma}$};
		\node[latent] at (4.9,-4.5) (epsilon) {$\epsilon$};

		\draw[fill=gray!20!white] (5.5,-3.5) rectangle (7.5,-2.5);
	    \node at (6.5, -3) {$\mathring{s}^r_{t+1}$};
    }}
}

\tikzset{
    pics/dense/.style args={#1/#2/#3/#4}{
    code = {
    		\draw[fill=green!30!white] (\fpeval{#1-1.75},\fpeval{#2-0.25}) rectangle (\fpeval{#1+1.75},\fpeval{#2+0.75});
		\node[rotate=90] at (\fpeval{#1-1.95},\fpeval{#2+0.25}) {#4};
		\node at (#1,#2+0.25) {\textbf{size:} #3};
    }}
}

\tikzset{
    pics/dropout/.style args={#1/#2/#3}{
    code = {
    		\draw[fill=blue!30!white] (\fpeval{#1-1.75},\fpeval{#2-0.25}) rectangle (\fpeval{#1+1.75},\fpeval{#2+0.75});
		\node at (#1,#2+0.25) {\textbf{rate:} #3};
    }}
}

\tikzset{
    pics/upconv/.style args={#1/#2/#3/#4/#5/#6}{
    code = {
    		\draw[fill=yellow!30!white] (\fpeval{#1-1.75},\fpeval{#2-0.75}) rectangle (\fpeval{#1+1.75},\fpeval{#2+1.25});
		\node at (#1,#2+0.75) {\textbf{size:} #3};
		\node at (#1,#2+0.25) {\textbf{kernel:} #4};
		\node at (#1,#2+-0.25) {\textbf{strides:} (#5,#5)};
		\node[rotate=90] at (\fpeval{#1-1.95},\fpeval{#2+0.25}) {#6};
    }}
}

\tikzset{
    pics/conv/.style args={#1/#2/#3/#4/#5/#6}{
    code = {
    		\draw[fill=orange!30!white] (\fpeval{#1-1.75},\fpeval{#2-0.75}) rectangle (\fpeval{#1+1.75},\fpeval{#2+1.25});
		\node at (#1,#2+0.75) {\textbf{size:} #3};
		\node at (#1,#2+0.25) {\textbf{kernel:} #4};
		\node at (#1,#2+-0.25) {\textbf{strides:} (#5,#5)};
		\node[rotate=90] at (\fpeval{#1-1.95},\fpeval{#2+0.25}) {#6};
    }}
}

\jmlrheading{1}{2020}{1-48}{4/00}{10/00}{meila00a}{Théophile Champion and Marek Grze\'s and Lisa Bonheme and Howard Bowman}

\ShortHeadings{Deconstructing deep active inference.}{Champion et al.}
\firstpageno{1}

\begin{document}

\title{Deconstructing deep active inference.}

\author{\name Théophile Champion \email tmac3@kent.ac.uk \\
       \addr University of Kent, School of Computing\\
       Canterbury CT2 7NZ, United Kingdom
       \AND
       \name Marek Grze\'s \email m.grzes@kent.ac.uk \\
       \addr University of Kent, School of Computing\\
       Canterbury CT2 7NZ, United Kingdom
       \AND
       \name Lisa Bonheme \email lb732@kent.ac.uk \\
       \addr University of Kent, School of Computing\\
       Canterbury CT2 7NZ, United Kingdom
       \AND
       \name Howard Bowman \email H.Bowman@kent.ac.uk \\
       \addr University of Birmingham, School of Psychology,\\
       Birmingham B15 2TT, United Kingdom\\
       University of Kent, School of Computing\\
       Canterbury CT2 7NZ, United Kingdom\\
       University College London, Wellcome Centre for Human Neuroimaging (honorary)\\
       London WC1N 3AR, United Kingdom
       }
       
\editor{\textbf{TO BE FILLED}} 

\maketitle


\begin{abstract}
Active inference is a theory of perception, learning and decision making, which can be applied to neuroscience, robotics, psychology, and machine learning. Recently, intensive reasearch has been taking place to scale up this framework using Monte-Carlo tree search and deep learning. The end-goal of this activity is to solve more complicated tasks using deep active inference. First, we review the existing literature, then, we progresively build a deep active inference agent as follows: (i) implement a variational auto-encoder (VAE), (ii) implement a deep hidden Markov model (HMM), (iii) implement a deep critical hidden Markov model (CHMM), and (iv) implement a complete deep active inference agent (DAI). For the CHMM and DAI agents, we have experimented with five definitions of the expected free energy and three different action selection strategies. According to our experiments, the models able to solve the dSprites environment are the ones that maximise rewards. Finally, we compare the similarity of the representation learned by the layers of various models (e.g., deep Q-network, CHMM, DAI) using centered kernel alignment. Importantly, the CHMM maximising reward and the CHMM minimising expected free energy learn very similar representations except for the last layer of the critic network (reflecting the difference in learning objective), and the variance layers of the transition and encoder networks. While performing further inspection of those (variance) layers, we found that the transition network of the reward maximising CHMM is a lot more certain than the transition network of the CHMM minimising expected free energy. More precisely, the CHMM minimising expected free energy is only confident about the world transition when performing action down. This suggests that the CHMM minimising expected free energy always picks the action down, and does not gather enough data for the other actions. In contrast, the CHMM maximising reward, keeps on selecting the actions left and right, enabling it to successfully solve the task. The only difference between those two CHMMs is the epistemic value, which aims to make the outputs of the transition and encoder networks as close as possible. Thus, the CHMM minimising expected free energy repeatedly picks a single action (down), and becomes an expert at predicting the future when selecting this action. This effectively makes the KL divergence between the output of the transition and encoder networks small. Additionally, when selecting the action down the average reward is zero, while for all the other actions, the expected reward will be negative. Therefore, if the CHMM has to stick to a single action to keep the KL divergence small, then the action down is the most rewarding. Thus, the appropriate formulation of the epistemic value in deep active inference remains an open question.
\end{abstract}


\section{Introduction}

Active inference is a unified framework for perception, learning, and planning that has emerged from theoretical neuroscience \citep{AI_TUTO,AI_VMP,AITs_tHEORY,
AITS_PRACTICE,BTAI_BF}. This framework has successfully explained a wide range of brain phenomena  \citep{FRISTON2016862,bayes_surprise,curiosity,dopamine}, and has been applied to a large number of tasks in robotics and artificial intelligence \citep{DeepAIwithMCMC,pezzato2020active,
sancaktar2020endtoend,ccatal2020learning,CULLEN2018809,cart_pole}.

A promising area of research revolves around scaling up this theoretical framework to tackle increasingly complex tasks. Research towards this goal is generally driven from recent advances in machine learning. For example, variational auto-encoders \citep{VAE,beta-VAE,Kingma2013,Rezende2014} have been key to the integration of deep neural networks within active inference \citep{sancaktar2020endtoend,ccatal2020learning,DeepAI}, and the Monte Carlo tree search algorithm \citep{MCTS,Go} has been used to improve planning efficency \citep{DeepAIwithMCMC,AITs_tHEORY,
AITS_PRACTICE,BTAI_BF,BTAI_3MF}.

Another closely related field is reinforcement learning \citep{DeepRL,DDQN,lample2016playing}, which addresses the same kind of tasks, where an agent must interact with its environment. A known challenge in this field is the correlation between the consecutive samples, which violates the standard i.i.d. assumption on which most of machine learning relies. To break this correlation, researchers proposed to store past experiences of the agent inside a replay buffer \citep{DeepRL}. Experiences can then be re-sampled randomly from the buffer to train the Q-network, which is used to approximate Q-values. The Q-network is trained to minimize the mean squared error between its output and a target value, which is defined as:
$$y(o_t,a_t) = \mathbb{E}_{o_{t+1} \sim E(o_t,a_t)}\Big[ r_t + \gamma \max_{a_{t+1} \in \mathcal{A}} \mathcal{Q}_{\theta_a}(o_{t+1}, a_{t+1})\Big],$$
where $t$ is the present time step, $\mathcal{A}$ is the set of available actions, $y(o_t,a_t)$ is the target Q-value to be predicted, $\mathbb{E}$ is the expectation w.r.t the observations received from the environment, $r_t$ is the reward obtained by the agent when performing action $a_t$ in state\footnote{Note, we are using the notation $o_\tau$ for the (observable) state at (an arbitrary) time step $\tau$, instead of the more standard notation $s_\tau$. This is because we reserve the notation $s_\tau$ for the (unobserved) states that arise in the context of active inference.} $o_t$, $o_{t+1}$ is the state reached when performing action $a_t$ in state $o_t$, $E$ is the environment emulator from which $o_{t+1}$ is sampled, $\gamma$ is the discount factor that discounts future rewards, and $\mathcal{Q}_{\theta_a}(o_{t+1}, a_{t+1})$ is the output of the Q-network, i.e., the estimated Q-value of performing action $a_{t+1}$ in state $o_{t+1}$.

Unfortunatly, using the above target to train the Q-network can make the training unstable. Generally, the problem is addressed by introducing a target network $\hat{\mathcal{Q}}_{\hat{\theta}_a}(o_{t+1}, a_{t+1})$, which is simply a copy of the Q-network. The weights of the target network are then synchronized with the weights of the Q-network every $K$ (learning) iterations \citep{DeepRL}. The new target is obtained by replacing the Q-network by the target network, i.e.,
$$y(o_t,a_t) = \mathbb{E}_{o_{t+1} \sim E(o_t,a_t)}\Big[ r_t + \gamma \max_{a_{t+1} \in \mathcal{A}} \hat{\mathcal{Q}}_{\hat{\theta}_a}(o_{t+1}, a_{t+1})\Big].$$
In Section \ref{sec:existing_research}, we review the existing literature and present: the Deep Q-network (DQN) agent \citep{DeepRL}, the deep active inference with Monte-Carlo methods ($DAI_{MC}$) agent by \citet{DeepAIwithMCMC}, the deep active inference as variational policy gradients ($DAI_{VPG}$) approach by \citet{DeepAI}, the deep active inference agent of rubber hand illusion ($DAI_{RHI}$) by \citet{rood2020deep}, the deep active inference agent for humanoid robot control ($DAI_{HR}$) by \citet{sancaktar2020endtoend,DAI_HR,DAI_HR2}, the deep active inference agent based on the free action objective ($DAI_{FA}$) by \citet{DAI_Kai}, a deep active inference agent for partially observable Markov decision processes ($DAI_{POMDP}$) by \citet{DAI_POMDP}, as well as various methods for which the code is not available online. We argue that while all these approaches illuminate important issues associated with realising a deep active inference agent, a fully complete implementation has not yet been published. Consequently, to systematically explore the construction of deep active inference agents, in Section \ref{sec:build_dai}, we incrementally build such an agent. We start with a simple variational auto-encoder (VAE) composed of an encoder and decoder network. Next, a transition network is added to create a deep hidden Markov model (HMM). Then, a critic network is added to define a prior over actions, which leads to the critical HMM (CHMM). Lastly, the policy network is added to approximate the posterior over actions leading to the full deep active inference (DAI) agent. Then, in Section \ref{sec:results}, we discuss our findings regarding the abilities and limitations of each intermediate step. This section also presents an analysis and discussion of the representations learned by each intermediate model. Finally, Section \ref{sec:conclusion} puts our findings in context and concludes this paper.

\section{Review of existing research} \label{sec:existing_research}

In this section, we discuss the DQN agent from the reinforcement learning literature, six agents from the active inference literature for which the code is available online ($DAI_{MC}$, $DAI_{VPG}$, $DAI_{RHI}$, $DAI_{HR}$, $DAI_{FA}$, and $DAI_{POMDP}$), and a few other deep active inference agents for which the code is unavailable. Finally, we explain how the representations learned by the agents can be compared using centered kernel alignment. Note, the notation used throughout this section is summarised in Appendix A.

\subsection{$DQN$ agent \citep{DeepRL}} \label{ssec:dqn}

\begin{figure}[H]
	\begin{center}
	\begin{tikzpicture}[square/.style={regular polygon,regular polygon sides=4}, scale=1]
		\draw[fill=blue!20!white] (-1,-12) rectangle (1,-8);
	    \node at (0, -10) {$o_t$};

		\coordinate (A) at (1.5,-12);
		\coordinate (B) at (4,-11);
		\coordinate (C) at (4,-9);
		\coordinate (D) at (1.5,-8);
		\draw[fill=green!20!white] (A) -- coordinate[pos=.6] (AB) (B)--(C)
         --coordinate[pos=.45] (CD) (D)
         --coordinate[pos=.55] (DA) cycle;
	    \node at (2.75, -10) {$\mathcal{Q}_{\theta_a}$};

		\draw[fill=gray!20!white] (4.5,-11) rectangle (6.5,-9);
	    \node at (5.5, -10) {$\mathcal{Q}_{\theta_a}(o_t, \bigcdot\,)$};

		\draw[fill=blue!20!white] (-1,-7) rectangle (1,-3);
	    \node at (0, -5) {$o_{t+1}$};

		\coordinate (A) at (1.5,-7);
		\coordinate (B) at (4,-6);
		\coordinate (C) at (4,-4);
		\coordinate (D) at (1.5,-3);
		\draw[fill=red!20!white] (A) -- coordinate[pos=.6] (AB) (B)--(C)
         --coordinate[pos=.45] (CD) (D)
         --coordinate[pos=.55] (DA) cycle;
	    \node at (2.75, -5) {$\hat{\mathcal{Q}}_{\hat{\theta}_a}$};

		\draw[fill=orange!20!white] (4.5,-6) rectangle (6.5,-4);
	    \node at (5.5, -5) {$\hat{\mathcal{Q}}_{\hat{\theta}_a}(o_{t+1}, \bigcdot\,)$};

		\draw[fill=blue!20!white] (6,-6.5) rectangle (7,-7.5);
	    \node at (6.5, -7) {$r_t$};

		\draw[fill=blue!20!white] (7.5,-7.5) rectangle (8.5,-8.5);
	    \node at (8, -8) {$\gamma$};

		\draw[fill=yellow!20!white] (7,-4) rectangle (9,-6);
	    \node at (8, -5) {$y(o_t,\bigcdot\,)$};

		\draw[fill=pink!60!white] (10,-6.5) rectangle (12,-8.5);
	    \node at (11, -7.5) {$MSE$};

		\draw[-latex] (4,-10) -- (4.5,-10);
		\draw[-latex] (1,-10) -- (1.5,-10);
		\draw[-latex] (4,-5) -- (4.5,-5);
		\draw[-latex] (1,-5) -- (1.5,-5);
		\draw[-latex] (6.5,-5) -- (7,-5);
		\draw[-latex] (7,-7) -- (8,-7) -- (8,-6);
		\draw[-latex] (9,-5) -- (11,-5) -- (11,-6.5);
		\draw[-latex] (6.5,-10) -- (11,-10) -- (11,-8.5);         
		\draw[-latex] (8,-7.5) -- (8,-6);
    \end{tikzpicture}

	\end{center}
  \caption{This figure illustrates the DQN agent. Briefly, the image $o_t$ is fed into the Q-network, and the image $o_{t+1}$ is fed into the target network. The Q-network outputs the Q-values for each action at time $t$, and the target network outputs the Q-values for each action at time $t+1$. Then, the reward, the discount factor, and Q-values of each action at time $t+1$ are used to compute the target values $y(o_t,\bigcdot\,)$. Finally, the goal is to minimise the MSE between the prediction of the Q-network and the target values by changing the weights of the Q-network.}
   \label{fig:DQN}
\end{figure}
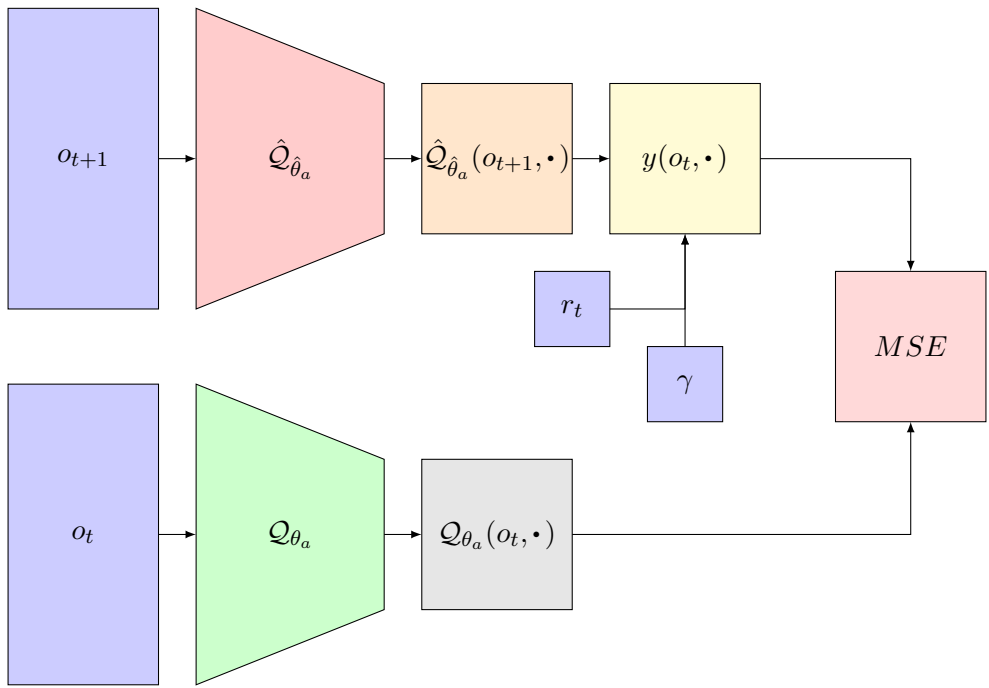

Let us start with the DQN agent \citep{DeepRL}, whose goal is to maximise the amount of reward obtained over time. At each time step $\tau$, the agent is observing an image $o_\tau$, and is allowed to perform one action $a_\tau \in \mathcal{A}$. After performing $a_\tau$ when observing $o_\tau$, the agent receives a reward $r_\tau$. The Q-learning algorithm \citep{sutton1998} aims to maximise reward by computing the Q-values $Q(o_\tau, a_\tau)$, for each state-action pair ($o_\tau$, $a_\tau$). The Q-values represent the expected amount of rewards obtained by taking action $a_\tau$ in state $o_\tau$. This approach is intractable for image based domains such as Atari games, since one would need to store a vector of Q-values for each possible image. Instead, the DQN algorithm (illustrated in Figure \ref{fig:DQN}) has been developed, which uses a deep neural network $\mathcal{Q}_{\theta_a}$ to approximate the Q-values. More formally, $\mathcal{Q}_{\theta_a}$ maps any observation to a vector of size $\#A$ containing the Q-values of each possible action, and we denote by $\mathcal{Q}_{\theta_a}(o_\tau, a_\tau)$ the element at position $a_\tau$ in the output vector predicted by $Q_{\theta_a}$ when provided with the image $o_\tau$. As we discussed in the introduction, the training stability of the Q-network is improved by introducing a target network $\hat{\mathcal{Q}}_{\hat{\theta}_a}$, which is structurally identical to the Q-network and whose weights are synchronised with the weights of the Q-network every $K$ (learning) iterations. The Q-network's weights are then optimised using gradient descent to minimise the mean square error between the output of the Q-network and a target value, i.e., $\theta^*_a = \argmin_{\theta_a} \text{MSE}[\mathcal{Q}_{\theta_a}(o_t,\bigcdot\,), y(o_t,\bigcdot\,)]$, where $y(o_t,a_t)$ is the target Q-value for each state-action pair, and, as highlighted earlier, is defined as follows:
$$y(o_t,a_t) = \mathbb{E}_{o_{t+1} \sim E(o_t,a_t)}\Big[ r_t + \gamma \max_{a_{t+1} \in \mathcal{A}} \hat{\mathcal{Q}}_{\hat{\theta}_a}(o_{t+1}, a_{t+1})\Big].$$

\subsection{$DAI_{MC}$ agent \citep{DeepAIwithMCMC}} \label{ssection:fountas_paper}

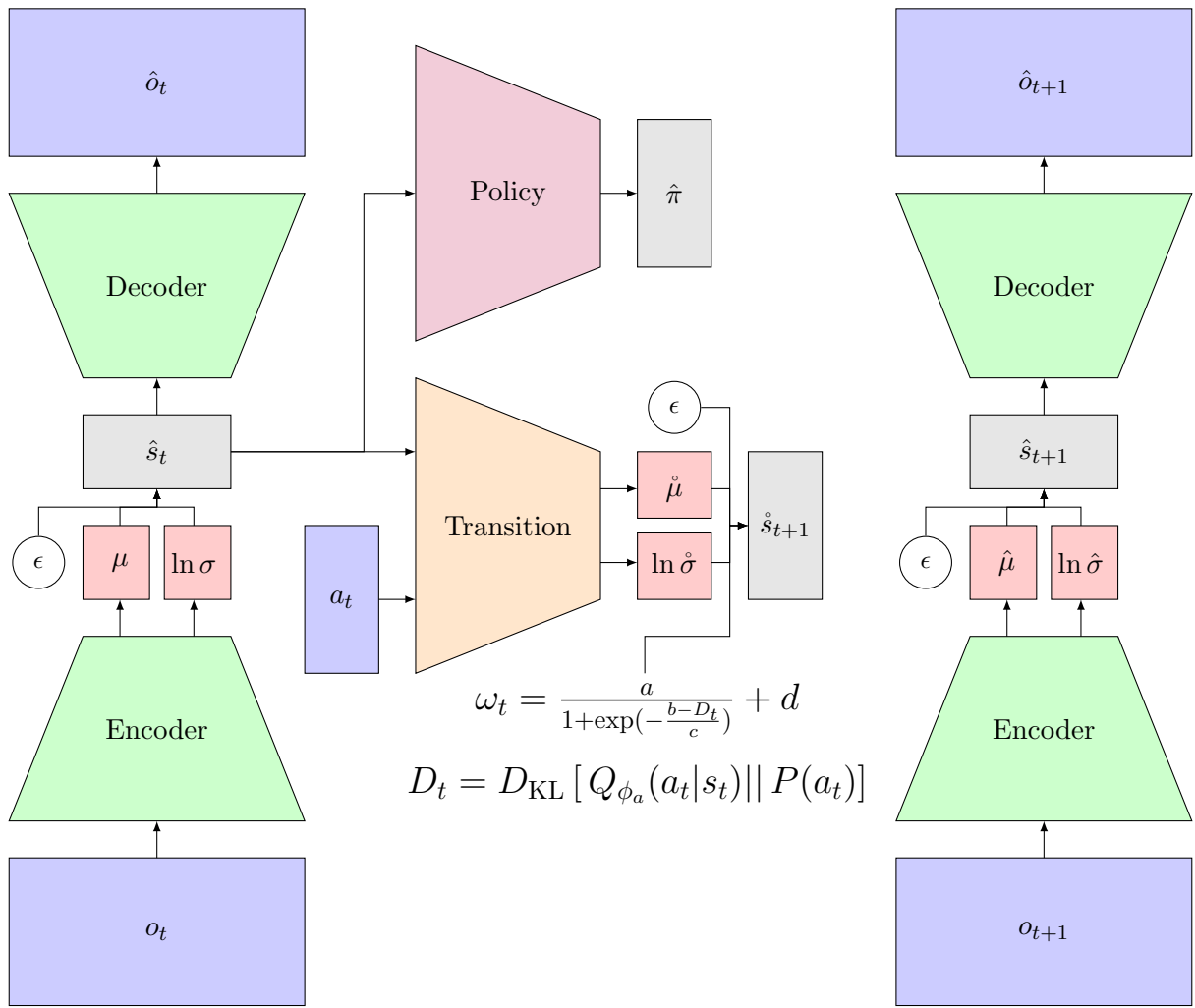
\begin{figure}[h]
	\begin{center}
	\begin{tikzpicture}[square/.style={regular polygon,regular polygon sides=4}]
		\coordinate (A) at (13.5,8);
		\coordinate (B) at (16,9);
		\coordinate (C) at (16,11);
		\coordinate (D) at (13.5,12);
		\draw[fill=purple!20!white] (A) -- coordinate[pos=.6] (AB) (B)--(C)
         --coordinate[pos=.45] (CD) (D)
         --coordinate[pos=.55] (DA) cycle;
	    \node at (14.75, 10) {Policy};

		\draw[-latex] (11,6.5) -- (12.8,6.5) -- (12.8,10) -- (13.5,10);
		\draw[-latex] (16, 10) -- (16.5, 10);
		
		\draw[fill=gray!20!white] (16.5,9) rectangle (17.5,11);
	    \node at (17, 10) {$\hat{\pi}$};

		\draw (16.6, 3.5) -- (16.6, 4) -- (17.75, 4) -- (17.75, 5.5);
	    \node at (16.5, 3) {\Large $\omega_t = \frac{a}{1 + \exp(-\frac{b - D_t}{c})} + d$};
		\node at (16.5, 2) {\Large $D_t = \kl{Q_{\phi_a}(a_t|s_t)}{P(a_t)}$};

		\pic{hmm};
    \end{tikzpicture}
	\end{center}
  \caption{This figure illustrates the $DAI_{MC}$ agent, which is composed of an encoder, a decoder, a transition network, and a policy network. The same VAE (encoder and decoder) is repeated in the figure to reflect successive time-points.}
  \label{fig:DAI_MC_agent}
\end{figure}

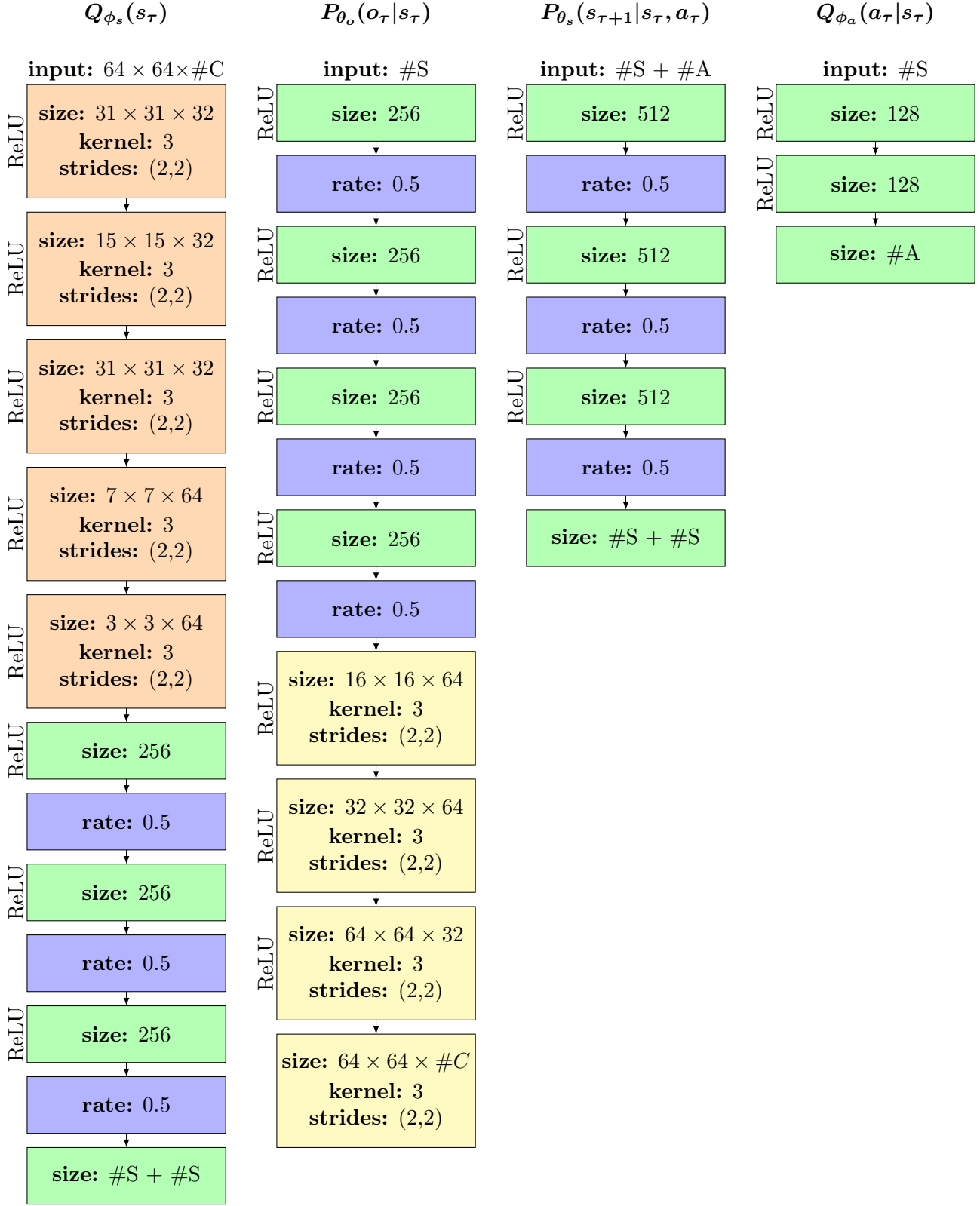
\begin{figure}
	\begin{center}
	\begin{tikzpicture}
		\node (Q_s) at (-6.2,8) {$\bm{Q_{\phi_s}(s_\tau)}$};
		\node at (-6.2,7) {\textbf{input:} $64\times64\times$\#C};
		\pic{conv=-6.2/5.5/$31\times31\times32$/3/2/ReLU};
		\pic{conv=-6.2/3.25/$15\times15\times32$/3/2/ReLU};
		\pic{conv=-6.2/1/$31\times31\times32$/3/2/ReLU};
		\pic{conv=-6.2/-1.25/$7\times7\times64$/3/2/ReLU};
		\pic{conv=-6.2/-3.5/$3\times3\times64$/3/2/ReLU};
		\pic{dense=-6.2/-5.25/256/ReLU};
		\pic{dropout=-6.2/-6.5/0.5};
		\pic{dense=-6.2/-7.75/256/ReLU};
		\pic{dropout=-6.2/-9/0.5};
		\pic{dense=-6.2/-10.25/256/ReLU};
		\pic{dropout=-6.2/-11.5/0.5};
		\pic{dense=-6.2/-12.75/\#S + \#S/};

		\draw[-latex] (-6.2,4.75) -- (-6.2,4.5);
		\draw[-latex] (-6.2,2.5) -- (-6.2,2.25);
		\draw[-latex] (-6.2,0.25) -- (-6.2,0);
		\draw[-latex] (-6.2,-2) -- (-6.2,-2.25);
		\draw[-latex] (-6.2,-4.25) -- (-6.2,-4.5);
		\draw[-latex] (-6.2,-5.5) -- (-6.2,-5.75);
		\draw[-latex] (-6.2,-6.75) -- (-6.2,-7);
		\draw[-latex] (-6.2,-8) -- (-6.2,-8.25);
		\draw[-latex] (-6.2,-9.25) -- (-6.2,-9.5);
		\draw[-latex] (-6.2,-10.5) -- (-6.2,-10.75);
		\draw[-latex] (-6.2,-11.75) -- (-6.2,-12);

		\node (P_o) at (-1.8,8) {$\bm{P_{\theta_o}(o_\tau|s_\tau)}$};
		\node at (-1.8,7) {\textbf{input:} \#S};
		\pic{dense=-1.8/6/256/ReLU};
		\pic{dropout=-1.8/4.75/0.5};
		\pic{dense=-1.8/3.5/256/ReLU};
		\pic{dropout=-1.8/2.25/0.5};
		\pic{dense=-1.8/1/256/ReLU};
		\pic{dropout=-1.8/-0.25/0.5};
		\pic{dense=-1.8/-1.5/256/ReLU};
		\pic{dropout=-1.8/-2.75/0.5};
		\pic{upconv=-1.8/-4.5/$16\times16\times64$/3/2/ReLU};
		\pic{upconv=-1.8/-6.75/$32\times32\times64$/3/2/ReLU};
		\pic{upconv=-1.8/-9/$64\times64\times32$/3/2/ReLU};
		\pic{upconv=-1.8/-11.25/$64\times64\times\#C$/3/2/};

		\draw[-latex] (-1.8,5.75) -- (-1.8,5.5);
		\draw[-latex] (-1.8,4.5) -- (-1.8,4.25);
		\draw[-latex] (-1.8,3.25) -- (-1.8,3);
		\draw[-latex] (-1.8,2) -- (-1.8,1.75);
		\draw[-latex] (-1.8,0.75) -- (-1.8,0.5);
		\draw[-latex] (-1.8,-0.5) -- (-1.8,-0.75);
		\draw[-latex] (-1.8,-1.75) -- (-1.8,-2);
		\draw[-latex] (-1.8,-3) -- (-1.8,-3.25);
		\draw[-latex] (-1.8,-5.25) -- (-1.8,-5.5);
		\draw[-latex] (-1.8,-7.5) -- (-1.8,-7.75);
		\draw[-latex] (-1.8,-9.75) -- (-1.8,-10);

		\node (P_s) at (2.6,8) {$\bm{P_{\theta_s}(s_{\tau+1}|s_\tau,a_\tau)}$};
		\node at (2.6,7) {\textbf{input:} \#S + \#A};
		\pic{dense=2.6/6/512/ReLU};
		\pic{dropout=2.6/4.75/0.5};
		\pic{dense=2.6/3.5/512/ReLU};
		\pic{dropout=2.6/2.25/0.5};
		\pic{dense=2.6/1/512/ReLU};
		\pic{dropout=2.6/-0.25/0.5};
		\pic{dense=2.6/-1.5/\#S + \#S/};

		\draw[-latex] (2.6,5.75) -- (2.6,5.5);
		\draw[-latex] (2.6,4.5) -- (2.6,4.25);
		\draw[-latex] (2.6,3.25) -- (2.6,3);
		\draw[-latex] (2.6,2) -- (2.6,1.75);
		\draw[-latex] (2.6,0.75) -- (2.6,0.5);
		\draw[-latex] (2.6,-0.5) -- (2.6,-0.75);
		
		\node (Q_a) at (7,8) {$\bm{Q_{\phi_a}(a_\tau|s_\tau)}$};
		\node at (7,7) {\textbf{input:} \#S};
		\pic{dense=7/6/128/ReLU};
		\pic{dense=7/4.75/128/ReLU};
		\pic{dense=7/3.5/\#A/};

		\draw[-latex] (7,5.75) -- (7,5.5);
		\draw[-latex] (7,4.5) -- (7,4.25);

	\end{tikzpicture}
	\end{center}
	\caption{Neural network architectures of the $DAI_{MC}$ agent. Orange blocks correspond to convolutional layers, green blocks correspond to fully connected layers, blue blocks correspond to dropout, and yellow blocks correspond to up-convolutional layers. For the dSprites environment, there are four actions (i.e., \#A = 4), ten states (i.e., \#S = 10), and only one channel (i.e., \#C = 1). For the Animal-AI environment, there are three actions (i.e., \#A = 3), ten states (i.e., \#S = 10), and three channels (i.e., \#C = 3). These are all trained to minimize variational free energy.}
	\label{fig:fountas_dnn}
\end{figure}

In this section, we review the $DAI_{MC}$ agent proposed by \citet{DeepAIwithMCMC}, which represents the most ambitious and complete implementation of a deep active inference agent that accordingly adds important new concepts to the field. The relevant code is available at the following URL: \url{https://github.com/zfountas/deep-active-inference-mc}. The $DAI_{MC}$ agent is composed of four deep neural networks, as illustrated in Figures \ref{fig:DAI_MC_agent} and \ref{fig:fountas_dnn}. The encoder $\mathcal{E}_{\phi_s}$ takes images as input, and outputs the mean and variance of the variational distribution over hidden states, i.e., $Q_{\phi_s}(s_t) = \mathcal{N}(s_t;\mu, \sigma)$, where $\mu, \sigma = \mathcal{E}_{\phi_s}(o_t)$. The decoder $\mathcal{D}_{\theta_o}$ takes a state as input, and outputs the parameters of a product of Bernoulli distributions, which can be interpreted as the expected (reconstructed) image $\hat{o}_t$, i.e., 
\begin{align*}
P_{\theta_o}(o_t|s_t) = \MultiBernoulli(o_t;\hat{o}_t),
\end{align*}
where $\hat{o}_t = \mathcal{D}_{\theta_o}(s_t)$ are the values predicted by the decoder, and $\MultiBernoulli(o_t;\hat{o}_t)$ is a product of Bernoulli distributions defined as:
$$\MultiBernoulli(o_t;\hat{o}_t) = \prod_{x,y} \text{Bernoulli}(o_t[x,y];\hat{o}_t[x,y]),$$
where $\text{Bernoulli}(\,\bigcdot\,;\,\bigcdot\,)$ is a Bernoulli distribution over the possible values of the pixel $o_t[x,y]$, parameterized by the parameter $\hat{o}_t[x,y]$. The transition network $\mathcal{T}_{\theta_s}$ takes a state-action pair as input, and outputs the mean and variance of a Gaussian distribution over hidden states, i.e., $P_{\theta_s}(s_{\tau+1}|s_\tau, a_\tau) = \mathcal{N}(s_{\tau+1};\mathring{\mu}, \frac{\mathring{\sigma}}{\omega_t})$, where $\mathring{\mu}, \mathring{\sigma} = \mathcal{T}_{\theta_s}(s_\tau, a_\tau)$, and $\omega_t$ is the top-down attention parameter modulating the precision of the transition mapping (see below). The policy network $\mathcal{P}_{\phi_a}$ takes a state as input, and outputs a distribution over actions, i.e., $Q_{\phi_a}(a_t|s_t) = \text{Cat}(a_t;\hat{\pi})$, where $\hat{\pi} = \mathcal{P}_{\phi_a}(s_t)$. Finally, the prior over actions is defined as follows:
\begin{align}
P(a_t) = \sum_{\pi \in \Pi} [\pi_t = a_t]P(\pi),\label{eq:initial_prior_actions}
\end{align}
where $\Pi$ is the set of all possible policies, $\pi_t$ is the action precribed by policy $\pi$ at time $t$, the square brackets represent an indicator function that equals one if the condition within the bracket is satisfied and zero otherwise, and $P(\pi)$ is the prior over policies defined as: \begin{align*}
P(\pi) = \sigma[-G(\pi)],
\end{align*}
where $\sigma[\,\bigcdot\,]$ is the softmax function, and $G(\pi)$ is the expected free energy (EFE) of policy $\pi$, which is defined as:
\begin{align}
G(\pi) = \sum_{\tau = t}^T G_{\tau}(\pi) = \sum_{\tau = t}^T \mathbb{E}_{\tilde{Q}}\Big[ \ln Q(s_\tau, \theta|\pi) - \ln \tilde{P}(o_\tau, s_\tau,\theta|\pi) \Big],\label{eq:efe_fountas_defin}
\end{align}
where $\tilde{Q} = Q(o_\tau, s_\tau, \theta|\pi) = Q(\theta|\pi)Q(s_\tau|\theta,\pi)Q(o_\tau|s_\tau,\theta,\pi)$ is the predictive posterior, and $\tilde{P}(o_\tau,s_\tau,\theta|\pi) = P(\theta|s_\tau,o_\tau,\pi)P(s_\tau|o_\tau,\pi)P(o_\tau|\pi)$ is the target distribution. However, Equation \ref{eq:efe_fountas_defin} needs to be re-arranged to be computed in practice\footnote{By ``in practice", we mean ``in the code" or equivalently ``when implementing the approach".}, and Section \ref{ssection:efe_derivation_assumptions} will present this derivation. Finally, as shown in Figure \ref{fig:DAI_MC_agent}, the top-down attention parameter is computed as follows:
\begin{align*}
\omega_t = \frac{a}{1 + \exp(-\frac{b - D_t}{c})} + d,
\end{align*}
where $D_t = \kl{Q_{\phi_a}(a_t|s_t)}{P(a_t)}$, and $\{a, b, c, d\}$ are fixed hyperparameters. Intuitively, $\omega_t$ is high when the posterior over actions (from the policy network) is close to the prior over actions (from the expected free energy), and low when the posterior is far away from the prior. This, in turn, means that extra uncertainty is introduced into the transition mapping (see paragraph before Equation \ref{eq:initial_prior_actions}) when posterior over actions and prior over actions are very different. Finally, note that the number of terms required to compute the prior over actions (defined in Equation \ref{eq:initial_prior_actions}) grows exponentially with the time horizon of planning. Because this is intractable in practice, \citet{DeepAIwithMCMC} implemented a Monte-Carlo tree search (MCTS) algorithm to evaluate the expected free energy of each action (see below). Finally, action selection is performed by sampling from the following distribution:
\begin{align*}
\tilde{P}(a_t) = \frac{N(\hat{s}_t, a_t)}{\sum_{\hat{a}_t}N(\hat{s}_t, \hat{a}_t)},
\end{align*}
where $\hat{s}_t$ is the current state of the environment, and $N(s_t, a_t)$ is the number of times action $a_t$ has been visited from state $s_t$ during MCTS.

\subsubsection{The Monte-Carlo tree search}

In this section, we describe the planning algorithm used by $DAI_{MC}$, i.e., Monte-Carlo tree search (MCTS). MCTS is used to enhance the planning ability of the agent by allowing it to look into the future. At the beginning of an action-perception cycle, the agent is provided with an image $o_t$. This image can be feed into the encoder to get the mean vector $\mu$ of the posterior over the latent states, i.e., $Q_{\phi_s}(s_t) = \mathcal{N}(s_t;\mu, \sigma)$. Since $\mu$ is the mean of the Gaussian posterior, it can be interpreted as the maximum a posteriori (MAP) estimate of the latent states at time step $t$. This MAP estimate will constitute the root node of the Monte-Carlo tree search (MCTS).

The first step of the MCTS is to use the Upper Confidence bounds for Trees (UCT) criterion to determine which node in the tree should be expanded. Let the tree's root $\hat{s}_t$ be called the current node, which is denoted $\hat{s}_\tau$. If the current node has no children (i.e., no previously selected actions from the current node), then it is selected for expansion. Alternatively, the child with the highest UCT criterion becomes the new current node and the process is iterated until we reach a leaf node (i.e. a node from which no action has previously been selected). The UCT criterion \citep{MCTS} of the child of $\hat{s}_\tau$ corresponding to action $\hat{a}_\tau$ is given by:
\begin{align*}
UCT(\hat{s}_\tau,\hat{a}_\tau) = - \bar{G}(\hat{s}_\tau,\hat{a}_\tau) + C_{explore} \cdot \frac{Q_{\phi_a}(a_\tau=\hat{a}_\tau|s_\tau=\hat{s}_\tau)}{1 + N(\hat{s}_\tau,\hat{a}_\tau)},
\end{align*}
where $\bar{G}(\hat{s}_\tau,\hat{a}_\tau)$ is the average expected free energy of taking action $\hat{a}_\tau$ in state $\hat{s}_\tau$, $C_{explore}$ is the exploration constant that modulates the amount of exploration at the tree level, $N(\hat{s}_\tau,\hat{a}_\tau)$ is the number of times action $\hat{a}_\tau$ was visited in state $\hat{s}_\tau$, and $Q_{\phi_a}(a_\tau=\hat{a}_\tau|s_\tau=\hat{s}_\tau)$ is the posterior probability of action $\hat{a}_\tau$ in state $\hat{s}_\tau$ as predicted by the policy network.

Let $\mathring{s}_\tau$ be the (leaf) node selected by the above selection procedure. The MCTS then expands one of the children of $\mathring{s}_\tau$. The expansion uses the transition network to compute the mean $\mathring{\mu}$ of $P_{\theta_s}(s_{\tau+1}|s_\tau=\mathring{s}_\tau, a_\tau=\mathring{a}_\tau)$, which is viewed as a MAP estimate of the states at time $\tau+1$. Then, we need to estimate the cost of (virtually) taking action $\mathring{a}_\tau$. By definition, the cost is the expected free energy given by \eqref{eq:efe_fountas_defin}, and Monte-Carlo rollouts can be run to improve its estimation. The final step of the planning iteration is to back-propagate the cost of the newly expanded (virtual) action toward the root of the tree. Formally, we write the update as follows:
\begin{align}\label{eq:backprop}
\forall s \in \mathbb{A}_{\mathring{s}_\tau} \cup \{\mathring{s}_\tau \}, \quad \bm{G}_{s} \leftarrow \bm{G}_{s} + \bm{G}_{\mathring{s}_\tau},
\end{align}
where $\mathring{s}_\tau$ is the node that was selected for expansion, $\bm{G}_{s}$ is the expected free energy of $s$, and $\mathbb{A}_{\mathring{s}_\tau}$ is the set of all ancestors of $\mathring{s}_\tau$ in the tree. During the back propagation, we also update the number of visits as follows:
\begin{align}\label{eq:backprop_n}
\forall s \in \mathbb{A}_{\mathring{s}_\tau} \cup \{\mathring{s}_\tau \}, \quad \bm{N}_{s} \leftarrow \bm{N}_{s} + 1.
\end{align}
If we let $\bm{G}^{aggr}_s$ be the aggregated cost of an arbitrary node $s$ obtained by applying Equation \ref{eq:backprop} after each expansion, then we are now able to express $\bar{\bm{G}}_s$ formally as:
$$\bar{\bm{G}}_s = \frac{\bm{G}^{aggr}_s}{\bm{N}_s}.$$
Importantly, if the node $s$ corresponds to the state reached from state $\hat{s}_\tau$ by performing action $\hat{a}_\tau$, then $\bar{G}(\hat{s}_\tau, \hat{a}_\tau) = \bar{\bm{G}}_s$ and $N(\hat{s}_\tau, \hat{a}_\tau) = \bm{N}_s$. The planning procedure described above ends when the maximum number of planning iterations is reached, or when a clear winner has been identified, i.e., if $\max_{a_t} P(a_t)-\frac{1}{\#A}>T_{dec}$ where $\#A$ is the number of possible actions, and $T_{dec}$ is a (threshold) hyperparameter.

\subsubsection{Derivation of the variational free energy}\label{ssec:derive_vfe_in_fountas}

In this section, we provide a derivation for the variational free energy used by \citet{DeepAIwithMCMC}. This derivation was introduced in \citep{DeepAI}, and can be adapted to derive the variational free energy of the models presented in Section \ref{sec:build_dai}. Recall, the goal of the variational free energy as classically presented is to make the approximate posterior $Q_\phi(s_t,a_t)$ as close as possible to the true posterior\footnote{By posterior we mean a conditional distribution, where the given variables are those for which we observe a specific value. Note, the value of $s_{t-1}$ is unknown but can be sampled from the posterior over $s_{t-1}$ (from the previous action-perception cycle).} $P(s_t,a_t|o_t,s_{t-1},a_{t-1})$, i.e.,
\begin{align*}
Q^*_\phi(s_t,a_t) = \argmin_{Q_\phi(s_t,a_t)} \kl{Q_\phi(s_t,a_t)}{P(s_t,a_t|o_t,s_{t-1},a_{t-1})}.
\end{align*}
Using Bayes theorem, the linearity of expectation, and the fact that $Q_\phi(s_t,a_t)$ integrates to one:
\begin{align*}
Q_\phi^*(s_t,a_t) &= \argmin_{Q_\phi(s_t,a_t)} \kl{Q_\phi(s_t,a_t)}{P(s_t,a_t|o_t,s_{t-1},a_{t-1})}\\
&= \argmin_{Q_\phi(s_t,a_t)} \kl{Q_\phi(s_t,a_t)}{P(s_t,a_t,o_t,s_{t-1},a_{t-1})} + \underbrace{\ln P(o_t,s_{t-1},a_{t-1})}_{\text{Constant w.r.t }Q_\phi(s_t,a_t)}\\
&= \argmin_{Q_\phi(s_t,a_t)} \kl{Q_\phi(s_t,a_t)}{P(s_t,a_t,o_t,s_{t-1},a_{t-1})}.
\end{align*}
Using the d-separation criteria \citep{koller2009probabilistic}, it can be shown that:
$$P(s_t,a_t,o_t,s_{t-1},a_{t-1}) = P_{\theta_o}(o_t|s_t)P(a_t)P_{\theta_s}(s_t|s_{t-1},a_{t-1})Q_{\phi}(s_{t-1}, a_{t-1}),$$
where $Q_\phi(s_{t-1}, a_{t-1})$ is the variational posterior obtained through the inference process at the previous time step. In the above equation, $Q_\phi(s_{t-1}, a_{t-1})$ was used to replace $P(s_{t-1}, a_{t-1})$, i.e., $Q_\phi(s_{t-1}, a_{t-1})$ was used as an empirical prior. Additionally, since $Q_\phi(s_{t-1}, a_{t-1})$ is a constant w.r.t $Q_\phi(s_t,a_t)$, the above minimization problem reduces to:
\begin{align}
Q^*_{\phi}(s_t,a_t)\quad &= \,\, \argmin_{\quad Q_{\phi}(s_t,a_t) \,\,\,\,\,\, \quad }\underbrace{\kl{Q_{\phi}(s_t,a_t)}{P_{\theta_o}(o_t|s_t)P(a_t)P_{\theta_s}(s_t|s_{t-1},a_{t-1})}}_{\text{variational free energy}}\label{eq:def_vfe_fountas}\\
&= \argmin_{Q_{\phi_a}(a_t|s_t)Q_{\phi_s}(s_t)}\, \mathbb{E}_{Q_{\phi_s}(s_t)}\Big[ \kl{Q_{\phi_a}(a_t|s_t)}{P(a_t)} \Big] + \kl{Q_{\phi_s}(s_t)}{P_{\theta_s}(s_t|s_{t-1},a_{t-1})}\nonumber\\
&\hspace{8.9cm}- \mathbb{E}_{Q_{\phi_s}(s_t)}\Big[\ln P_{\theta_o}(o_t|s_t)\Big].\nonumber
\end{align}
By comparing the VFE in \eqref{eq:def_vfe_fountas} and the EFE in \eqref{eq:efe_fountas_defin}, one can see an inconsistency. Namely, the parameters are seen as latent variables in the EFE definition, c.f., $\theta$ in \eqref{eq:efe_fountas_defin}, but they are regarded as parameters of neural networks in the VFE, c.f., $\theta_s$ and $\theta_o$ in \ref{eq:def_vfe_fountas}. Note, $\theta$ cannot be both a parameter (i.e, parameter, vector, or matrix) and a random variable, and if $\theta$ is a random variable, one must define its probability density, i.e., $P(\theta)$. Additionally, this inconsistency raises the question of whether the EFE is really the expectation of the VFE. To sum up, the $DAI_{MC}$ agent is equipped with four deep neural networks modelling $Q_{\phi_a}(a_t|s_t)$, $Q_{\phi_s}(s_t)$, $P_{\theta_s}(s_t|s_{t-1},a_{t-1})$, and $P_{\theta_o}(o_t|s_t)$. The weights of those networks are optimised using back-propagation to minimise the VFE given by \eqref{eq:def_vfe_fountas}. Note, \eqref{eq:def_vfe_fountas} decomposes into two KL-divergence terms that can be computed analytically, and the expectations can be approximated using a Monte-Carlo estimate. Also, because $P_{\theta_o}(o_t|s_t)$ is modelled as a product of Bernoulli distributions, the logarithm of $P_{\theta_o}(o_t|s_t)$ reduces to the binary cross entropy.

\subsubsection{Independence assumptions and the expected free energy} \label{ssection:efe_derivation_assumptions}

The EFE as stated in Equation \eqref{eq:efe_fountas_defin} needs to be re-arranged because it cannot be easily evaluated. We therefore present the derivation proposed by \citet{DeepAIwithMCMC}. Then, we highlight two independence assumptions, i.e., $s_\tau \indep \theta \,\,|\,\, \pi$ and $s_\tau \indep \theta \,\,| \,\,\pi, o_\tau$, used without explicitly presented proofs. Finally, we propose an alternative derivation that does not require these two assumptions and produces a simpler result. Using the product rule of probability, one can see that $Q(s_\tau, \theta|\pi) = Q(\theta|s_\tau,\pi)Q(s_\tau|\pi)$ and $\tilde{P}(o_\tau,s_\tau,\theta|\pi) = P(o_\tau|\pi)P(s_\tau|o_\tau,\pi)P(\theta|s_\tau,o_\tau,\pi)$. Using those two factorisations, the EFE given in \eqref{eq:efe_fountas_defin}, i.e.,
\begin{align*}
G_{\tau}(\pi) = \mathbb{E}_{\tilde{Q}}\Big[ \ln Q(s_\tau, \theta|\pi) - \ln \tilde{P}(o_\tau, s_\tau,\theta|\pi) \Big],
\end{align*}
where $\tilde{Q} = Q(o_\tau, s_\tau, \theta|\pi)$, and can be re-arranged as follows:
\begin{align}
G_{\tau}(\pi) = &- \mathbb{E}_{\tilde{Q}}\Big[ \ln \tilde{P}(o_\tau|\pi)\Big]\nonumber\\
&+ \mathbb{E}_{\tilde{Q}}\Big[ \ln Q(s_\tau|\pi) - \ln \tilde{P}(s_\tau|o_\tau, \pi) \Big]\nonumber\\
&+ \mathbb{E}_{\tilde{Q}}\Big[ \ln Q(\theta|s_\tau, \pi) - \ln \tilde{P}(\theta|s_\tau,o_\tau,\pi) \Big].\label{eq:efe_rearrangedd}
\end{align}
Note, the above derivation follows the work of \citet{DeepAIwithMCMC}.

\subsubsubsection{Re-arranging the second term of Equation \eqref{eq:efe_rearrangedd} according to \citet{DeepAIwithMCMC}}
First, the second term of Equation \eqref{eq:efe_rearrangedd} is re-arranged into entropy terms for which an analytical solution exists. In the supplementary material of \citep{DeepAIwithMCMC}, the derivation proceeds as follows:
\begin{align}
\mathbb{E}_{\tilde{Q}}\Big[ \ln Q(s_\tau|\pi) - \ln \tilde{P}(s_\tau|o_\tau, \pi) \Big] &\delequal \mathbb{E}_{\tilde{Q}}\Big[ \ln Q(s_\tau|\pi) - \ln Q(s_\tau|o_\tau, \pi) \Big]\label{eq:switching_point}\\
&= \mathbb{E}_{Q(\theta|\pi)Q(s_\tau|\theta,\pi)Q(o_\tau|s_\tau,\theta,\pi)}\Big[ \ln Q(s_\tau|\pi) - \ln Q(s_\tau|o_\tau, \pi) \Big]\nonumber\\
&= \mathbb{E}_{Q(\theta|\pi)}\Big[ \mathbb{E}_{Q(s_\tau|\theta,\pi)}[\ln Q(s_\tau|\pi)] - \mathbb{E}_{Q(s_\tau|\theta,\pi)Q(o_\tau|s_\tau,\theta,\pi)}[\ln Q(s_\tau|o_\tau, \pi)] \Big]\nonumber\\
&= \mathbb{E}_{Q(\theta|\pi)}\Big[ \mathbb{E}_{Q(s_\tau|\theta,\pi)}[\ln Q(s_\tau|\pi)] - \mathbb{E}_{Q(o_\tau|\theta,\pi)Q(s_\tau,|o_\tau,\theta,\pi)}[\ln Q(s_\tau|o_\tau, \pi)] \Big],\nonumber
\end{align}
where in the first line a distribution was renamed, i.e., $\tilde{P}(s_\tau|o_\tau, \pi) \delequal Q(s_\tau|o_\tau, \pi)$. The next step in the derivation (c.f. supplementals of \citet{DeepAIwithMCMC}) re-arranges this final expression to the following:
\begin{align*}
\mathbb{E}_{\tilde{Q}}\Big[ \ln Q(s_\tau|\pi) - \ln Q(s_\tau|o_\tau, \pi) \Big] &= \mathbb{E}_{Q(\theta|\pi)}\Big[ \mathbb{E}_{Q(o_\tau|\theta,\pi)}[H[Q(s_\tau|o_\tau, \pi)]] - H[Q(s_\tau|\pi)] \Big].
\end{align*}
However, the above equation assumes that $s_\tau \indep \theta\,\, | \,\,\pi$ and $s_\tau \indep \theta \,\,| \,\,\pi, o_\tau$. In other words, some of the conditioning on $\theta$ has been dropped, i.e.,
\begin{align*}
\mathbb{E}_{Q(\theta|\pi)}\Big[ &\mathbb{E}_{Q(s_\tau|\theta,\pi)}[\ln Q(s_\tau|\pi)] - \mathbb{E}_{Q(o_\tau|\theta,\pi)Q(s_\tau,|o_\tau,\theta,\pi)}[\ln Q(s_\tau|o_\tau, \pi)] \Big] \tag{last expression of derivation \ref{eq:switching_point}}\\
&\neq \mathbb{E}_{Q(\theta|\pi)}\Big[ \mathbb{E}_{Q(s_\tau|\pi)}[\ln Q(s_\tau|\pi)] - \mathbb{E}_{Q(o_\tau|\theta,\pi)Q(s_\tau|o_\tau,\pi)}[\ln Q(s_\tau|o_\tau, \pi)] \Big] \\
&= \mathbb{E}_{Q(\theta|\pi)}\Big[ \mathbb{E}_{Q(o_\tau|\theta,\pi)}\big[H[Q(s_\tau|o_\tau, \pi)]\big] - H[Q(s_\tau|\pi)] \Big].
\end{align*}
Whether this conditioning can be dropped or not depends on the factorisation of the distribution. In other words, the two assumptions (i.e., $s_\tau \indep \theta \,\,|\,\, \pi$ and $s_\tau \indep \theta \,\,| \,\,\pi, o_\tau$) would have to be checked using the d-separation criterion. However, this is difficult to do, since as mentioned previously, the parameters $\theta$ are latent variables in \eqref{eq:efe_fountas_defin} but are regarded as parameters in \eqref{eq:def_vfe_fountas}, which makes the graphical model unclear. Instead of attempting to prove that $s_\tau \indep \theta \,\,|\,\, \pi$ and $s_\tau \indep \theta \,\,| \,\,\pi, o_\tau$, we propose an alternative derivation that does not require such independence assumptions. 

\subsubsubsection{Alternative derivation of the second term of Equation  \eqref{eq:efe_rearrangedd}}

Restarting from the second term of Equation \eqref{eq:efe_rearrangedd}, we can  re-arrange as follows:
\begin{align*}
\mathbb{E}_{\tilde{Q}}\Big[ \ln Q(s_\tau|\pi) - \ln \tilde{P}(s_\tau|o_\tau, \pi) \Big] &\delequal \mathbb{E}_{\tilde{Q}}\Big[ \ln Q(s_\tau|\pi) - \ln Q(s_\tau|o_\tau, \pi) \Big]\\
&= \mathbb{E}_{Q(s_\tau|\pi)Q(\theta,o_\tau|s_\tau,\pi)}\Big[ \ln Q(s_\tau|\pi)\Big] - \mathbb{E}_{Q(o_\tau|\pi)Q(s_\tau|o_\tau,\pi)Q(\theta|s_\tau,o_\tau,\pi)}\Big[ \ln Q(s_\tau|o_\tau, \pi) \Big]\\
&= \mathbb{E}_{Q(s_\tau|\pi)}\Big[ \ln Q(s_\tau|\pi)\Big] - \mathbb{E}_{Q(o_\tau|\pi)Q(s_\tau|o_\tau,\pi)}\Big[ \ln Q(s_\tau|o_\tau, \pi) \Big]\\
&= \mathbb{E}_{Q(o_\tau|\pi)}\Big[ H[Q(s_\tau|o_\tau, \pi)] \Big] - H[Q(s_\tau|\pi)],
\end{align*}
where in the first line a distribution was renamed, i.e., $\tilde{P}(s_\tau|o_\tau, \pi) \delequal Q(s_\tau|o_\tau, \pi)$, two different factorizations of 
$\tilde{Q}$ are used going from the first to the second line, the linearity of expectations was used between the first and second line, and the expectation 
w.r.t $\theta$ was dropped (between lines two and three) because the expectation of a constant is the constant itself. Importantly, the above derivation 
does not make any assumption of independence, and leads to a simpler result. This alternative line of reasoning is beneficial as it produces a stronger 
derivation that relies upon fewer assumptions. The simpler result produced by this derivation, also have a practical implication. Indeed, the expectation 
w.r.t. $Q(\theta|\pi)$ disappears and the expectation w.r.t. $Q(o_\tau|\theta,\pi)$ is now w.r.t. $Q(o_\tau|\pi)$. Those two changes sugguest that a 
different implementation of this term is required.

\subsubsubsection{Re-arranging the third term of Equation \eqref{eq:efe_rearrangedd} from \citet{DeepAIwithMCMC}}

For completeness, we now focus on the third term of \eqref{eq:efe_rearrangedd}, which can be re-arranged as follows:
\begin{align*}
\mathbb{E}_{\tilde{Q}}\Big[ \ln Q(\theta|s_\tau, \pi) - \ln \tilde{P}(\theta|s_\tau,o_\tau,\pi) \Big] &\delequal \mathbb{E}_{\tilde{Q}}\Big[ \ln Q(\theta|s_\tau, \pi) - \ln Q(\theta|s_\tau,o_\tau,\pi) \Big] \\
&= \mathbb{E}_{\tilde{Q}}\Big[ \ln Q(o_\tau|s_\tau, \pi) - \ln Q(o_\tau|s_\tau,\theta,\pi) \Big],
\end{align*}
where $\tilde{P}(\theta|s_\tau,o_\tau,\pi)$ was renamed as $Q(\theta|s_\tau,o_\tau,\pi)$, and Bayes theorem was used to get:
\begin{align*}
Q(\theta|s_\tau, \pi)
= \frac{Q(\theta|s_\tau,o_\tau,\pi)Q(o_\tau|s_\tau, \pi)}{Q(o_\tau|s_\tau,\theta,\pi)} \Leftrightarrow \frac{Q(\theta|s_\tau, \pi)}{Q(\theta|s_\tau,o_\tau,\pi)}
= \frac{Q(o_\tau|s_\tau, \pi)}{Q(o_\tau|s_\tau,\theta,\pi)}.
\end{align*}
Finally, by recalling that $\tilde{Q} = Q(o_\tau, s_\tau, \theta|\pi) = Q(\theta|\pi)Q(s_\tau|\theta,\pi)Q(o_\tau|s_\tau,\theta,\pi)$, and using the linearity of expectation, we get:
\begin{align*}
\mathbb{E}_{\tilde{Q}}\Big[ \ln Q(\theta|s_\tau, \pi) - \ln Q(\theta|s_\tau,o_\tau,\pi) \Big] &= \mathbb{E}_{Q(o_\tau,s_\tau|\pi)}\Big[ \ln Q(o_\tau|s_\tau, \pi)\Big] + \mathbb{E}_{Q(\theta|\pi)Q(s_\tau|\theta,\pi)}\Big[H[Q(o_\tau|s_\tau,\theta,\pi)] \Big]\\
&= \mathbb{E}_{Q(o_\tau|s_\tau,\pi)Q(s_\tau|\pi)}\Big[ \ln Q(o_\tau|s_\tau, \pi)\Big] + \mathbb{E}_{Q(\theta|\pi)Q(s_\tau|\theta,\pi)}\Big[H[Q(o_\tau|s_\tau,\theta,\pi)] \Big]\\
&= - \mathbb{E}_{Q(s_\tau|\pi)}\Big[ H\big[ Q(o_\tau|s_\tau, \pi) \big] \Big] + \mathbb{E}_{Q(\theta|\pi)Q(s_\tau|\theta,\pi)}\Big[H[Q(o_\tau|s_\tau,\theta,\pi)] \Big],
\end{align*}
where because $\ln Q(o_\tau|s_\tau, \pi)$ is a constant w.r.t $\theta$, we have been able to use:
\begin{align*}
\mathbb{E}_{Q(o_\tau,\theta,s_\tau|\pi)}[ \ln Q(o_\tau|s_\tau, \pi)] = \mathbb{E}_{Q(o_\tau,s_\tau|\pi)}[ \ln Q(o_\tau|s_\tau, \pi)].
\end{align*}
To sum up, this derivation provides an expression based on the following two entropy terms: $H[ Q(o_\tau|s_\tau, \pi) ]$ and $H[Q(o_\tau|s_\tau,\theta,\pi)]$, 
which the authors claim can be estimated (c.f. Appendix B for more details). Note that our proposed alternative to the EFE (see below) uses 
this derivation for the third term of Equation \eqref{eq:efe_rearrangedd}.

\subsubsubsection{The EFE from \citet{DeepAIwithMCMC}}

If one follows the derivation proposed by \citet{DeepAIwithMCMC}, then the EFE is given by:
\begin{align}
G_{\tau}(\pi) = &- \mathbb{E}_{\tilde{Q}}\Big[ \ln \tilde{P}(o_\tau|\pi)\Big]\nonumber\\
&+ \mathbb{E}_{Q(\theta|\pi)}\Big[ \mathbb{E}_{Q(o_\tau|\theta,\pi)}\big[H[Q(s_\tau|o_\tau, \pi)]\big] - H[Q(s_\tau|\pi)] \Big]\nonumber\\
&+ \mathbb{E}_{Q(\theta|\pi)Q(s_\tau|\theta,\pi)}\Big[H[Q(o_\tau|s_\tau,\theta,\pi)] \Big] - \mathbb{E}_{Q(s_\tau|\pi)}\Big[ H\big[ Q(o_\tau|s_\tau, \pi) \big] \Big].\label{eq:efe_rearranged_fountas}
\end{align}

\noindent Note, in Section \ref{ssection:fountas_paper}, we focused on presenting the approach given in \citet{DeepAIwithMCMC}, with some adjustments for consistency. More details about the implementation of Equation \ref{eq:efe_rearranged_fountas} are presented in Appendix B, along with some discrepancies between the paper and the code.

\subsubsubsection{Our proposed alternative to the EFE}

If one follows our alternative derivation, then the EFE is given by:
\begin{align*}
G_{\tau}(\pi) = &- \mathbb{E}_{\tilde{Q}}\Big[ \ln \tilde{P}(o_\tau|\pi)\Big]\\
&+ \mathbb{E}_{Q(o_\tau|\pi)}\Big[ H[Q(s_\tau|o_\tau, \pi)] \Big] - H[Q(s_\tau|\pi)],\\
&+ \mathbb{E}_{Q(\theta|\pi)Q(s_\tau|\theta,\pi)}\Big[H[Q(o_\tau|s_\tau,\theta,\pi)] \Big] - \mathbb{E}_{Q(s_\tau|\pi)}\Big[ H\big[ Q(o_\tau|s_\tau, \pi) \big] \Big].
\end{align*}

\noindent Finally, $DAI_{MC}$ does solve the dSprites environment.

\subsection{$DAI_{VPG}$ agent \citep{DeepAI}}

\begin{figure}[H]
	\begin{center}
	\begin{tikzpicture}[square/.style={regular polygon,regular polygon sides=4}]
		\coordinate (A) at (13.5,12.5);
		\coordinate (B) at (16,13.5);
		\coordinate (C) at (16,15.5);
		\coordinate (D) at (13.5,16.5);
		\draw[fill=purple!20!white] (A) -- coordinate[pos=.6] (AB) (B)--(C)
         --coordinate[pos=.45] (CD) (D)
         --coordinate[pos=.55] (DA) cycle;
	    \node at (14.75, 14.5) {Policy};
		\draw[-latex] (12.8,6.5) -- (12.8,14.5) -- (13.5,14.5);
		\draw[-latex] (16, 14.5) -- (16.5, 14.5);
		\draw[fill=gray!20!white] (16.5,13.5) rectangle (17.5,15.5);
	    \node at (17, 14.5) {$\hat{\pi}$};

		\coordinate (A) at (13.5,8);
		\coordinate (B) at (16,9);
		\coordinate (C) at (16,11);
		\coordinate (D) at (13.5,12);
		\draw[fill=yellow!20!white] (A) -- coordinate[pos=.6] (AB) (B)--(C)
         --coordinate[pos=.45] (CD) (D)
         --coordinate[pos=.55] (DA) cycle;
	    \node at (14.75, 10) {Critic};

		\draw[-latex] (12.8,6.5) -- (12.8,10) -- (13.5,10);
		\draw[-latex] (16, 10) -- (16.5, 10);
		
		\draw[fill=gray!20!white] (16.5,9) rectangle (17.5,11);
	    \node at (17, 10) {$\bm{G}$};

		\draw[fill=gray!20!white] (11,9) rectangle (12,11);
	    \node at (11.5, 10) {$o_t$};

		\draw[fill=blue!20!white] (11,3.5) rectangle (12,5.5);
	    \node at (11.5, 4.5) {$a_t$};

		\coordinate (A) at (13.5,3.5);
		\coordinate (B) at (16,4.5);
		\coordinate (C) at (16,6.5);
		\coordinate (D) at (13.5,7.5);
		\draw[fill=orange!20!white] (A) -- coordinate[pos=.6] (AB) (B)--(C)
         --coordinate[pos=.45] (CD) (D)
         --coordinate[pos=.55] (DA) cycle;
	    \node at (14.75, 5.5) {Transition};

		\draw[fill=gray!20!white] (16.5,4.5) rectangle (17.5,6.5);
	    \node at (17, 5.5) {$\mathring{o}_{t+1}$};

		\draw[-latex] (12,4.5) -- (13.5,4.5);
		\draw[-latex] (12,10) -- (13.5,10);
		\draw[-latex] (12.8,6.5) -- (13.5,6.5);

		\draw[-latex] (16,5.5) -- (16.5,5.5);

    \end{tikzpicture}
	\end{center}
  \caption{This figure illustrates the $DAI_{VPG}$ agent. The only new part is the policy network, which takes as input the hidden state at time $t$ and ouputs the parameters $\hat{\pi}$ of the variational posterior over actions. Importantly, the $DAI_{VPG}$ takes actions based on the EFE.}
   \label{fig:DAI_VPG_agent}
\end{figure}
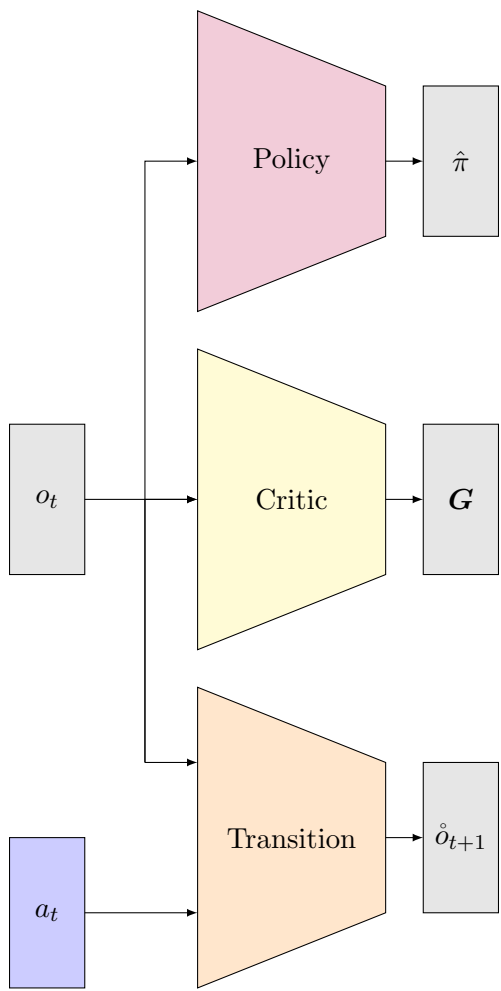

In this section, we explain and discuss the approach of \citet{DeepAI}. The code is available at the following URL: \url{https://github.com/BerenMillidge/DeepActiveInference}. Note that, though the mathematics in the paper are based on the formalism of a partially observable Markov decision process (POMDP), the code does not implement an encoder/decoder architecture, which means that the code implements a fully observable decision process, i.e., MDP. Additionally, the $DAI_{VPG}$ is composed of three neural networks, as illustrated in Figures \ref{fig:beren_dnn} and \ref{fig:DAI_VPG_agent}. The first is the transition network that predicts the future observations based on the current observations and action, i.e., $\mathring{o}_{\tau+1} = \mathcal{T}_{\theta_o}(o_\tau, a_\tau)$. The second is the policy network that models the variational distribution over actions $Q_{\phi_a}(a_\tau|o_\tau)$. The third is the critic network that predicts the expected free energy of each action given the current observation. Moreover, \citet{DeepAI} defines the prior over actions as follows:
\begin{align*}
P(a_\tau|o_\tau) = \sigma[-\zeta G(o_\tau, a_\tau)],
\end{align*}
where $\zeta$ is the precision of the prior over actions, $\sigma[\,\bigcdot\,]$ is a softmax function, and $G(o_\tau, a_\tau)$ is the expected free energy (EFE) of taking action $a_\tau$ when observing $o_\tau$. In the paper, the mathematics are based on the POMDP formalism. Therefore, $G(o_\tau, a_\tau)$ is denoted $G(s_\tau, a_\tau)$, and is defined as follows:
\begin{align}
G(s_\tau, a_\tau) = - r_\tau + \underbrace{\kl{Q(s_\tau)}{Q(s_\tau|o_\tau)}}_{\text{intrinsic value}} + \, \hat{\mathcal{G}}_{\hat{\theta}_a}(a_{\tau+1},s_{\tau+1}),\label{eq:efe_beren}
\end{align}
where $r_\tau$ is the reward gathered by the agent at time step $\tau$, and $\hat{\mathcal{G}}_{\hat{\theta}_a}(a_{\tau+1},s_{\tau+1})$ is the target network (i.e., a copy of the critic network whose weights are synchronised every $K$ iterations of learning). Now, remember that in the implementation, there is no encoder $Q(s_\tau)$ and no decoder $P(o_\tau|s_\tau)$. In other words, there are no hidden states $s_\tau$, raising some uncertainty about how the intrinsic value is computed. The code available on Github\footnote{We are referring to the version of the code that was available on github on the 6th of June 2022.} at the following URL: \url{https://github.com/BerenMillidge/DeepActiveInference}, in the file \url{active_inference_with_Tmodel.jl} (see line 51) suggests that the following equation is used:
\begin{align}
\text{intrinsic value} = \sum_{i} \Big[ o_{\tau+1}[i] - \mathring{o}_{\tau+1}[i] \Big]^2, \label{eq:intrisic_values}
\end{align}
where $o_{\tau+1}[i]$ is the i-th observation received at time step $\tau + 1$, and $\mathring{o}_{\tau+1}[i]$ is the (numerical) value of the i-th observation (at time step $\tau + 1$) predicted by the transition network. More formally, the above formulation for the intrisic value corresponds to the KL-divergence between two Gaussian distributions both having an identity covariance matrix, i.e.,
\begin{align*}
\text{intrinsic value} = \kl{Q(o_{\tau+1})}{P(o_{\tau+1}|o_\tau, a_\tau)} = \sum_{i} \Big[ o_{\tau+1}[i] - \mathring{o}_{\tau+1}[i] \Big]^2,
\end{align*}
where $P(o_{\tau+1}|o_\tau, a_\tau) = \mathcal{N}(o_{\tau+1};\mathring{o}_{\tau+1},I)$ and $Q(o_{\tau+1})$ is a Gaussian distribution with mean vector $o_{\tau+1}$ and an identity covariance matrix. However, note that \eqref{eq:efe_beren} is the definition of the expected free energy in the POMDP setting. As explained by \citet{dacosta2020relationship}, the expected free energy in the MDP setting is given by:
\begin{align*}
G(a_{t:T-1},o_t) \approx \sum_{\tau = t+1}^T \kl{P(o_\tau|a_{t:T-1}, o_t)}{P(o_\tau)},
\end{align*}
where $P(o_\tau)$ are the prior preferences of the agent (related to rewards in reinforcement learning), and $P(o_\tau |a_{t:T-1}, o_t)$ is the transition mapping. Importantly, this definition for the expected free energy does not decompose into extrinsic and intrinsic terms as in \eqref{eq:efe_beren}. Thus, (as it stands) the implementation of the $DAI_{VPG}$ agent is a mixture between the POMDP and MDP setting, where the generative model corresponds to an MDP, and the expected free energy is adapted from the POMDP setting.

We conclude this section by dicussing the training procedure of the transition, policy and critic networks. As explained in the paper, the transition network is trained to minimise the variational free enery. Additionally, because of the Gaussian assumptions (with identity covariance matrices) mentioned above, the KL-divergence reduces to the mean square error (MSE). Thus, the transition network is updated to minimise the MSE between the observations made by the agent at time $\tau + 1$, and the observations ($\mathring{o}_{\tau+1}$) predicted by the transition network, i.e., 
\begin{align*}
\theta_o^* = \argmin_{\theta_o} \text{MSE}\Big[o_{\tau+1}, \mathring{o}_{\tau+1}\Big],
\end{align*}
where $\mathring{o}_{\tau+1} = \mathcal{T}_{\theta_o}(o_\tau, a_\tau)$. The policy network is trained to minimise the KL-divergence between the variational posterior over actions $Q_{\phi_a}(a_\tau|o_\tau)$ and the prior over actions $P(a_\tau|o_\tau)$, i.e., 
\begin{align*}
\phi_a^* = \argmin_{\phi_a} \kl{Q_{\phi_a}(a_\tau|o_\tau)}{P(a_\tau|o_\tau)},
\end{align*}
which minimises the variational free energy. Finally, the critic is trained by minimising the MSE between the target EFE as defined in \eqref{eq:efe_beren} and the ouput of the critic $\mathcal{G}_{\theta_a}(o_\tau, \bigcdot\,)$:
\begin{align*}
\theta_a^* = \argmin_{\theta_a} \text{MSE}\Big[G(o_\tau, \bigcdot\,), \mathcal{G}_{\theta_a}(o_\tau, \bigcdot\,)\Big].
\end{align*}
$DAI_{VPG}$ is able to solve the CartPole environment. 

\begin{figure}[H]
	\begin{center}
	\resizebox{1\textwidth}{!}{%
	\begin{tikzpicture}
		\node[text width=4cm, align=center] at (-5.7,5.5) {There is no encoder $\bm{Q(s_\tau)}$, and no decoder $\bm{P(o_\tau|s_\tau)}$.};
		\draw (-7.7,4.5) rectangle (-3.7,6.5);
		
		\node at (0,8) {$\mathcal{T}_{\theta_o}(o_\tau, a_\tau)$};
		\node at (0,7) {\textbf{input:} \#O + \#A};
		\pic{dense=0/5.75/100/ReLU};
		\pic{dense=0/4.25/\#O/};
		\draw[-latex] (0,5.5) -- (0,5);

		\node at (5,8) {$Q_{\phi_a}(a_\tau| o_\tau)$};
		\node at (5,7) {\textbf{input:} \#O};
		\pic{dense=5/5.75/100/ReLU};
		\pic{dense=5/4.25/\#A/};
		\draw[-latex] (5,5.5) -- (5,5);

		\node at (10,8) {$\mathcal{G}_{\theta_a}(o_\tau, \bigcdot\,)$};
		\node at (10,7) {\textbf{input:} \#O};
		\pic{dense=10/5.75/100/ReLU};
		\pic{dense=10/4.25/\#A/};
		\draw[-latex] (10,5.5) -- (10,5);
	\end{tikzpicture}
	  }%
	\end{center}
	\caption{Neural networks architecture of the $DAI_{VPG}$ agent. Green blocks correspond to fully connected layers. The first neural network is the transition network that takes as input the observation and action at time step $\tau$, and outputs the mean of a Gaussian distribution over observation at time step $\tau + 1$. The second neural network is the policy network that models the variational posterior over actions. The third neural network is the critic that takes as input an observation and outputs the expected free energy of each action.}
	\label{fig:beren_dnn}
\end{figure}
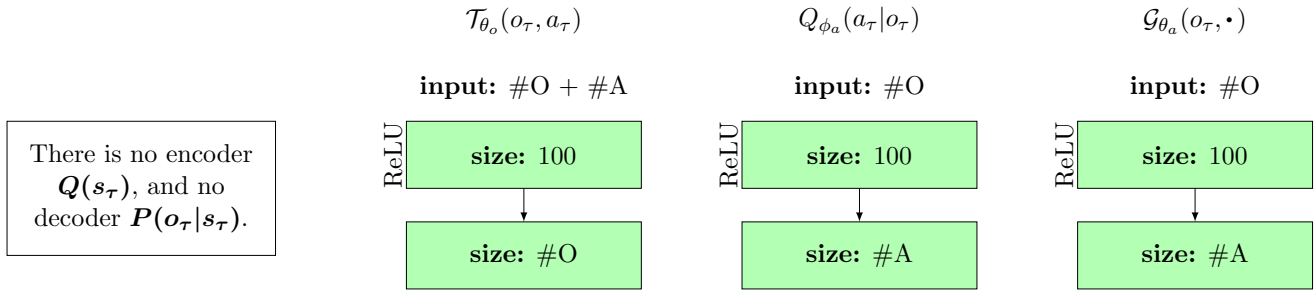

\subsection{$DAI_{RHI}$ agent \citep{rood2020deep}}

In this section, we explain and discuss the approach of \citet{rood2020deep}. Put simply, this paper proposes a variational auto-encoder (VAE), which is able to account for results that were observed in the context of the rubber-hand illusion (RHI) experiment. In the experiment in \citet{rood2020deep}, an agent (i.e., either a human or a computer) is able to move an arm in a 3D space. However, the agent does not observe the real position of the arm, instead, the agent sees an artificial hand placed in a different location. This can be implemented using virtual reality (for humans) or within a simulator (for computers). Since, \citet{rood2020deep} restricted themself to the context of a VAE, this approach cannot be considered as a complete implementation of deep active inference. More precisely, the transition and critic (or policy) networks are missing.

\subsection{$DAI_{HR}$ agent \citep{sancaktar2020endtoend,DAI_HR,DAI_HR2}}

In this section, we explain and discuss the following approaches: \citet{sancaktar2020endtoend}, \citet{DAI_HR}, and \citet{DAI_HR2}. Briefly, those papers propose a free energy minimisation scheme based on a single decoder network, which is used to control Nao, TIAGo, and iCub robots, respectively. Since, \citet{sancaktar2020endtoend,DAI_HR,DAI_HR2} restricted themself to the context of a single decoder, this approach can not be considered as a complete implementation of deep active inference. More precisely, the encoder, transition and critic (or policy) networks are missing.

\subsection{$DAI_{FA}$ agent \citep{DAI_Kai}} \label{ssec:dai_approach}

\begin{figure}[H]
	\begin{center}
    \resizebox{1\textwidth}{!}{%
    \begin{tikzpicture}
		\node at (-5,8) {$Q_{\phi_s}(s_t|o_t, s_{t-1})$};
		\node at (-5,7) {\textbf{input:} \#O + \#S};
		\pic{dense=-5/5.75/\#S/ReLU};
		\pic{dense=-5/4.25/\#S/ReLU};
		\pic{dense=-5/2.75/\#S + \#S/THS};
		\draw[-latex] (-5,5.5) -- (-5,5);
		\draw[-latex] (-5,4) -- (-5,3.5);

		\node at (0,8) {$P_{\theta_o}(o_\tau|s_\tau)$};
		\node at (0,7) {\textbf{input:} \#S};
		\pic{dense=0/5.75/\#S/ReLU};
		\pic{dense=0/4.25/\#S/ReLU};
		\pic{dense=0/2.75/\#S/ReLU};
		\pic{dense=0/1.25/\#O + \#O/LS};
		\draw[-latex] (0,5.5) -- (0,5);
		\draw[-latex] (0,4) -- (0,3.5);
		\draw[-latex] (0,2.5) -- (0,2);

		\node at (5,8) {$P_{\theta_s}(s_\tau|s_{\tau-1})$};
		\node at (5,7) {\textbf{input:} \#S};
		\pic{dense=5/5.75/\#S + \#S/THS};

		\node at (10,8) {$P_{\theta_a}(a_\tau| s_\tau)$};
		\node at (10,7) {\textbf{input:} \#S};
		\pic{dense=10/5.75/\#S/};
		\pic{dense=10/4.25/\#A + \#A/LS};
		\draw[-latex] (10,5.5) -- (10,5);
	\end{tikzpicture}
	}%
	\end{center}
	\caption{Neural networks architecture of the $DAI_{FA}$ agent. Green blocks correspond to fully connected layers. The first neural network is the encoder that takes as input the state at time $t-1$ and the observation at time $t$, and outputs the parameters of a distribution over the state at time $t$. The second neural network is the decoder that takes as input the state at time $\tau$, and outputs the parameters of a distribution over the observation at time $\tau$. The third is the transition network that takes as input the state at time step $\tau - 1$, and outputs the parameters of a distribution over the state at time step $\tau$. The fourth neural network is the policy network that models the prior over actions, i.e., the policy takes as input a state at time $\tau$ and outputs the parameters of a distribution over the actions at time step $\tau$. Finally, THS stands for tangent hyperbolic and softplus, i.e., the tangent hyperbolic activation is over the first half of the neurons and the softplus activation function is over the second half, and LS stands for linear activation function and softplus, i.e., the linear activation is over the first half of the neurons and the softplus activation function is over the second half.}
	\label{fig:kai_dnn}
\end{figure}
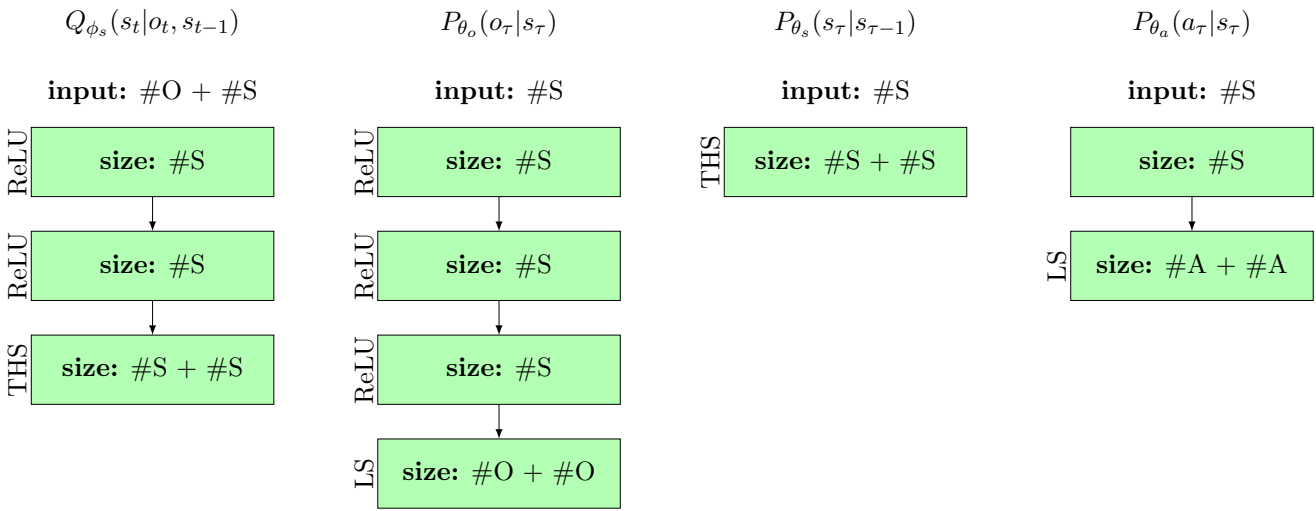

\begin{figure}[H]
	\begin{center}
	\begin{tikzpicture}
		\node at (-5,8.1) {$s_\tau$};
		\node at (-4.3,7.3) {$\mathcal{P}_{\theta_a}(s_\tau)$};
		\draw[-latex] (-5,7.8) -- (-5,7) -- (-5.5,7) -- (-5.5,6.5);
		\draw[-latex] (-5,7.8) -- (-5,7) -- (-4.5,7) -- (-4.5,6.5);
		\node at (-4.5,6.2) {$\sigma_a$};
		\node at (-5.5,6.2) {$\mu_a$};
		\node[latent] at (-6.5,6.2) (epsilon) {$\epsilon$};
		\draw (-6.5,5.85) -- (-6.5,5.35) -- (-4.5,5.35) -- (-4.5,5.85);
		\draw[-latex] (-5.5,5.85) -- (-5.5,4.85);
		\node at (-5.5,4.5) {$\hat{a}_\tau$};
		\draw[-latex] (-5.5,4.15) -- (-5.5,3.5);
		\node at (-4,3.925) {env.execute($\hat{a}_\tau$)};
		\node at (-5.5,3.1) {$o_{\tau+1}^x$};
		\draw (-5,3.1) -- (-1.5,3.1);
		\draw[-latex] (-5.2,4.5) -- (1,4.5);
		\draw (-4.7,8.1) -- (-1.5,8.1) -- (-1.5,3.1);
		\draw[-latex] (0.6,4.5) -- (0.6,3.7) -- (1,3.7);
		\node at (-0.3,4.8) {$\mathcal{E}_{\phi_s}(o_{\tau+1}, s_\tau)$};
		\node[latent] at (1.3,5.5) (epsilon) {$\epsilon$};
		\node at (1.3,4.5) {$\hat{\mu}$};
		\node at (1.3,3.7) {$\hat{\sigma}$};
		\draw[-latex] (1.6,4.5) -- (2.6,4.5);
		\draw (1.6,3.7) -- (2.1,3.7) -- (2.1,5.5) -- (1.65,5.5);
		\node at (3.1,4.5) {$\hat{s}_{\tau+1}$};
		\draw[-latex] (3.1,4.8) -- (3.1,8.8) -- (-5,8.8) -- (-5,8.4);

		\draw[-latex, red!70!black] (3.1,4.2) -- (3.1,3.3) -- (3.8,3.3) -- (3.8,2.9);
		\draw[-latex, red!70!black] (3.1,4.2) -- (3.1,2.9);
		\node[red!70!black] at (3.15,2.6) {$\mu_o$};
		\node[red!70!black] at (3.85,2.6) {$\sigma_o$};
		\node[red!70!black] at (4,3.6) {$\mathcal{D}_{\theta_o}(\hat{s}_{\tau+1})$};
		\draw[-latex, red!70!black] (1.6,4.4) -- (2,4.4) -- (2,2.4);
		\draw[red!70!black] (1.6,3.6) -- (2,3.6);
		\node[red!70!black] at (2,2) {$\text{\textbf{VFE}}_\tau$};
		\draw[-latex, red!70!black] (3.1,2.3) -- (3.1,2) -- (2.7,2);
		\draw[-latex, red!70!black] (3.85,2.3) -- (3.85,2) -- (2.7,2);
		\draw[-latex, red!70!black] (-5.4,8.1) -- (-7.3,8.1) -- (-7.3,2) -- (-2,2);
		\node[red!70!black] at (-2.7,2.3) {$\mathcal{T}_{\theta_s}(s_\tau$)};
		\draw[-latex, red!70!black] (-2.4,2) -- (-2.4,1) -- (-2,1);
		\node[red!70!black] at (-1.7,2) {$\mathring{\mu}$};
		\node[red!70!black] at (-1.7,1) {$\mathring{\sigma}$};

		\draw[-latex, red!70!black] (-1.4,2) -- (1.25,2);
		\draw[red!70!black] (-1.4,1) -- (-1,1) -- (-1,2);

	\end{tikzpicture}
	\end{center}
	\caption{Action-perception cycles (in black) and estimation of the free action objective (in red). Note, $\mathring{\mu}$, $\mathring{\sigma}$, $\hat{\mu}$ and $\hat{\sigma}$ are used to compute the complexity terms of the variational free energy, while $\mu_o$ and $\sigma_o$ are used to compute the accuracy term of the variational free energy.}
	\label{fig:kai_fa_estimate}
\end{figure}
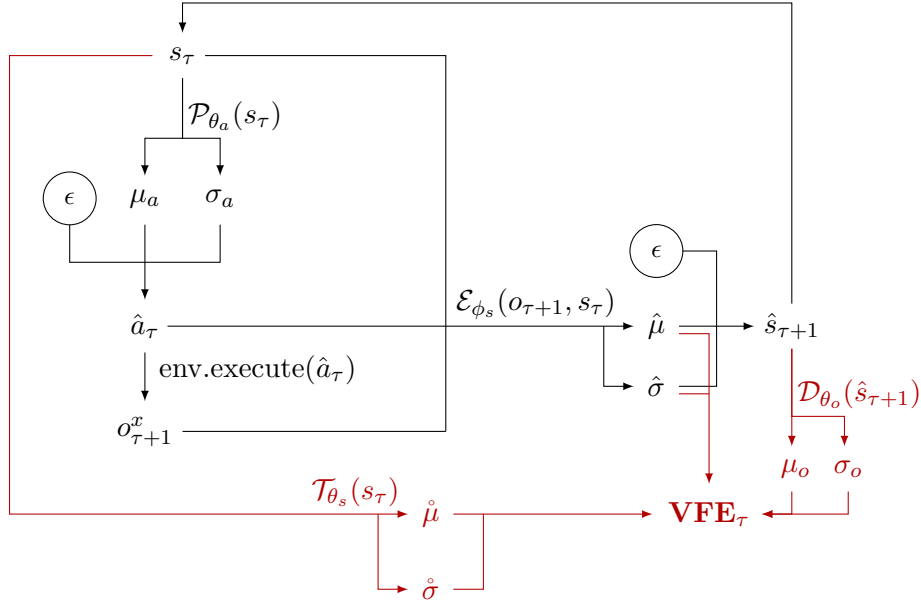

In this section, we review the approach proposed by \citet{DAI_Kai}. The original code of this paper is available on GitHub at the following URL: \url{https://github.com/kaiu85/deepAI_paper}. This approach is composed of four deep neural networks. The encoder $\mathcal{E}_{\phi_s}$ models the approximate posterior over states $Q_{\phi_s}(s_t|s_{t-1}, o_t)$ as a Gaussian distribution, i.e., $Q_{\phi_s}(s_t|s_{t-1}, o_t) = \mathcal{N}(s_t;\mu, \sigma)$ where $\mu, \sigma = \mathcal{E}_{\phi_s}(s_{t-1}, o_t)$. The decoder $\mathcal{D}_{\theta_o}$ models the likelihood mapping $P_{\theta_o}(o_\tau|s_\tau)$ as a Gaussian distribution, i.e., $P_{\theta_o}(o_\tau|s_\tau) = \mathcal{N}(o_\tau;\mu_o, \sigma_o)$ where $\mu_o, \sigma_o = \mathcal{D}_{\theta_o}(s_\tau)$. The transition network $\mathcal{T}_{\theta_s}$ models the transition mapping $P_{\theta_s}(s_\tau|s_{\tau-1})$ as a Gaussian distribution, i.e., $P_{\theta_s}(s_\tau|s_{\tau-1}) = \mathcal{N}(s_\tau;\mathring{\mu}, \mathring{\sigma})$ where $\mathring{\mu}, \mathring{\sigma} = \mathcal{T}_{\theta_s}(s_{\tau-1})$. Note that, the transition network is only conditioned on the previous state. This is because the action is contained in the observations predicted by the decoder. More precisely, the experiments were run in the MountainCar environment, which means that the agent is observing the x position of the car $o_\tau^x$. Additionally, according to the idea of proprioception, the agent observes its own action, i.e., $o^a_{\tau-1} = a_{\tau-1}$ where $a_{\tau-1}$ is the action performed by the agent at time $\tau-1$. In what follows, we let $o_\tau = (o^x_\tau, o^a_{\tau-1})$ be the concatenation of the x position of the car and the action taken by the agent. Importantly, because $o_\tau$ contains $o^a_{\tau-1}$, the latent space has to (implicitly) encode the action for the decoder to successfully predict the observations. Finally, the policy network $\mathcal{P}_{\theta_a}$ models the prior over actions $P_{\theta_a}(a_\tau|s_\tau)$ as a Gaussian distribution, i.e., $P_{\theta_a}(a_\tau|s_\tau) = \mathcal{N}(a_\tau;\mu_a, \sigma_a)$ where $\mu_a, \sigma_a = \mathcal{P}_{\theta_a}(s_\tau)$. Figure \ref{fig:kai_dnn} illustrates the architectures of those deep neural networks. Then, \citet{DAI_Kai} defines the free action objective as the cumulated variational free energy over time:
\begin{align*}
FA(o_{1:T}, \phi, \theta) = \sum_{\tau = 1}^T \Bigg[\underbrace{\overbrace{- \mathbb{E}_{Q_{\phi_s}(s_\tau|s_{\tau-1}, o_\tau)}\Big[\ln P_{\theta_o}(o_\tau|s_\tau)\Big]}^{\text{accuracy}} + \overbrace{\kl{Q_{\phi_s}(s_\tau|s_{\tau-1}, o_\tau)}{P_{\theta_s}(s_\tau|s_{\tau-1})}}^{\text{complexity}} }_{\text{VFE}_\tau}\Bigg],
\end{align*}
where $s_0 = (0, ..., 0)$ is a vector of zeros representing the initial hidden state, $T$ is the time horizon, $P_{\theta_o}(o_\tau|s_\tau)$, $Q_{\phi_s}(s_\tau|s_{\tau-1}, o_t)$, $P_{\theta_s}(s_\tau|s_{\tau-1})$ are modeled using Gaussian distributions whose parameters are predicted by the decoder, encoder and transition network, respectively. Figure \ref{fig:kai_fa_estimate} illustrates the computation of the free action objective, and the action-perception cycle of the agent. The first action-perception cycle is initiated when the intital hidden state $s_0$ is being fed into the policy network, which outputs the parameters of a Gaussian distribution over actions. Then, an action $\hat{a}_0$ is sampled from this Gaussian, and executed in the environment leading to a new observation $o_1^x$. Next, the action $\hat{a}_0$ is concatenated with $o_1^x$ to form $o_1$. The observation $o_1$ and the state $s_0$ are then fed into the encoder that outputs the parameters of a Gaussian distribution over $\hat{s}_1$. Lastly, a state is sampled from this Gaussian distribution and is used as input to the next action-perception cycle. This process continues until reaching the time horizon.

Within each action-perception cycle, the variational free energy of this time step is computed. To compute $\text{VFE}_\tau$, the state $s_\tau$  is fed into both the tansition network and the encoder. Both networks output the parameters of a Gaussian distribution over $s_{\tau+1}$. A state is sampled from the distribution predicted by the encoder, and is used as input to the decoder that outputs the parameters of a Gaussian distribution over $o_{\tau + 1}$. Finally, the parameters of the Gaussian distribution over $o_{\tau + 1}$ is used to compute the accuracy term, and the parameters of the two Gaussian distributions over $s_{\tau+1}$ are used to compute the complexity term.

We now focus on the prior preferences of the agent. Usually, prior preferences are part of the expected free energy. However, \citet{DAI_Kai} takes a different approach. Recall, the latent variable $s_\tau$ is modeled using a multivariate Gaussian. The $DAI_{FA}$ agent reserves the first dimension of the latent space to the encoding of the prior preferences. Specifically, the transition network predicts the mean vector and the diagonal of the covariance matrix (i.e., another vector) of a multivariate Gaussian over latent states. The first element in the mean vector is clamped to the target x position, and the first element of the variance vector is set to a relatively small value. This effectively propels the agent towards the target location. Additionally, the encoder predicts another set of mean and variance vectors. The first element of the mean vector predicted by the encoder is clamped to the current x position observed by the agent, and the first element of the variance vector is set to a relatively small value. Note, clamping the value of the first element of the mean and variance vectors predicted by the transition is uncontentious, i.e., this is simply how the generative model is defined. However, clamping the value of the first element of the mean and variance vectors predicted by the encoder may be debated. Specifically, the encoder is supposed to predict the variational distribution, which is an approximation of the true posterior. However, clamping the value of the first element of the mean and variance vectors predicted by the encoder is likely to push the variational posterior further from the true posterior.

There are a number of important aspects of the $DAI_{FA}$ agent. First, there is no expected free energy, instead the agent is trained to minimise the cumulated variational free energy over time. Second, this approach unrolls the partially observable Markov decision process over time. In other words, the code builds a huge computational graph containing the encoder, decoder, transition and policy networks for each action-perception cycle. Therefore, the approach is computationally intensive and can become intractable for a large time horizon. Third, the $DAI_{FA}$ requires the modeller to encode the prior preferences within the distributions predicted by the encoder and transition network. This can limit the applicability of the approach. Indeed, as previously explained, one can encode the prior preferences of the agent for the MountainCar problem within the first dimension of the latent space. 

However, manually encoding the prior preferences in the latent space has two major drawbacks. First, the model needs to be modified from one environment to the next. This is because for each environment, the prior preferences of the agent will be different. Second, for some environments, it is unclear how the prior preferences may be defined. For example, when playing PacMan, the agent needs to eat all the dots, while simultaneously avoiding the ghosts. How can this be encoded in the model's latent space? This is particularly challenging because the only observation made by the agent is an image of the game, i.e., the agent does not directly have access to the positions of PacMan and the ghosts.

\subsection{$DAI_{POMDP}$ agent \citep{DAI_POMDP}}

In this section, we review the approach proposed by \citet{DAI_POMDP}. The code is available here: \url{https://github.com/Grottoh/Deep-Active-Inference-for-Partially-Observable-MDPs}. The $DAI_{POMDP}$ agent is composed of five deep neural networks. 

The decoder $\mathcal{D}_{\theta_o}$ models $P_{\theta_o}(o_\tau|s_\tau)$ as a product of Bernoulli distributions, therefore: $P_{\theta_o}(o_\tau|s_\tau) = \MultiBernoulli(o_\tau;\hat{o}_\tau)$ where $\hat{o}_\tau = \mathcal{D}_{\theta_o}(s_\tau)$. The transition network $\mathcal{T}_{\theta_s}$ models $P_{\theta_s}(s_{\tau+1}|s_\tau,a_\tau)$ as a Gaussian distribution, i.e., $P_{\theta_s}(s_{\tau+1}|s_\tau,a_\tau) = \mathcal{N}(s_{\tau+1}|\mathring{\mu},\mathring{\sigma})$ where $\mathring{\mu},\ln \mathring{\sigma} = \mathcal{T}_{\theta_s}(s_\tau,a_\tau)$. The critic $\mathcal{G}_{\theta_a}$ outputs a vector containing the predicted expected free energy of each action, which is used to define the prior over action as $P_{\theta_a}(a_\tau|s_\tau) = \sigma[-\zeta \mathcal{G}_{\theta_a}(s_\tau, \bigcdot\,)]$, where $\sigma[\bigcdot]$ is a softmax function, $\zeta$ is the precision of the prior over actions, and $\mathcal{G}_{\theta_a}(s_\tau, \bigcdot\,)$ is the expected free energy of each action as predicted by the critic network when state $s_\tau$ is provided as input. The variational posterior over states $Q_{\phi_s}(s_t)$ is a Gaussian distribution modelled by the encoder $\mathcal{E}_{\phi_s}$, i.e., $Q_{\phi_s}(s_t) = \mathcal{N}(s_t;\mu, \sigma)$ where $\mu, \ln\sigma = \mathcal{E}_{\phi_s}(o_t)$. The variational posterior over actions $Q_{\phi_a}(a_t|s_t)$ is a categorical distribution modelled by the policy network $\mathcal{P}_{\phi_a}$, i.e., $Q_{\phi_a}(a_t|s_t) = \text{Cat}(a_t;\hat{\pi})$ where $\hat{\pi} = \mathcal{P}_{\phi_a}(s_t)$. Then, the agent is supposed to minimise the variational free energy defined as follows:
\begin{align*}
Q^*_{\phi}(s_t, a_t) &= \argmin_{Q_{\phi}(s_t, a_t)}\kl{Q_{\phi_a}(a_t|s_t)Q_{\phi_s}(s_t)}{P_{\theta_o}(o_t|s_t)P_{\theta_s}(s_t|s_{t-1},a_{t-1})P_{\theta_a}(a_t|s_t)}\\
&= \argmin_{Q_{\phi}(s_t, a_t)}\kl{Q_{\phi_s}(s_t)}{P_{\theta_s}(s_t|s_{t-1},a_{t-1})} + \kl{Q_{\phi_a}(a_t|s_t)}{P_{\theta_a}(a_t|s_t)} - \mathbb{E}_{Q_{\phi_s}(s_t)}[\ln P_{\theta_o}(o_t|s_t)].
\end{align*}
However, as explained in the paper, the KL-divergence (over states) is replaced by the mean square error (MSE) as follows:
\begin{align*}
Q^*_{\phi}(s_t, a_t) &= \argmin_{Q_{\phi}(s_t, a_t)} \text{MSE}(\mu, \mathring{\mu}) + \kl{Q_{\phi_a}(a_t|s_t)}{P_{\theta_a}(a_t|s_t)} - \mathbb{E}_{Q_{\phi_s}(s_t)}[\ln P_{\theta_o}(o_t|s_t)],
\end{align*}
where $\mu$ and $\mathring{\mu}$ are the mean vectors predicted by the encoder and the transition network, respectively. The paper justifies this substitution by saying that the maximum a posteriori (MAP) estimate is used to compute the state prediction error, instead of using the KL-divergence over the densities. However, the state prediction error and the KL-divergence over states are two different quantities, which are only equal when the two densities over states are Gaussian distributions with identity covariance matrix. However, the distribution predicted by the encoder network does not have an identity covariance matrix.

Put simply, in this context, the MSE and the KL-divergence between the densities over state are not necessarily equivalent. As a result, the $DAI_{POMDP}$ agent may not always follow the free energy principle.

\subsection{$DAI_{SSM}$ agent \citep{DAI_POMDP}}

The deep active inference agent proposed by \citep{ccatal2020learning} is based on a state space model, and is therefore called $DAI_{SSM}$. The code of this approach was not available online, but we were able to retrieve it from the authors. Importantly, $DAI_{SSM}$ is an offline approach meaning that the model is trained first on a fixed dataset gathered either by taking random actions in the environment or by manually controlling the robot. Then, when the model is trained, the expected free energy of different sequences of actions can be estimated using imaginary rollouts, and the first action of the policy with the lowest expected free energy is executed in the environment.

\subsection{Active exploration for robotic manipulation \citep{schneider2022active}}

The last paper that we review \citep{schneider2022active} is motivated from a reinforcement learning perspective, where the agent aims to maximise reward. However, instead of greedily maximising reward, the agent also maximises the information gain between the model parameters and the expected states and rewards, i.e.,
\begin{align}
\max_\pi \quad \underbrace{\mathbb{E}_{P(\bm{r}|\pi)}\big[f(\bm{r})\big]}_{\text{Expected reward}} + \,\, \beta \underbrace{\mathbb{E}_{P(\bm{s}, \bm{r}| \pi, s_t)}\big[\kl{P(\theta|\bm{s}, \bm{r}, \pi, s_t)}{P(\theta)}\big]}_{\text{Information gain}}, \quad \text{where:} \quad f(\bm{r}) = \sum_{\tau=t+1}^{t+H} r_\tau \label{eq:efe_like_eq}
\end{align}
where $t$ is the current time step, $H$ is the time horizon of planning, $\pi$ is the agent policy, $\theta$ are the parameters of the model, $\bm{r} = \bm{r}_{t+1:t+H}$ is the sequence of rewards between time step $t+1$ and $t+H$, $\bm{s} = \bm{s}_{t+1:t+H}$ is the sequence of states between time step $t+1$ and $t+H$, and $\beta$ is a hyper-parameter modulating the impact of information gain. Note, the distribution over the sequence of rewards $\bm{r}$ obtained by the agent when behaving according to a policy $\pi$, i.e., $P(\bm{r}|\pi)$, is not deterministic. Indeed, the same policy can produce different trajectories of states as the transition mapping is stochastic, and those different states can produce different rewards. Notice, the expectation w.r.t $P(\bm{r}|\pi)$ is iterating over all possible sequences of rewards, and passing each sequence to $f(\bm{r})$.
Then, the authors demonstrate the relationship between Equation \ref{eq:efe_like_eq} and the expected free energy. This relationship is explained in more details in \citep{schneider_master_thesis}, and relies on the the following assumptions: 
\begin{enumerate}
\item the states $s_\tau$ and rewards $r_\tau$ are latent variables for all $\tau \in \{t+1, ..., t+H\}$
\item the generative model contains observed variables for states $o^s_\tau$ and rewards $o^r_\tau$
\item the likelihood mappings, i.e., $P(o^r_\tau | r_\tau)$ and $P(o^s_\tau | s_\tau)$, are delta distributions which effectively make the latent variables fully observable, except for the model parameters $\theta$ that are distributed according to a delta distribution at time step zero and remain fixed over time
\item the variance of the transition mapping $P(s_{\tau+1} | s_\tau, a_\tau, \theta)$ does not depend on the parameters  $\theta$, e.g., $P(s_{\tau+1} | s_\tau, a_\tau, \theta) =  \mathcal{N}(s_{\tau+1}; \bm{\mu}, \bm{\sigma})$ where $\bm{\mu} = f_{\theta}(s_\tau, a_\tau)$ is predicted by a neural network with parameters $\theta$, and $\bm{\sigma} = \sigma I$ is a diagonal matrix whose diagonal elements are equal to $\sigma$.
\end{enumerate}
Importantly, while the above assumptions allow the authors to derive the expected free energy from Equation \ref{eq:efe_like_eq}, these assumptions also impose a lot of constraints on the model. For example, the proof does not hold if the likelihood mappings are not delta distributions. In this paper, we focus on such models. To conclude, this approach is a great contribution to reinforcement learning applied to robotic control, but cannot be considered as a complete deep active inference agent. The code of this approach is available at the following URL: \url{https://github.com/TimSchneider42/aerm}.

\subsection{Representational similarity with centered kernel alignment}\label{ssec:similarity}
The goal of representational similarity metrics is, as its name indicates, to measure the similarity between two representations.
In the context of deep learning, these representations correspond to $\mathbb{R}^{n \times p}$ matrices of activations, where
$n$ is the number of data examples and $p$ the number of neurons in a layer.~In this paper, we aim to use such metrics to compare the representations learned by the deep learning models described in Section~\ref{sec:build_dai} and the
representations learned by a DQN.

For our analysis, we will use Centred Kernel Alignment (CKA)~\citep{Cortes2012,Cristianini2002}, a normalised version of the Hillbert-Schmidt Independence Criterion (HSIC)~\citep{Gretton2005}, which measures the alignment between the $n \times n$ kernel matrices of two representations. \citet{Kornblith2019} have shown that for deep learning applications, linear kernels with centred layer activations worked well.~We thus focus on the linear CKA, also known as RV-coefficient~\citep{Robert1976}. Moreover, it has been shown to provide results similar to other representational similarity metrics, while being faster to compute~\citep{Bonheme2022}.

For conciseness, we will refer to linear CKA as CKA in the rest of this paper.
We now define CKA more formally. Given the centered layer activations $x \in \mathbb{R}^{n \times m}$ and $y \in \mathbb{R}^{n \times p}$ taken over $n$ data examples, CKA is defined as:
\begin{equation*}
    CKA(x, y) = \frac{\lVert y^Tx \rVert_F^2}{\lVert x^Tx \rVert_F\lVert y^Ty \rVert_F},
\end{equation*}
where $\lVert{\cdot}\rVert_F$ is the Frobenius norm, which is defined as:
\begin{align*}
\lVert{a}\rVert_F = \sqrt{\text{tr}(aa^T)} = \sqrt{\sum_{i=1}^k\sum_{j=1}^l |a_{ij}|^2},
\end{align*}
where $a \in \mathbb{R}^{k\times l}$ is an arbitrary $k \times l$ matrix, and $a^T$ is the transpose of $a$.

\paragraph{Limitations of CKA}
While CKA leads to accurate results in practice, it can be overly sensitive to differences in neural architectures \citep{Maheswaranathan2019}, and can thus underestimate the similarity between activations coming from layers of different type (e.g., convolutional and deconvolutional). Thus, we will only discuss the variation of similarity when analysing such cases.
For example, we will not compare $CKA(a, b)$ and $CKA(a, c)$ if $a$ and $b$ are convolutional layers but
$c$ is linear. We will, however, compare $CKA(a, c)$ and $CKA(b, c)$.

\section{Incrementally building a deep active inference agent} \label{sec:build_dai}

All of the deep active inference models we have presented make important contributions, illustrating a range of possible implementations. However, we do not 
feel that any of these approaches is a complete and definitive realisation of deep active inference. We have highlighted limitations of these published 
approaches throughout our presentation. Accordingly, in the remainder of this paper, we step back to first principles and ``build up" an agent 
component-by-component to determine which parts of a ``natural" deep active inference framework underlie its capacity to solve or fail to solve inference 
problems. Additionally, throughout this component-by-component investigation, we compare the different variants of deep active inference that result with a 
standard (well-attested) approach: a deep Q-network \citep{DeepRL}. Thus, in this section, we progresively build a deep active inference agent. Section 
\ref{ssec:dSprites_env} presents the dSprites environment in which all our simulations will be run. This environment was picked to test whether an active 
inference agent is able to solved the dSprites problem, as explored in \citep{DeepAIwithMCMC}. Section \ref{ssec:env_agent_iter} describes how the agents 
introduced later in this paper interact with the environment. Then, Section \ref{ssec:VAE} introduces a variational auto-encoder (VAE) agent, Section 
\ref{ssec:HMM} discusses a deep hidden Markov model (HMM) agent, Section \ref{ssec:CHMM} presents a deep critical HMM (CHMM) agent, and finally, Section 
\ref{ssec:DAI} introduces a complete deep active inference agent. Note, the notation used throughout this section are summarised in Appendix A.

\subsection{dSprites environment} \label{ssec:dSprites_env}

\begin{figure}[h]
	\begin{center}
	\includegraphics[scale=0.4]{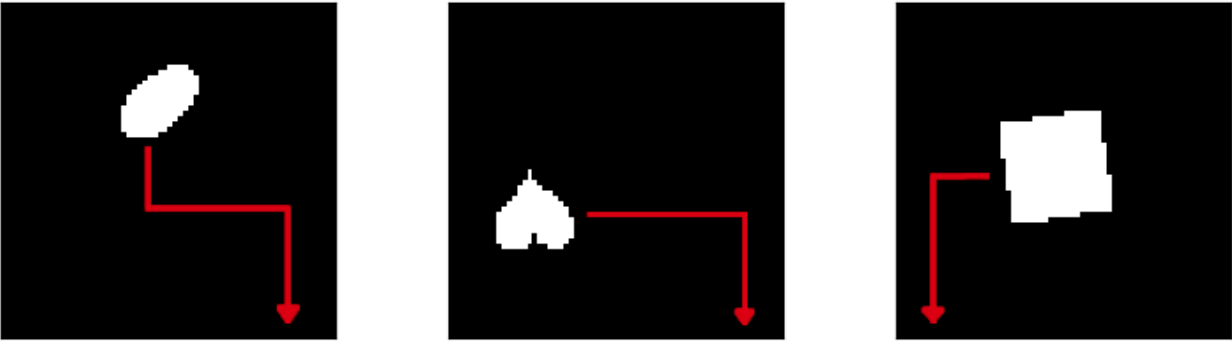}
	\end{center}
  \caption{This figure illustrates the dSprites environment, in which the agent must move all squares towards the bottom-left corner of the image and all ellipses and hearts towards the bottom-right corner of the image. The red arrows show the behaviour expected from the agent.}
   \label{fig:dSprites_env}
\end{figure}

The dSprites environment is based on the dSprites dataset \citep{dsprites17}, initially designed for analysing the latent representation learned by variational auto-encoders \citep{VAE}. The dSprites dataset is composed of images of squares, ellipses and hearts. Each image contains one shape (square, ellipse or heart) with its own scale, orientation, and $(X,Y)$ position. In the dSprites environment, the agent is able to move those shapes around by performing four actions (i.e., UP, DOWN, LEFT, RIGHT). To make the task tractable, the action selected by the agent is executed eight times in the environment before the beginning of the next action-perception cycle, i.e., the $X$ or $Y$ position is increased or decreased by eight between time step $t$ and $t+1$. The goal of the agent is to move all squares towards the bottom-left corner of the image and all ellipses and hearts towards the bottom-right corner of the image, c.f. Figure \ref{fig:dSprites_env}.

\subsection{Agent-environment interaction} \label{ssec:env_agent_iter}

In this section, we present how all the agents introduced in the next sections interact with the environment. Each agent was trained for $N = 500K$ iterations. At the begining of a trial, the environment is reset to a random state and the agent receives an observation\footnote{Each observation contains a sequence of three images, i.e., the image corresponding to the current state of the environment, and the two images gathered during the previous two time steps.} $o_t$. Using $o_t$, the agent selects an action $a_t$, which is then executed in the environment. This leads the agent to receive a new obervation $o_{t+1}$, a reward $r_{t+1}$ and a boolean $done$ describing whether the trial is over or not. Then, the new experience ($o_t$, $a_t$, $o_{t+1}$, $r_{t+1}$, $done$) is added to the replay buffer, from which a batch is sampled to train the various neural networks of the agent. Finally, if the trial has ended, then the environment is reset to a random state leading to a new observation $o_t$, otherwise $o_{t+1}$ becomes the new $o_t$ closing the action-perception cycle. Algorithm \ref{algo:ap_cycles} summarises the agent-environment interaction.

\begin{algorithm}[H]
\label{algo:ap_cycles}
\SetAlgoLined\DontPrintSemicolon
\SetKwInOut{Input}{Input}
\SetKwFor{RepTimes}{repeat}{times}{end}
\SetAlgoLined
\Input{$env$ the environment,\\
       $agent$ the agent,\\
       $buffer$ the replay buffer,\\
       $N$ the number of training iterations.
}
 {\color{white}space}\;
$o_t$ = env.reset() \tcp*{Get the initial observation from environment}
 \RepTimes{$N$} {
   $a_t \leftarrow $ select\_action($o_t$) \tcp*{Select an action}
   $o_{t+1}, r_{t+1}, done \leftarrow $ env.execute($a_t$) \tcp*{Execute the action in the environment}
   buffer.push\_new\_experience($o_t$, $a_t$, $o_{t+1}$, $r_{t+1}$, $done$) \tcp*{Add the experience to the replay buffer}
   agent.learn(buffer) \tcp*{Perform one iteration of training}
   \eIf{done == True}{
      $o_t \leftarrow$ env.reset() \tcp*{Reset the environment when a trial ends}
   } {
      $o_t \leftarrow o_{t+1}$
   }
 }
 \caption{The interaction between the agent and the environment.}
\end{algorithm}

\subsection{Variational auto-encoder} \label{ssec:VAE}

In this section, we present our first agent based on a variational auto-encoder. The agent is composed of two deep neural networks, i.e., an encoder and a decoder. The encoder $\mathcal{E}_{\phi_s}$ takes as input an image $o_t$ and outputs the parameters of the variational posterior $Q_{\phi_s}(s_t) = \mathcal{N}(s_t;\mu,\sigma)$, where $\mu$ is the mean vector of the Gaussian distribution, and $\sigma$ are the diagonal elements of the covariance matrix. The decoder $\mathcal{D}_{\theta_o}$ models the likelihood mapping $P_{\theta_o}(o_t|s_t)$, which attributes a probability to each image $o_t$ given a state $s_t$, and is defined as:
\begin{align*}
P_{\theta_o}(o_t|s_t) = \MultiBernoulli(o_t;\hat{o}_t),
\end{align*}
where $\hat{o}_t = \mathcal{D}_{\theta_o}(s_t)$ are the values predicted by the decoder, and $\MultiBernoulli(o_t;\hat{o}_t)$ is a product of Bernoulli distributions defined as:
$$\MultiBernoulli(o_t;\hat{o}_t) = \prod_{x,y} \text{Bernoulli}(o_t[x,y];\hat{o}_t[x,y]),$$
where $\text{Bernoulli}(\,\bigcdot\,;\,\bigcdot\,)$ is a Bernoulli distribution over the possible values of the pixel $o_t[x,y]$, parameterized by the parameter $\hat{o}_t[x,y]$, which is predicted by the decoder network. The goal of the agent is to minimise the variational free energy (VFE):
$$\bm{F} = \kl{Q_{\phi_s}(s_t)}{P_{\theta_o}(o_t,s_t)} = \kl{Q_{\phi_s}(s_t)}{P_{\theta_o}(o_t|s_t)P(s_t)},$$
where $P(s_t) = \mathcal{N}(s_t; 0, I)$ is an isotropic (multivariate) Gaussian with variance one. The VFE can be re-arranged as follows:
$$\bm{F} = \kl{Q_{\phi_s}(s_t)}{P(s_t)} - \mathbb{E}_{Q_{\phi_s}(s_t)}[\ln P_{\theta_o}(o_t|s_t)],$$
where the KL-divergence between two Gaussian distributions can be computed using an analytical solution, and the expectation of the logarithm of $P_{\theta_o}(o_t|s_t)$ is approximated by a Monte-Carlo estimate using a single sample $\hat{s}_t \sim Q_{\phi_s}(s_t)$. The sample $\hat{s}_t$ is obtained using the reparameterisation trick as follows: $\hat{s}_t = \mu + \sigma \odot \hat{\epsilon}$, where $\odot$ is an element-wise product between two vectors, and $\hat{\epsilon} \sim \mathcal{N}(\epsilon;0,I)$.

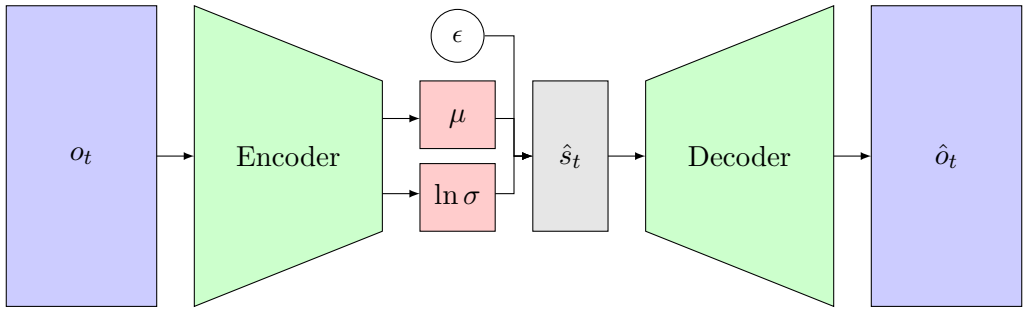
\begin{figure}[h]
	\begin{center}
	\begin{tikzpicture}[square/.style={regular polygon,regular polygon sides=4}]
		\pic{vae=$o_t$/$\mu$/$\ln \sigma$/$\hat{s}_t$/$\hat{o}_t$};
    \end{tikzpicture}

	\end{center}
  \caption{This figure illustrates the VAE agent. From left to right, we have the input image $o_t$, the encoder network, the layer of mean $\mu$ and log variance $\ln \sigma$, the epsilon random variable used for the reparameterisation trick, the latent state $\hat{s}_t$, the decoder network, and finally, the reconstructed image $\hat{o}_t$. Note, there are no actions in this agent's generative model. Therefore, the VAE agent takes random actions.}
   \label{fig:VAE}
\end{figure}

To sum up, this agent takes random actions, and stores its experiences in a replay buffer (c.f. Section \ref{ssec:env_agent_iter}). Then, batches of experiences ($o_t$, $a_t$, $o_{t+1}$, $r_{t+1}$, $done$) are sampled from the replay buffer. The observations at time step $t$ are then fed into the encoder, which outputs the mean and log variance of a Gaussian distribution $Q_{\phi_s}(s_t) = \mathcal{N}(s_t;\mu, \sigma)$. A latent state is sampled from $Q_{\phi_s}(s_t)$ using the re-parameterisation trick, and is then provided as input to the decoder which outputs the parameters of Bernoulli distributions $\hat{o}_t$. The KL-divergence between $Q_{\phi_s}(s_t)$ and $P(s_t)$ is computed analytically, and the logarithm of $P_{\theta_o}(o_t|s_t)$ reduces to the binary cross entropy (BCE) because $P_{\theta_o}(o_t|s_t)$ is a product of Bernoulli distributions. Next, the VFE is obtained by subtracting the BCE from the KL-divergence, and back-propagation is used to update the weights of the encoder and decoder networks. Figure \ref{fig:VAE} illustrates the VAE agent presented in this section. Note, this agent takes random actions.

\subsection{Deep hidden Markov model} \label{ssec:HMM}

\begin{figure}[h]
	\begin{center}
	\begin{tikzpicture}
		\pic{hmm};
    \end{tikzpicture}
	\end{center}
	\caption{This figure illustrates the HMM agent. On the left and right, one can see two auto-encoders, i.e., one at time step $t$ and one at time step $t+1$. In the middle, the transition network takes as input the state and action at time $t$, i.e., $(\hat{s}_t, a_t)$, and outputs the mean $\mathring{\mu}$ and log variance $\ln\mathring{\sigma}$ of a Gaussian distribution. By sampling the latent variable $\epsilon$, and using the reparameterisation trick, we get the latent state outputed by the transition network: $\mathring{s}_{t+1} = \mathring{\mu} + \mathring{\sigma} \odot \hat{\epsilon}$ where $\hat{\epsilon}$ is sampled from a Gaussian distribution with mean zero and variance one. Importantly, the model seems to be composed of two disconnected parts, however, the variational free energy will have a complexity term between the Gaussian distributions outputed by the transition network and encoder at time $t+1$. Note, this agent takes random actions.}
   \label{fig:HMM}
\end{figure}
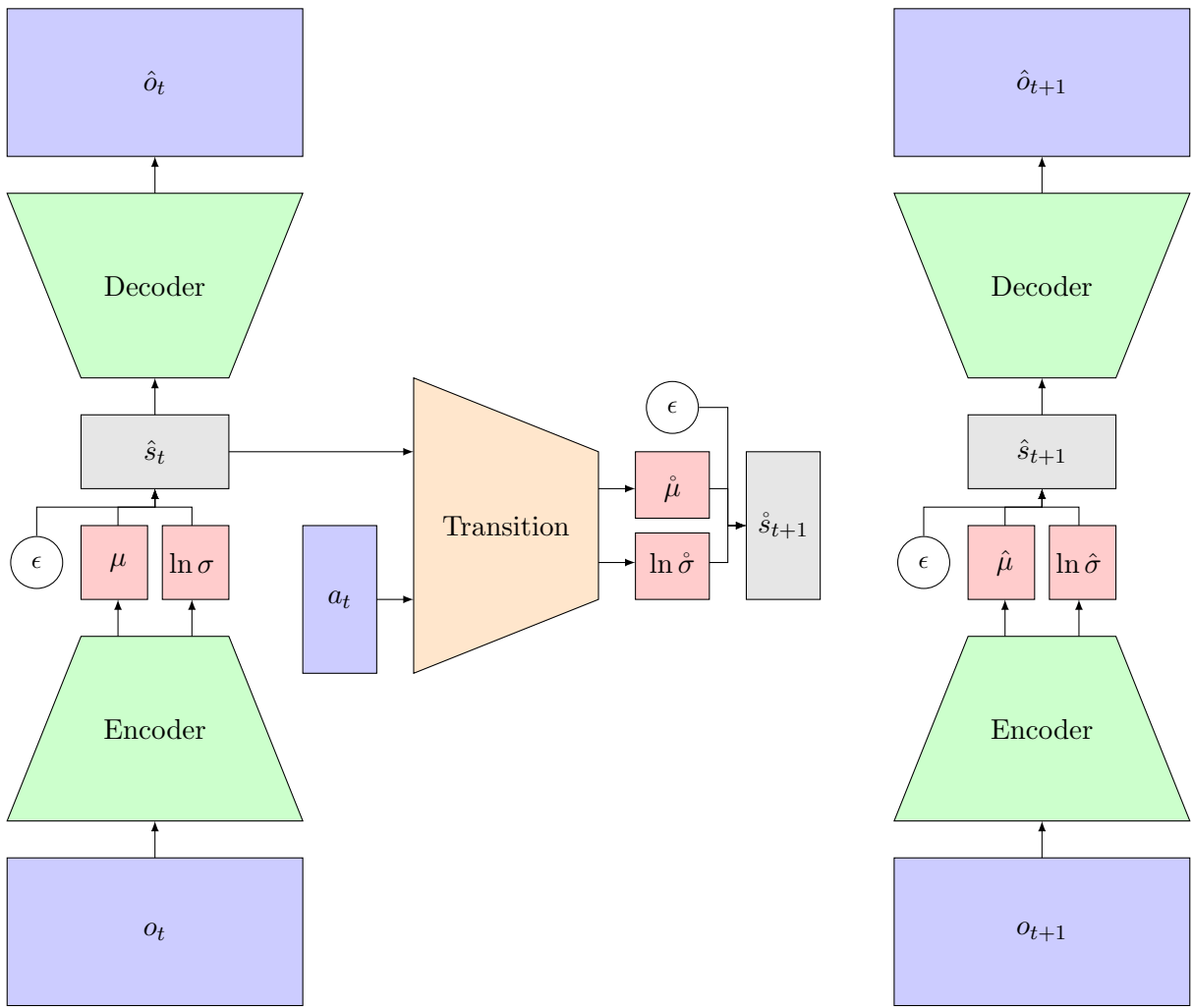

In this section, we present our second agent based on a hidden Markov model. Similarly to the VAE agent, the HMM agent is composed of an encoder network modelling $Q_{\phi_s}(s_\tau)$, and a decoder network modelling $P_{\theta_o}(o_\tau|s_\tau)$. However, the prior over the hidden states at time step $t+1$ depends on the hidden states and action at time step $t$. This prior is modelled by the transition network $\mathcal{T}_{\theta_s}$ that predicts the parameters of the Gaussian distribution $P_{\theta_s}(s_{t+1}|s_t,a_t) = \mathcal{N}(s_{t+1}; \mathring{\mu}, \mathring{\sigma})$, where $\mathring{\mu}$ is the mean of the Gaussian distribution, and $\mathring{\sigma}$ are the diagonal elements of the covariance matrix. Recall, that the goal of the inference process is to fit the approximate posterior $Q_{\phi_s}(s_t)$ to the true posterior $P(s_t|o_t, s_{t-1}, a_{t-1})$. Formally, this optimisation can be written as the minimization of the Kullback-Leibler divergence between the approximate and the true posterior, i.e.,
\begin{align*}
Q^*(s_t) &= \argmin_{Q_{\phi_s}(s_t)} \kl{Q_{\phi_s}(s_t)}{P(s_t|o_t, s_{t-1}, a_{t-1})}.
\end{align*}
Using a derivation almost identical to the one presented in Section \ref{ssec:derive_vfe_in_fountas}, the VFE can be proven to be:
\begin{align}
Q^*_{\phi_s}(s_t)\,\, &= \argmin_{\quad Q_{\phi_s}(s_t) \quad }\underbrace{\kl{Q_{\phi_s}(s_t)}{P_{\theta_o}(o_t|s_t)P_{\theta_s}(s_t|s_{t-1},a_{t-1})}}_{\text{variational free energy}}\label{eq:vfe_defi}\\
&= \argmin_{\quad Q_{\phi_s}(s_t)\quad } \kl{Q_{\phi_s}(s_t)}{P_{\theta_s}(s_t|s_{t-1},a_{t-1})} - \mathbb{E}_{Q_{\phi_s}(s_t)}[P_{\theta_o}(o_t|s_t)].\nonumber
\end{align}
The VFE of Equation \ref{eq:vfe_defi} can be computed in a similar way to the VAE agent. Put simply, this agent takes random actions, and stores its experiences in a replay buffer (c.f. Section \ref{ssec:env_agent_iter}). Then, batches of experiences ($o_{t-1}$, $a_{t-1}$, $o_t$, $r_t$, $done$) are sampled from the replay buffer. The observations at time step $t - 1$ are feed into the encoder, which outputs the mean and log variance of a Gaussian distribution $Q_{\phi_s}(s_{t-1}) = \mathcal{N}(s_{t-1};\mu, \sigma)$. A latent state is sampled from $Q_{\phi_s}(s_{t-1})$ using the re-parameterisation trick, and is then provided as input to the transition network along with action $a_{t-1}$. The transition network outputs the parameters of the Gaussian distributions $P_{\theta_s}(s_t|s_{t-1},a_{t-1}) = \mathcal{N}(s_t;\mathring{\mu}, \mathring{\sigma})$. Additionally, the observations at time step $t$ can be fed into the encoder to get the parameters of $Q_{\phi_s}(s_t) = \mathcal{N}(s_t;\hat{\mu}, \hat{\sigma})$. Sampling from $Q_{\phi_s}(s_t)$ using the reparameterisation trick gives a state $\hat{s}_t$ that when given as input to the decoder produces the parameters of a product of Bernoulli distributions $\hat{o}_{t+1}$. The KL-divergence between $Q_{\phi_s}(s_t)$ and $P_{\theta_s}(s_t|s_{t-1},a_{t-1})$ is computed analytically, and the logarithm of $P_{\theta_o}(o_t|s_t)$ reduces to the binary cross entropy (BCE) because $P_{\theta_o}(o_t|s_t)$ is a product of Bernoulli distributions. Next, the VFE is obtained by subtracting the BCE from the KL-divergence, and back-propagation is used to update the weights of the encoder, decoder and transition networks. Figure \ref{fig:HMM} illustrates the HMM agent. 

\subsection{Deep critical HMM} \label{ssec:CHMM}

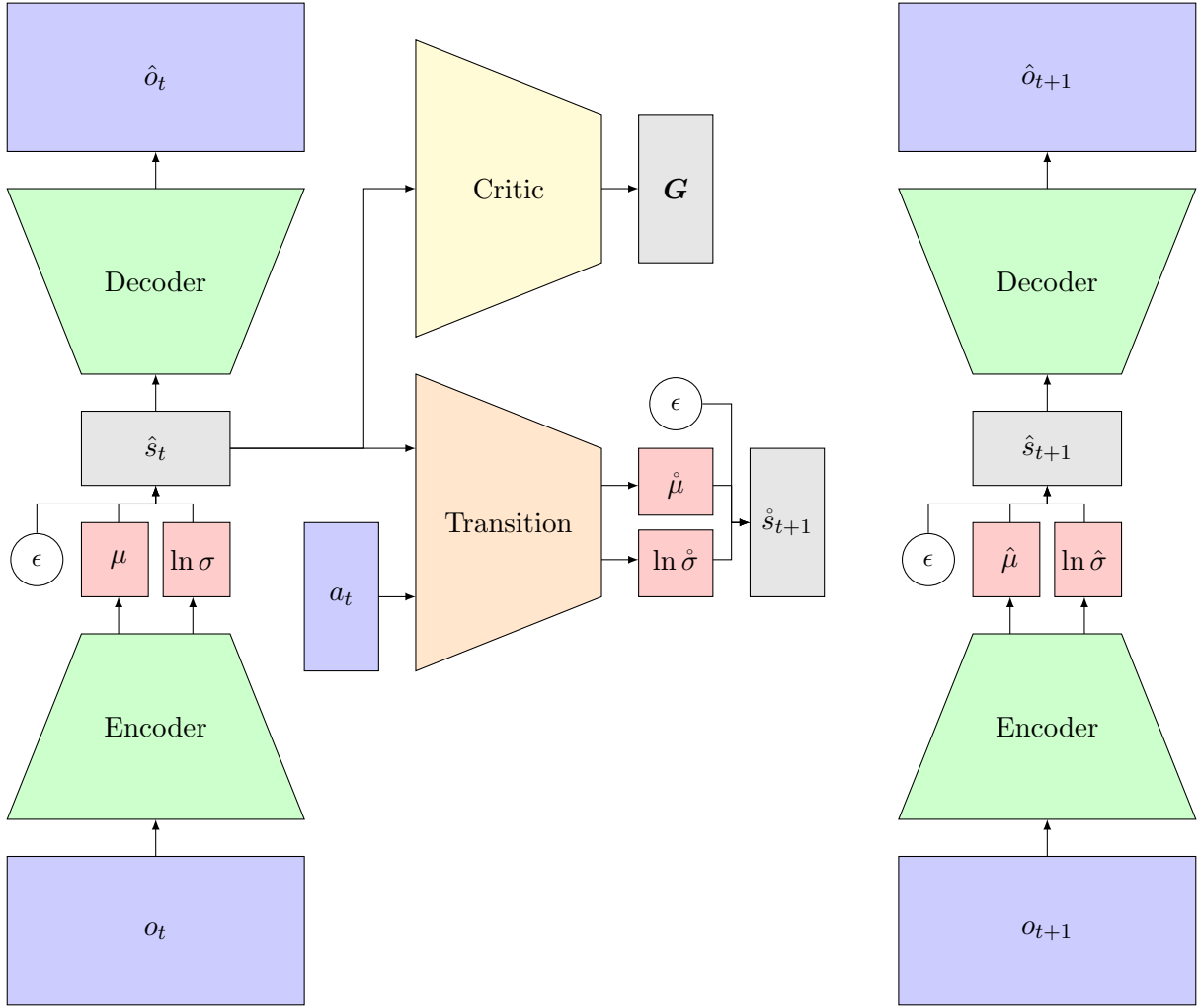
\begin{figure}[h]
	\begin{center}
	\begin{tikzpicture}[square/.style={regular polygon,regular polygon sides=4}]
		\coordinate (A) at (13.5,8);
		\coordinate (B) at (16,9);
		\coordinate (C) at (16,11);
		\coordinate (D) at (13.5,12);
		\draw[fill=yellow!20!white] (A) -- coordinate[pos=.6] (AB) (B)--(C)
         --coordinate[pos=.45] (CD) (D)
         --coordinate[pos=.55] (DA) cycle;
	    \node at (14.75, 10) {Critic};

		\draw[-latex] (11,6.5) -- (12.8,6.5) -- (12.8,10) -- (13.5,10);
		\draw[-latex] (16, 10) -- (16.5, 10);
		
		\draw[fill=gray!20!white] (16.5,9) rectangle (17.5,11);
	    \node at (17, 10) {$\bm{G}$};

		\pic{hmm};
    \end{tikzpicture}
	\end{center}
  \caption{This figure illustrates the CHMM agent. The only new part is the critic network, which takes as input the hidden state at time $t$ and ouputs the expected free energy of each action $\bm{G}$. Importantly, the CHMM takes actions based on the EFE.}
   \label{fig:CHMM}
\end{figure}

In this section, we present our third agent that incorporates a critic network to the deep HMM presented in the previous section. The resulting model is called a deep CHMM and is illustrated in Figure \ref{fig:CHMM}. The CHMM is equipped with an encoder $\mathcal{E}_{\phi_s}$ modelling $Q_{\phi_s}(s_\tau)$, a decoder $\mathcal{D}_{\theta_o}$  modelling $P_{\theta_o}(o_\tau|s_\tau)$, a transition network $\mathcal{T}_{\theta_s}$ modelling $P_{\theta_s}(s_t|s_{t-1},a_{t-1})$, and a critic network $\mathcal{G}_{\theta_a}$ that predicts the expected free energy (see below) of each action. The critic is then used to define the prior over actions as: $P_{\theta_a}(a_t|s_t) = \sigma[-\zeta \mathcal{G}_{\theta_a}(s_t,\bigcdot\,)]$, where $\zeta$ is the precision of the prior over actions, and $\mathcal{G}_{\theta_a}(s_t,\bigcdot\,)$ is the EFE of taking each action in state $s_t$ as predicted by the critic. The encoder, decoder and transition networks are all trained in the same way as before to minimise the VFE. The critic however is trained to minimise the smooth L1 norm between its output $\mathcal{G}_{\theta_a}(s_t,\bigcdot\,)$ and the target G-values $y(\,\bigcdot\,)$, i.e., the critic minimises $\text{SL1}[\mathcal{G}_{\theta_a}(s_t,\bigcdot\,), y(\,\bigcdot\,)]$. Note, the SL1 was picked because it is less sensitive to outliers than the MSE, and is defined as:
$$SL1[x, y]= \begin{cases}
      0.5 \times \frac{(x - y)^2}{\beta} & \text{if }\lvert x-y \rvert < \beta\\
      \lvert x-y \rvert - 0.5 \times \beta & \text{otherwise}
\end{cases},$$
where $\beta$ is an hyper-parameter such that as $\beta$ goes to zero, the SL1 loss converges to the L1 loss, and when $\beta$ tends to $+\infty$, the SL1 loss converges to a constant zero loss. Intuitively, the SL1 loss uses a squared term if the absolute element-wise error falls below $\beta$, and an L1 term otherwise.
Addtionally, the target G-values are defined as:
$$y(a_t) = G_{t+1}(a_t) + \gamma \mathbb{E}_{Q_{\phi_s}(s_{t+1})}\Big[ \max_{a_{t+1} \in \mathcal{A}} \hat{\mathcal{G}}_{\hat{\theta}_a}(s_{t+1}, a_{t+1})\Big],$$
where $Q_{\phi_s}(s_{t+1})$ can be computed by feeding the image $o_{t+1}$ sampled from the replay buffer as input to the encoder, $G_{t+1}(a_t)$ is the expected free energy at time $t+1$ after taking action $a_t$ (see below), and $\gamma$ is a discount factor. Note, the above Equation is an application of Bellman's equation \citep{Bellman1952} to the expected free energy. Also, as for the DQN agent, we improved the training stability by implementing a target network $\hat{\mathcal{G}}_{\hat{\theta}_a}$, which is structurally identical to the critic and whose weights are synchronised with the weights of the critic every $K$ (learning) iterations. The last question to answer before focusing on the subject of the EFE is: how does the CHMM select the action to be performed in the environment?

There are at least four possibilities: (i) select a random action, (ii) select the action that maximises EFE according to the critic, i.e., $a_t^* = \argmax_{a_t} \mathcal{G}_{\theta_a}(s_t,a_t)$, (iii) sample an action from a softmax function of the output of the critic, i.e., $a_t^* \sim \sigma[\mathcal{G}_{\theta_a}(s_t,\bigcdot\,)]$ where $\sigma[\,\bigcdot\,]$ is a softmax function, and (iv) use the $\mathring{\epsilon}$-greedy algorithm with exponential decay, i.e., select a random action with probabilty $\mathring{\epsilon}$ or select the best action with probability $1 - \mathring{\epsilon}$ where $\mathring{\epsilon}$ starts with a high value and decays exponentially fast.

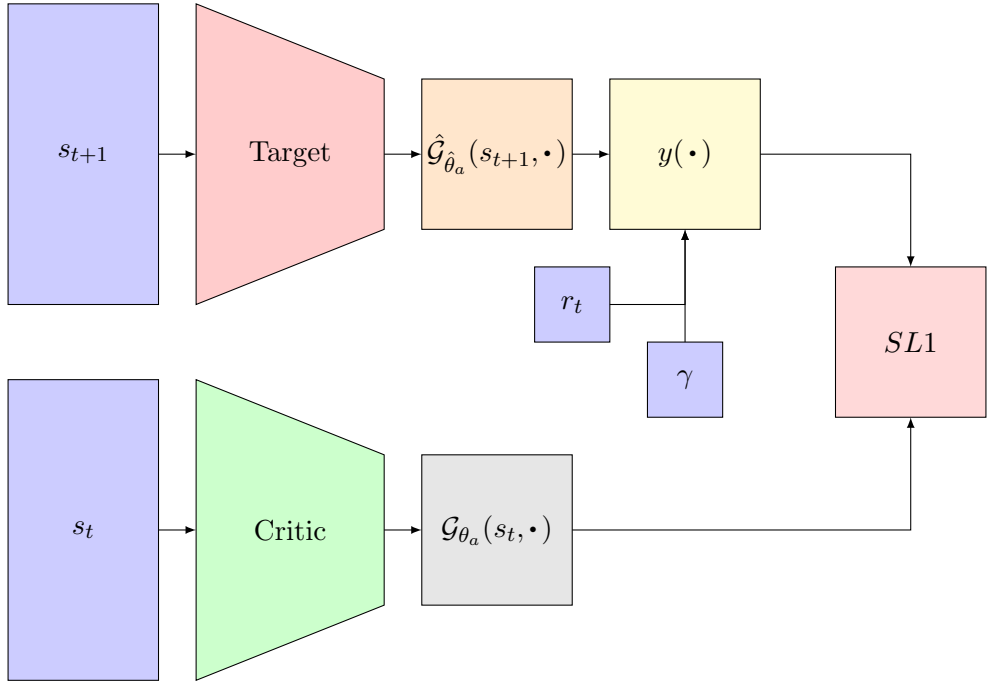
\begin{figure}[h]
	\begin{center}
	\begin{tikzpicture}[square/.style={regular polygon,regular polygon sides=4}, scale=1]
		\draw[fill=blue!20!white] (-1,-12) rectangle (1,-8);
	    \node at (0, -10) {$s_t$};

		\coordinate (A) at (1.5,-12);
		\coordinate (B) at (4,-11);
		\coordinate (C) at (4,-9);
		\coordinate (D) at (1.5,-8);
		\draw[fill=green!20!white] (A) -- coordinate[pos=.6] (AB) (B)--(C)
         --coordinate[pos=.45] (CD) (D)
         --coordinate[pos=.55] (DA) cycle;
	    \node at (2.75, -10) {Critic};

		\draw[fill=gray!20!white] (4.5,-11) rectangle (6.5,-9);
	    \node at (5.5, -10) {$\mathcal{G}_{\theta_a}(s_t, \bigcdot\,)$};

		\draw[fill=blue!20!white] (-1,-7) rectangle (1,-3);
	    \node at (0, -5) {$s_{t+1}$};

		\coordinate (A) at (1.5,-7);
		\coordinate (B) at (4,-6);
		\coordinate (C) at (4,-4);
		\coordinate (D) at (1.5,-3);
		\draw[fill=red!20!white] (A) -- coordinate[pos=.6] (AB) (B)--(C)
         --coordinate[pos=.45] (CD) (D)
         --coordinate[pos=.55] (DA) cycle;
	    \node at (2.75, -5) {Target};

		\draw[fill=orange!20!white] (4.5,-6) rectangle (6.5,-4);
	    \node at (5.5, -5) {$\hat{\mathcal{G}}_{\hat{\theta}_a}(s_{t+1}, \bigcdot\,)$};

		\draw[fill=blue!20!white] (6,-6.5) rectangle (7,-7.5);
	    \node at (6.5, -7) {$r_t$};

		\draw[fill=blue!20!white] (7.5,-7.5) rectangle (8.5,-8.5);
	    \node at (8, -8) {$\gamma$};

		\draw[fill=yellow!20!white] (7,-4) rectangle (9,-6);
	    \node at (8, -5) {$y(\,\bigcdot\,)$};

		\draw[fill=pink!60!white] (10,-6.5) rectangle (12,-8.5);
	    \node at (11, -7.5) {$SL1$};

		\draw[-latex] (4,-10) -- (4.5,-10);
		\draw[-latex] (1,-10) -- (1.5,-10);
		\draw[-latex] (4,-5) -- (4.5,-5);
		\draw[-latex] (1,-5) -- (1.5,-5);
		\draw[-latex] (6.5,-5) -- (7,-5);
		\draw[-latex] (7,-7) -- (8,-7) -- (8,-6);
		\draw[-latex] (9,-5) -- (11,-5) -- (11,-6.5);
		\draw[-latex] (6.5,-10) -- (11,-10) -- (11,-8.5);         
		\draw[-latex] (8,-7.5) -- (8,-6);
    \end{tikzpicture}

	\end{center}
  \caption{This figure illustrates the computation of the critic's loss function when the critic is only maximising reward, i.e., when Equation \ref{eq:reward_maximisation} is used for the expected free energy. Briefly, the state $s_t$ is fed into the Critic, and the state $s_{t+1}$ is fed into the target network. The critic outputs the G-values for each action at time $t$, and the target network outputs the G-values for each action at time $t+1$. Then, the reward, the discount factor, and G-values of each action at time $t+1$ are used to compute the target values $y(\,\bigcdot\,)$. Finally, the goal is to minimise the SL1 between the prediction of the critic and the target values by changing the weights of the critic.}
     \label{fig:reward_usage}
\end{figure}

\subsubsection{Expected free energy} \label{ssec:efe}

In this section, we discuss the definition of the expected free energy (EFE) before investigating various ways to implement it in the context of deep active inference. Recently in \citep{Parr304782}, the expected free energy was defined as:
\begin{align}
G(\pi) = \sum_{\tau = t+1}^T G_\tau(\pi) = \sum_{\tau = t+1}^T \mathbb{E}_{P(o_\tau|s_\tau)Q(s_\tau | \pi)}\big[\ln Q(s_\tau | \pi) - \ln P(o_\tau, s_\tau)\big],\label{eq:efe_definition}
\end{align}
where $P(o_\tau|s_\tau)$ is the likelihood mapping, $Q(s_\tau | \pi)$ is the variational distribution, and in the literature, $P(o_\tau, s_\tau)$ is called the generative model but is better understood as a target distribution encoding the prior preferences of the agent. Indeed, assuming the standard generative model of active inference (i.e., a partially observable Markov decision process), the hidden states $s_\tau$ should depend on $s_{\tau-1}$ and $a_{\tau-1}$. This point is important because it impacts the question of whether $P(o_\tau, s_\tau)$ is indeed the generative model, and therefore whether the expected free energy is the expectation of the variational free energy. According to the free energy principle, an active inference agent must minimise its (variational) free energy. However, if such a relationship cannot be established between the expected and variational free energy, then one cannot claim that an agent minimising expected free energy also minimises its variational free energy. Additionally, we need to re-arrange the definition of the EFE stated in \eqref{eq:efe_definition} to allow rewards to be incorporated:
\begin{align}
G_\tau(\pi) &= \mathbb{E}_{P(o_\tau|s_\tau)Q(s_\tau | \pi)}\big[\ln Q(s_\tau | \pi) - \ln P(o_\tau, s_\tau)\big]\nonumber\\
&= \mathbb{E}_{P(o_\tau|s_\tau)Q(s_\tau | \pi)}\big[\ln Q(s_\tau | \pi) - \ln P(s_\tau|o_\tau)- \ln P(o_\tau)\big]\nonumber\\
&\approx \mathbb{E}_{P(o_\tau|s_\tau)Q(s_\tau | \pi)}\big[\ln Q(s_\tau | \pi) - \ln Q(s_\tau)- \ln P(o_\tau)\big]\nonumber\\
&= \underbrace{\kl{Q(s_\tau | \pi)}{Q(s_\tau)}}_{\text{epistemic value}} - \underbrace{\mathbb{E}_{P(o_\tau|s_\tau)Q(s_\tau | \pi)}\big[\ln P(o_\tau)\big]}_{\text{extrinsic value}},\label{eq:efe_practice}
\end{align}

\subsubsection{A principled estimate of the EFE at time $t + 1$?}

Now, the question is how to estimate \eqref{eq:efe_practice}, and we focus on the case where $\tau = t + 1$. Note, because $\tau = t+1$, the policy $\pi$ contains only one action $a_t$, i.e., $\pi = a_t$. In the tabular version of active inference, the variational distribution is composed of a factor $Q(s_\tau | \pi)$. However, in the deep active inference literature, the variational distribution does not contain such a factor. Generally, a Monte-Carlo estimate is used as follows:
\begin{align}
Q(s_{t+1} | a_t) = \mathbb{E}_{Q_{\phi_s}(s_t)}\big[P_{\theta_s}(s_{t+1}|s_t, a_t)\big] \approx \frac{1}{N} \sum_{i = 1}^N P_{\theta_s}(s_{t+1}|s_t = \hat{s}_t^i, a_t), \label{eq:vd_or_gm}
\end{align}
where $\hat{s}_t^i \sim Q_{\phi_s}(s_t)$. Importantly, for the expected free energy to be the expectation of the variational free energy, $Q(s_{t+1} | a_t)$ should be a factor of the variational distribution. However, \eqref{eq:vd_or_gm} is estimated using a factor of the generative model $P_{\theta_s}(s_{t+1}|s_t = \hat{s}_t^i, a_t)$. This is a conceptual issue, associated with current deep active inference approaches, such as \citet{DeepAIwithMCMC}.
In what follows, we use $N=1$ leading to a simplified version of the estimate:
\begin{align}
Q(s_{t+1} | a_t) \approx P_{\theta_s}(s_{t+1}|s_t = \hat{s}_t, a_t).
\end{align}
Note, in the above equation, $\hat{s}_t^i$ is denoted $\hat{s}_t$ because there is only one sample, i.e., $N=1$. At this point, we have an estimate for $Q(s_{t+1} | a_t)$ and $Q_{\phi_s}(s_t)$ is the variational distribution. The only missing piece is an estimate of the extrinsic value. In the tabular version of active inference, the preferences of the agent can be related to the rewards from the reinforcement learning literature. In this paper, we follow \citep{dacosta2020relationship} and define the prior preferences as:
\begin{align*}
P(o_\tau) = \frac{\exp(\psi r_\tau[o_\tau])}{\sum_{o_\tau} \exp(\psi r_\tau[o_\tau])},
\end{align*}
where $\psi$ is the precision of the prior preferences, and $r_\tau[o_\tau]$ is the reward obtained when making observation $o_\tau$. Taking the logarithm of the above equation leads to:
\begin{align}
\ln P(o_\tau) &= \psi r_\tau[o_\tau] - \ln \sum_{o_\tau} \exp(\psi r_\tau[o_\tau])\nonumber\\
&= \psi r_\tau[o_\tau] + C, \label{eq:prior_pref}
\end{align}
where we used the fact that the summation over all $o_\tau$ is a normalisation term, i.e., a constant. Using \eqref{eq:prior_pref}, we can now create an estimate of the extrinsic value as follows:
\begin{align*}
\mathbb{E}_{P_{\theta_o}(o_\tau|s_\tau)Q(s_\tau | a_t)}\big[\ln P(o_\tau)\big] \approx \frac{1}{M} \sum_{i = 1}^M \ln P(o_\tau = \hat{o}_\tau) = \frac{1}{M} \sum_{i = 1}^M \psi r_\tau[o_\tau] + C,
\end{align*}
where $\hat{o}_\tau \sim P_{\theta_o}(o_\tau|s_\tau=\hat{s}_\tau)$ and $\hat{s}_\tau \sim Q(s_\tau | a_t)$. In what follows, we use $M=1$ and discard the constant\footnote{Removing a constant does not influence which policy is the best. Indeed, $\pi^* = \argmax_\pi G(\pi) = \argmax_\pi G(\pi) - C$.}, which leads to a simplified version of the estimate:
\begin{align*}
\mathbb{E}_{P_{\theta_o}(o_\tau|s_\tau)Q(s_\tau | a_t)}\big[\ln P(o_\tau)\big] \delequal \psi r_\tau[o_\tau] \delequal \psi r_\tau,
\end{align*}
where we simplied the notation by deno in which all our simulations will be ting $r_\tau[o_\tau]$ as $r_\tau$. To conclude, we have the following estimate for the EFE at time $\tau = t+1$:
\begin{align}
G_{t+1}(a_t) &\approx \kl{Q(s_{t+1} | a_t)}{Q_{\phi_s}(s_{t+1})} - \mathbb{E}_{P_{\theta_o}(o_{t+1}|s_{t+1})Q(s_{t+1} | a_t)}\big[\ln P(o_{t+1})\big]\nonumber\\
&\approx \kl{P_{\theta_s}(s_{t+1}|s_t = \hat{s}_t, a_t)}{Q_{\phi_s}(s_{t+1})} - \psi r_{t+1},
\end{align}
where $\hat{s}_t \sim Q_{\phi_s}(s_t)$, $P_{\theta_s}(s_{t+1}|s_t, a_t)$ is known from the generative model, $Q_{\phi_s}(s_{t+1})$ is known from the variational distribution, the KL-divergence can be estimated using an analytical solution, $\psi$ is a hyperparameter modulating the precision of the prior preferences, and $r_{t+1}$ is the reward obtained at time step $t+1$. As shown in Figure \ref{fig:reward_usage}, the reward at time step $t+1$ is used to compute the target values that must be predicted by the critic network.

\subsubsection{Other definitions of the EFE at time $t + 1$} \label{ssec:efe_other_defe}

In the previous section, we have presented what may be a principled way to estimate the EFE. As will be discussed later in this paper, this estimate of the EFE was not very fruitful empirically. To explore the range of alternatives, we also experimented with the following definitions, which contain relatively minor perturbations of the epistemic value term:
\begin{align*}
G^1_{t+1}(a_t) &= H[Q_{\phi_s}(s_{t+1})] - H[P_{\theta_s}(s_{t+1}|s_t =\hat{s}_t, a_t)] - \psi r_{t+1},\\
G^2_{t+1}(a_t) &= H[P_{\theta_s}(s_{t+1}|s_t =\hat{s}_t, a_t)] - H[Q_{\phi_s}(s_{t+1})] - \psi r_{t+1},\\
G^3_{t+1}(a_t) &= \kl{Q_{\phi_s}(s_{t+1})}{P_{\theta_s}(s_{t+1}|s_t = \hat{s}_t, a_t)} - \psi r_{t+1},
\end{align*}
where all the entropy terms and the KL-divergence were computed analytically. Also, we experimented with simply predicting the (negative) expected future reward as follows:
\begin{align}
G^4_{t+1}(a_t) &= - \psi r_{t+1},\label{eq:reward_maximisation}
\end{align}
which is effectively making the job of the critic identical to the job of the Q-network in the DQN agent (c.f. Section \ref{ssec:dqn} for details). More precisely, they are identical, except for the fact that the Q-network is taking observations as input, while the critic takes hidden states. 

\subsection{Deep active inference} \label{ssec:DAI}

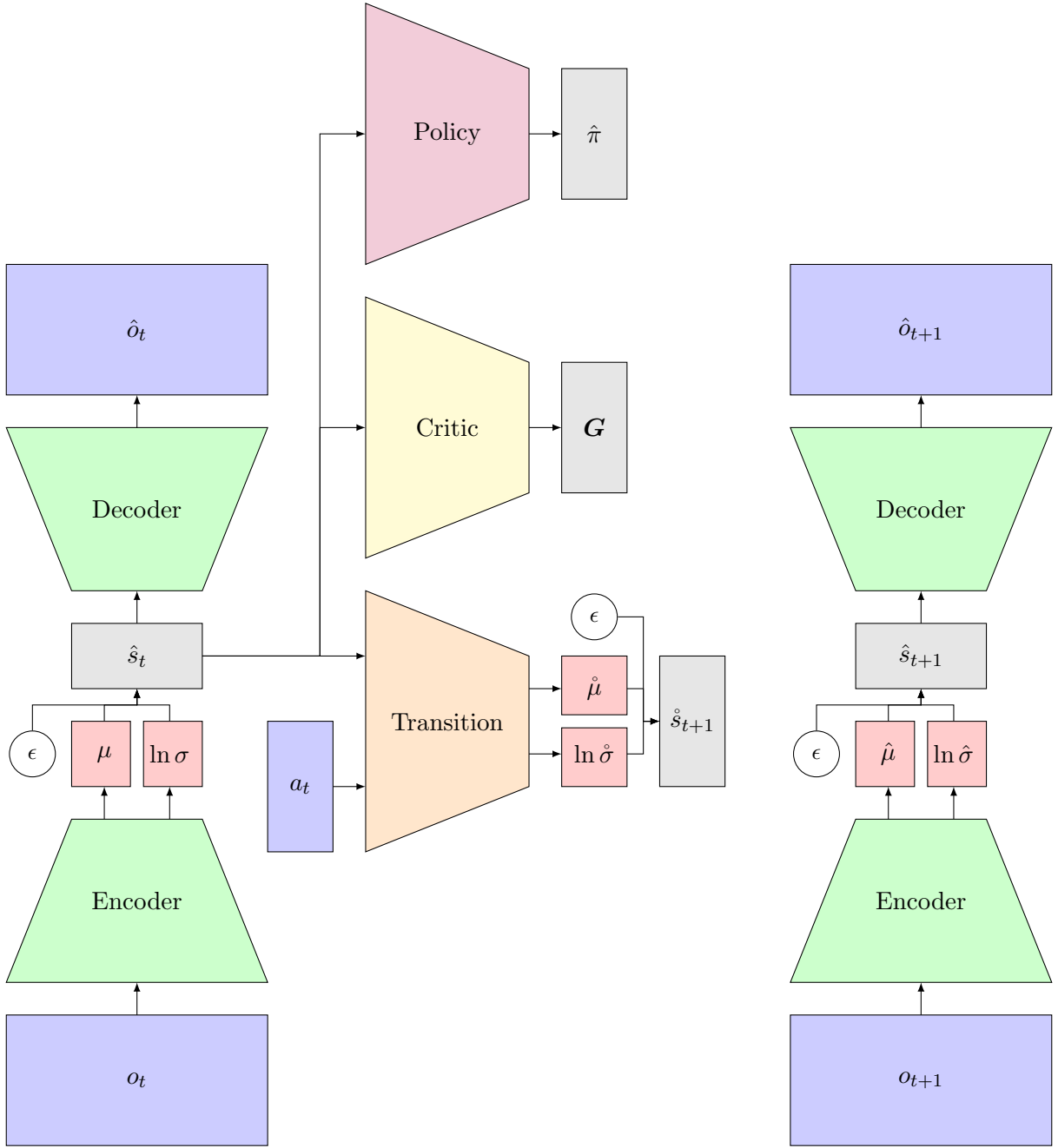
\begin{figure}[H]
	\begin{center}
		\begin{tikzpicture}[square/.style={regular polygon,regular polygon sides=4}]
			\coordinate (A) at (13.5,12.5);
			\coordinate (B) at (16,13.5);
			\coordinate (C) at (16,15.5);
			\coordinate (D) at (13.5,16.5);
			\draw[fill=purple!20!white] (A) -- coordinate[pos=.6] (AB) (B)--(C)
			--coordinate[pos=.45] (CD) (D)
			--coordinate[pos=.55] (DA) cycle;
			\node at (14.75, 14.5) {Policy};
			\draw[-latex] (11,6.5) -- (12.8,6.5) -- (12.8,14.5) -- (13.5,14.5);
			\draw[-latex] (16, 14.5) -- (16.5, 14.5);
			\draw[fill=gray!20!white] (16.5,13.5) rectangle (17.5,15.5);
			\node at (17, 14.5) {$\hat{\pi}$};

			\coordinate (A) at (13.5,8);
			\coordinate (B) at (16,9);
			\coordinate (C) at (16,11);
			\coordinate (D) at (13.5,12);
			\draw[fill=yellow!20!white] (A) -- coordinate[pos=.6] (AB) (B)--(C)
			--coordinate[pos=.45] (CD) (D)
			--coordinate[pos=.55] (DA) cycle;
			\node at (14.75, 10) {Critic};
			
			\draw[-latex] (11,6.5) -- (12.8,6.5) -- (12.8,10) -- (13.5,10);
			\draw[-latex] (16, 10) -- (16.5, 10);
			
			\draw[fill=gray!20!white] (16.5,9) rectangle (17.5,11);
			\node at (17, 10) {$\bm{G}$};
			
			\pic{hmm};
		\end{tikzpicture}
	\end{center}
	\caption{This figure illustrates the DAI agent. The only new part is the policy network, which takes as input the hidden state at time $t$ and ouputs the parameters $\hat{\pi}$ of the variational posterior over actions. Importantly, the DAI takes actions based on the EFE.}
	\label{fig:DAI}
\end{figure}

In this section, we discuss the full deep active inference (DAI) agent illustrated in Figure \ref{fig:DAI}. Put simply, this agent is composed of five deep neural networks, i.e., the encoder, the decoder, the transition, the critic and the policy network. The decoder $\mathcal{D}_{\theta_o}$ models $P_{\theta_o}(o_\tau|s_\tau)$ as a product of Bernoulli distributions, i.e., $P_{\theta_o}(o_\tau|s_\tau) = \MultiBernoulli(o_\tau;\hat{o}_\tau)$ where $\hat{o}_\tau = \mathcal{D}_{\theta_o}(s_\tau)$. The transition network $\mathcal{T}_{\theta_s}$ models $P_{\theta_s}(s_{\tau+1}|s_\tau,a_\tau)$ as a Gaussian distribution, i.e., $P_{\theta_s}(s_{\tau+1}|s_\tau,a_\tau) = \mathcal{N}(s_{\tau+1}|\mathring{\mu},\mathring{\sigma})$ where $\mathring{\mu},\ln \mathring{\sigma} = \mathcal{T}_{\theta_s}(s_\tau,a_\tau)$. The critic $\mathcal{G}_{\theta_a}$ outputs a vector containing the predicted expected free energy of each action, which is used to define the prior over action as $P_{\theta_a}(a_\tau|s_\tau) = \sigma[-\zeta \mathcal{G}_{\theta_a}(s_\tau, \bigcdot\,)]$, where $\sigma[\bigcdot]$ is a softmax function and $\zeta$ is the precision of the prior over actions. With this in mind, the full generative model of the agent is:
\begin{align*}
P_{\theta}(o_{o:T}, s_{o:T}, a_{o:T-1}) = P(s_0)\prod_{\tau = 0}^T P_{\theta_o}(o_\tau|s_\tau) \prod_{\tau = 0}^{T-1} P_{\theta_s}(s_{\tau+1}|s_\tau, a_\tau) P_{\theta_a}(a_\tau|s_\tau),
\end{align*}

\noindent where $P(s_0) = \mathcal{N}(s_0;\mu_0,\sigma_0)$ is a Gaussian prior over initial hidden states. Let $t$ be the present time step. The DAI agent maintains posterior beliefs over the present states $s_t$ and action $a_t$. The variational posterior over states $Q_{\phi_s}(s_t)$ is a Gaussian distribution modelled by the encoder $\mathcal{E}_{\phi_s}$, i.e., $Q_{\phi_s}(s_t) = \mathcal{N}(s_t;\mu, \sigma)$ where $\mu, \ln\sigma = \mathcal{E}_{\phi_s}(o_t)$. The variational posterior over actions $Q_{\phi_a}(a_t|s_t)$ is a categorical distribution modelled by the policy network $\mathcal{P}_{\phi_a}$, i.e., $Q_{\phi_a}(a_t|s_t) = \text{Cat}(a_t;\hat{\pi})$ where $\hat{\pi} = \mathcal{P}_{\phi_a}(s_t)$. The full variational distribution is therefore defined as:
\begin{align*}
Q_{\phi}(a_t,s_t) = Q_{\phi_a}(a_t|s_t)Q_{\phi_s}(s_t).
\end{align*}
The variational free energy of the DAI agent is derived in a similar way to the VFE of Section \ref{ssec:derive_vfe_in_fountas}, and is defined as:
\begin{align}
Q^*_{\phi}(s_t, a_t) &= \argmin_{Q_{\phi}(s_t, a_t)}\underbrace{\kl{Q_{\phi}(s_t, a_t)}{P_{\theta_o}(o_t|s_t)P_{\theta_s}(s_t|s_{t-1},a_{t-1})P_{\theta_a}(a_t|s_t)}}_{\text{variational free energy}}\label{eq:vfe_defi_dai}\\
&= \argmin_{Q_{\phi}(s_t, a_t)} \quad\,\,\, \mathbb{E}_{Q_{\phi_s}(s_t)}\big[\kl{Q_{\phi_a}(a_t|s_t)}{P_{\theta_a}(a_t|s_t)}\big] + \kl{Q_{\phi_s}(s_t)}{P_{\theta_s}(s_t|s_{t-1},a_{t-1})}\nonumber\\
&\quad\qquad\qquad - \mathbb{E}_{Q_{\phi_s}(s_t)}[\ln P_{\theta_o}(o_t|s_t)].\nonumber
\end{align}

\noindent The VFE is therefore a function of $s_{t-1}$, $a_{t-1}$, and $o_t$. Both $a_{t-1}$, and $o_t$ can be obtained from the replay buffer, and $s_{t-1}$ can be sampled from the variational distribution predicted by the encoder network when observation $o_{t-1}$ is provided as input. Also, the KL-divergences can be computed analytically, the expectations w.r.t $Q_{\phi_s}(s_t)$ can be approximated using a Monte-Carlo estimate, and the logarithm of the likelihood mapping reduces to the binary cross entropy because $P_{\theta_o}(o_t|s_t)$ is a product of Bernoulli distributions. Thus, all the VFE terms can be estimated, and the encoder, decoder, transition and policy networks can be trained to minimise the VFE using gradient descent. The critic's weights are optimised as in Section \ref{ssec:CHMM} using gradient descent to minimise the smooth L1 norm between the critic's output $\mathcal{G}_{\theta_a}(s_t,\bigcdot\,)$ and the target G-values $y(\,\bigcdot\,)$, i.e., $\text{ SL1}[\mathcal{G}_{\theta_a}(s_t,\bigcdot\,), y(\,\bigcdot\,)],$ where the target G-values are defined as:
$$y(a_t) = G_{t+1}(a_t) + \gamma \mathbb{E}_{Q_{\phi_s}(s_{t+1})}\Big[ \max_{a_{t+1} \in \mathcal{A}} \hat{\mathcal{G}}_{\hat{\theta}_a}(s_{t+1}, a_{t+1})\Big],$$
where $Q_{\phi_s}(s_{t+1})$ can be computed by feeding the image $o_{t+1}$ sampled from the replay buffer as input to the encoder, $\gamma$ is a discount factor, and $G_{t+1}(a_t)$ is the expected free energy received at time $t+1$ after taking action $a_t$, i.e.,
\begin{align}
G_{t+1}(a_t) &\approx \kl{P_{\theta_s}(s_{t+1}|s_t = \hat{s}_t, a_t)}{Q_{\phi_s}(s_{t+1})} - \psi r_{t+1},
\end{align}
where $\hat{s}_t \sim Q_{\phi_s}(s_t)$, $\psi$ is a hyperparamter modulating the precision of the prior preferences, and $r_{t+1}$ is the reward obtained at time step $t+1$. Note, we also experimented with other definitions of the EFE at time $t + 1$ as presented in Section \ref{ssec:efe_other_defe}. Finally, with regard to the action selection performed by the DAI agent, there are at least four possibilities: (i) select a random action, (ii) select the action with the highest posterior probability according to the policy network, i.e., $a_t^* = \argmax_{a_t} Q_{\phi_a}(a_t|s_t)$, (iii) sample an action from the posterior over actions, i.e., $a_t^* \sim Q_{\phi_a}(a_t|s_t)$, and (iv) use the $\mathring{\epsilon}$-greedy algorithm with exponential decay, i.e., random action with probabilty $\mathring{\epsilon}$ or best action with probability $1 - \mathring{\epsilon}$.

\section{Results} \label{sec:results}

In this section, we discuss the results obtained by the DQN agent and each model presented in Section \ref{sec:build_dai} at solving the dSprites problem. The code that can be used to reproduce all the experiements can be found on GitHub at the following URL: \url{https://github.com/ChampiB/Challenges_Deep_Active_Inference}. Section \ref{ssec:dqn_results} presents the results obtained by the DQN agent. Section \ref{ssec:vae_results} presents the VFE obtained by the VAE agent, and shows the reconstructed images produced by the VAE. Section \ref{ssec:hmm_results} shows the VFE of the HMM agent as well as the generated sequences of images. Section \ref{ssec:chmm_results} illustrates the VFE obtained by the CHMM as well as the reward obtained by this model when using different action selection schemes and different definitions for the EFE. Finally, Section \ref{ssec:dai_results} dicusses the VFE obtained by the DAI agent, as well as the rewards obtained by this model. Note, each time CKA is used in the following sections, we sampled 5K data examples, and we used them to compute all the CKA scores. 

\subsection{DQN agent} \label{ssec:dqn_results}

\begin{figure}[H]
	\begin{center}
    \resizebox{0.8\textwidth}{!}{%
    \begin{tikzpicture}[square/.style={regular polygon,regular polygon sides=4}, scale=0.75]
	    \node at (0, 0) {\includegraphics[scale=0.3]{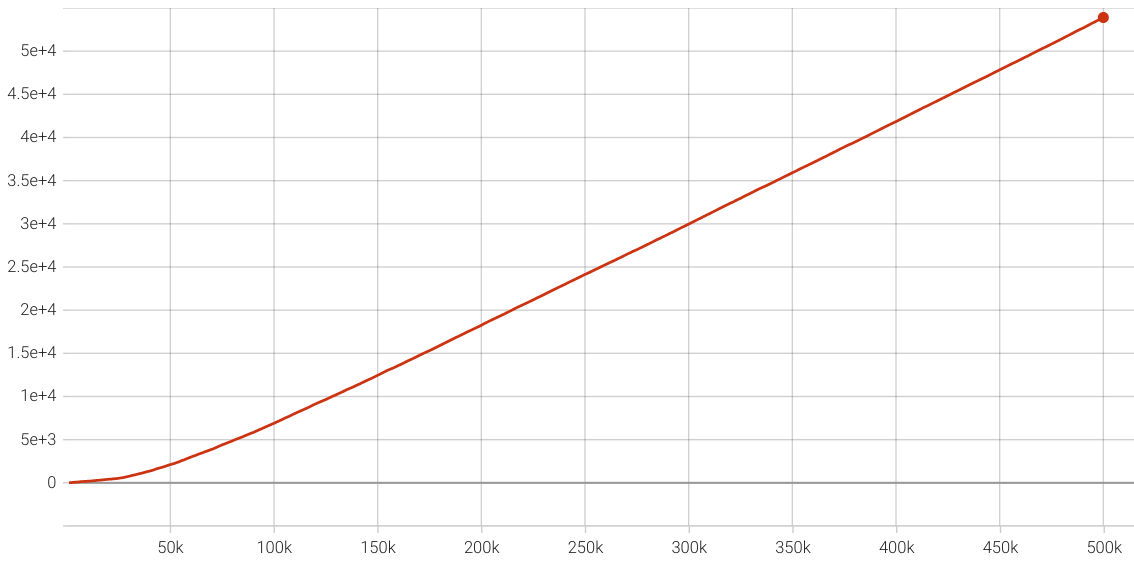}};
	    \node[gray!60!black,rotate=90] at (-8.3, 0.3) {Total Reward};
	    \node[gray!60!black] at (0, -4.4) {Training Iterations};
	\end{tikzpicture}
	}%
	\end{center}
   \caption{This figure illustrates the cumulated rewards obtained by the DQN agent during the 500K training iterations.}
   \label{fig:DQN_rewards}
\end{figure}

\begin{figure}[H]
    \centering
    \includegraphics[width=0.45\linewidth]{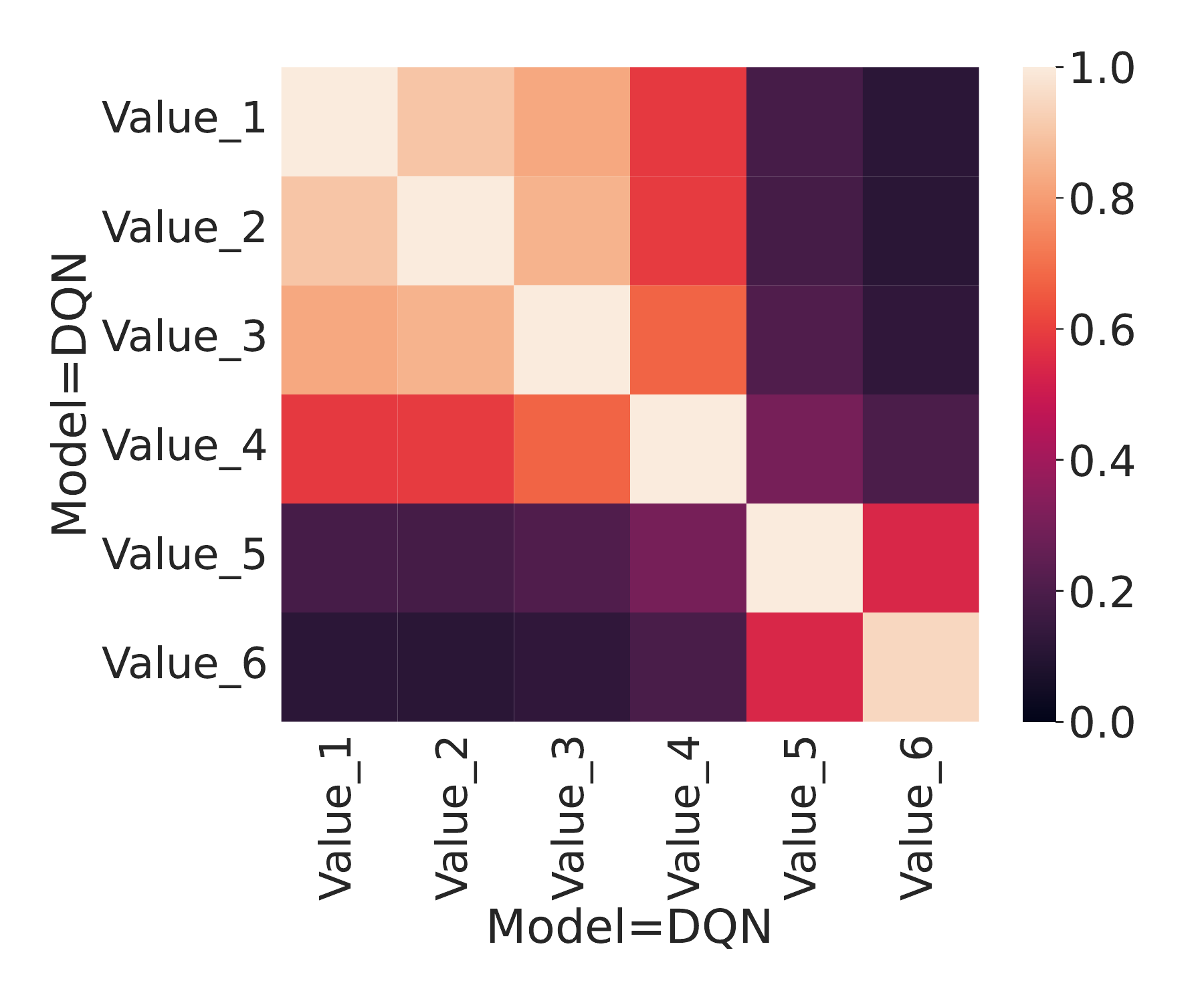}
    \caption{Value\_X labels the X-th layer of the value network, i.e., Value\_1 is the closest to the input and Value\_6 is the output layer. We can see that the first three layers of the DQN learn very similar representations (CKA is close to 1). The representations learned by the fourth layer start to diverge (CKA is lower), and the last two layers learn highly specific representations that are very different from the previous layers (CKA is close to 0), but slightly similar to each other.}\label{fig:cka-dqn}
\end{figure}

In this section, we report the results obtained from the DQN agent. As shown in Figure \ref{fig:DQN_rewards}, the DQN was able to accumulate a total amount of reward of around 50K. This result confirms the correctness of our implementation, and gives us a baseline which can be used to evaluate the performance of the CHMM and DAI agents. To better understand the representations learned by a DQN, we compute the CKA scores between the activations of its layers. We can see in Figure~\ref{fig:cka-dqn} that while the layers closer to the input retain some similarity with each other, the last two layers learn highly specific representations.

\subsection{VAE agent} \label{ssec:vae_results}

\begin{figure}[H]
	\begin{center}
    \resizebox{0.55\textwidth}{!}{%
	\begin{tikzpicture}[square/.style={regular polygon,regular polygon sides=4}, scale=0.75]
	    \node at (0, 0) {\includegraphics[scale=0.4]{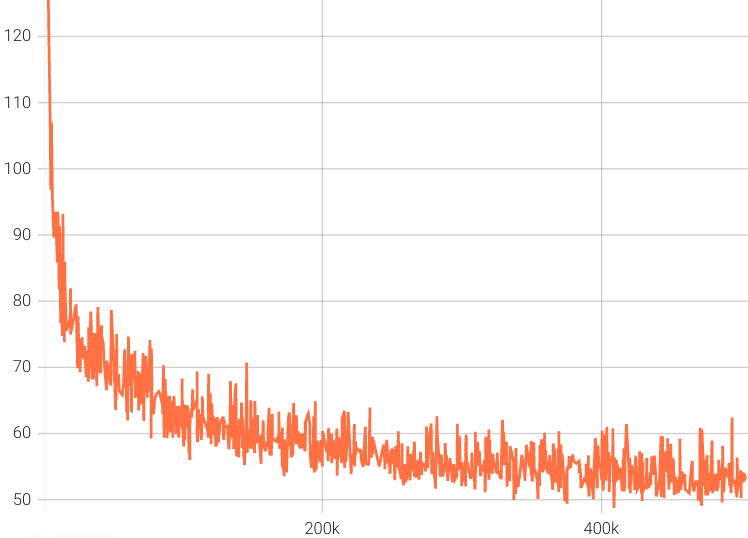}};
	    \node[gray!60!black,rotate=90] at (-7.6, 0) {VFE};
	    \node[gray!60!black] at (0, -5.8) {Training Iterations};
	\end{tikzpicture}
	} %
	\end{center}
   \caption{This figure illustrates the variational free energy of the VAE agent during the 500K iterations of training.}
   \label{fig:VAE_vfe}
\end{figure}

In this section, we report the results obtained by the VAE agent. As shown in Figure \ref{fig:VAE_vfe}, the VFE decreases as training progresses. Also, at the end of the 500K training iterations, the VAE is able to properly reconstruct the  images, c.f., Figure \ref{fig:VAE_reconstruction}. Additionally, since the VAE takes random actions in the environment, the agent was unable to solve the task and accumulated a total amount of reward of around -7K. Those results suggest our implementation is correct, and gives us a baseline for the amount of rewards obtained under random play in the dSprites environment.

\begin{figure}[ht!]
	\centering
	\vspace{-1cm}
	\begin{subfigure}{.4\textwidth}
		\includegraphics[draft=false,width=\linewidth]{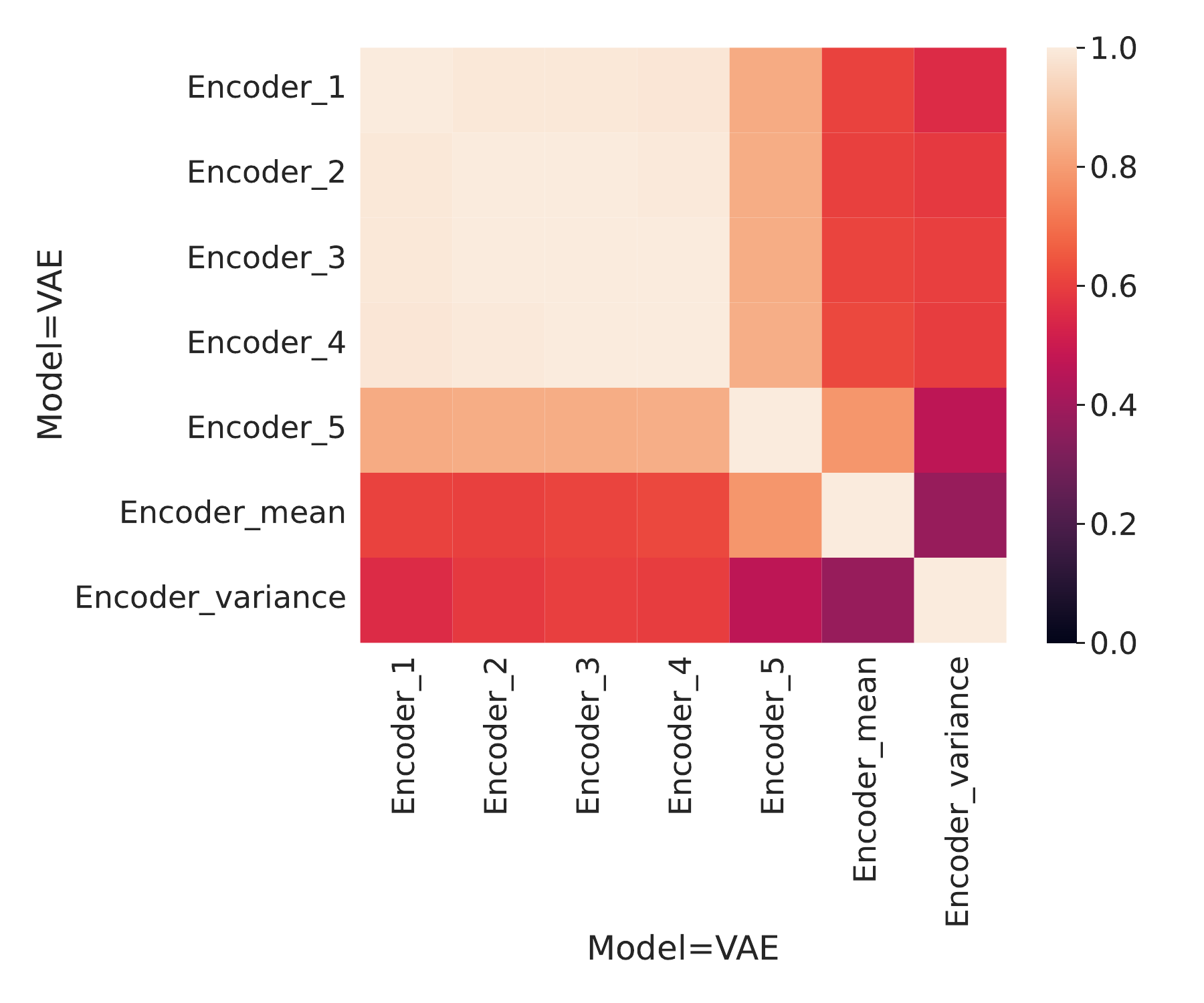}
		\caption{}\label{sfig:cka-vae-vae}
	\end{subfigure}%
	\begin{subfigure}{.4\textwidth}
		\includegraphics[draft=false,width=\linewidth]{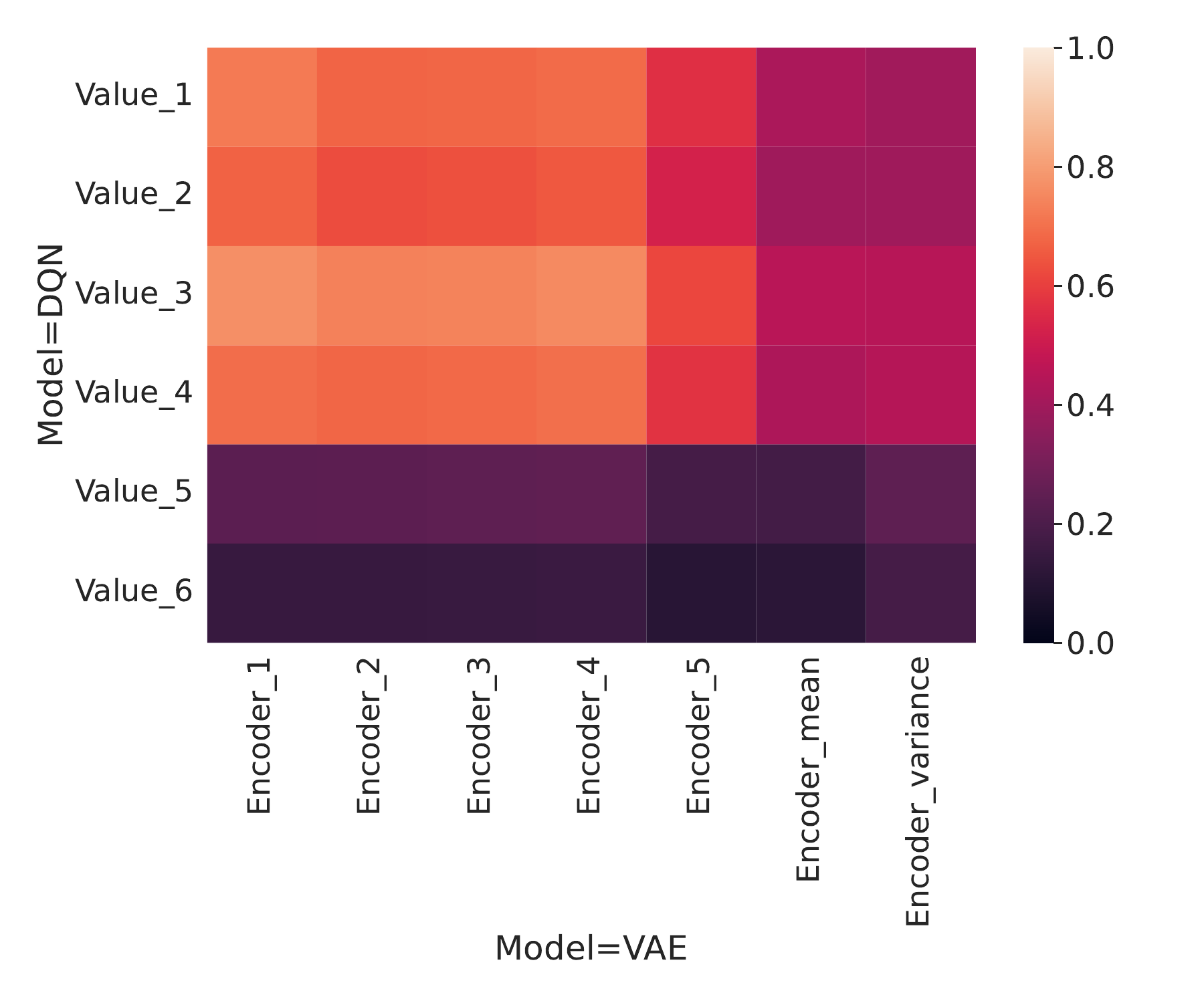}
		\caption{}\label{sfig:cka-vae-dqn}
	\end{subfigure}
	\caption{Encoder\_X is to the X-th layer of the encoder network, i.e., Encoder\_1 is the closest to the input and Encoder\_5 is the one just before the mean and variance layers. Encoder\_mean and Encoder\_variance are the mean and variance layers of the encoder network, respectively. (a) shows the similarity between the representations learned by different layers of the encoder of a VAE.
		(b) shows the similarity between the representations learned by a DQN and a VAE. Note, both the VAE and DQN take images as input and need to process them to either learn a compact representation and reconstruct the images or predict the cumulated reward, respectively. Thus, both learn to represent edges and combination of edges in their first layers.}
	\label{fig:cka-vae}
\end{figure}

\noindent To observe the representations learned by VAEs, we compute the CKA scores between the activations of the different layers of the encoder. We can see in Figure~\ref{sfig:cka-vae-vae} that the representations are strongly similar between all layers, with the exception of the mean and variance representations at the output end (last two layers), similarly to what was observed in~\citet{Bonheme2022}, and is therefore expected. As illustrated in Figure~\ref{sfig:cka-vae-dqn}, these representations are generally similar to those learned by the DQN in early layers but the representations of the last two layers strongly differ, reflecting the difference of learning objectives between the VAE and DQN.

{
\definecolor{Red}{RGB}{188, 9, 9}
\definecolor{Green}{RGB}{60, 128, 56}
\begin{figure}[H]
	\begin{center}
    \resizebox{1\textwidth}{!}{%
    \begin{tikzpicture}[square/.style={regular polygon,regular polygon sides=4}, scale=0.75]
	    \node at (0, 0) {\includegraphics[scale=0.5]{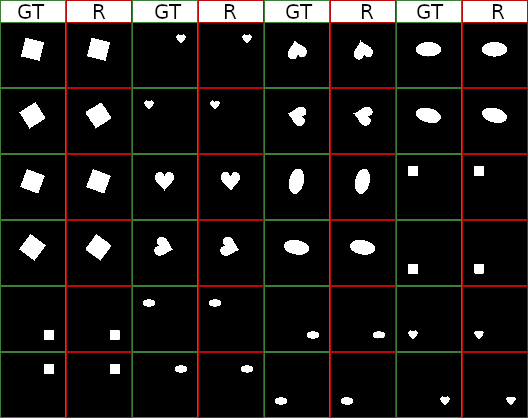}};
	    \node[gray!60!black] at (9, 0) {Ground Truth (GT):};
	    \node[gray!60!black] at (9.1, 0.5) {Reconstruction (R):};
		\draw[fill=Green] (11.5, 0.2) rectangle (12.5, -0.2);
		\draw[fill=Red] (11.5, 0.7) rectangle (12.5, 0.3);
	\end{tikzpicture}
	} %
	\end{center}
   \caption{This figure illustrates the reconstructed image produced by the VAE after 500K training iterations. The columns alternate between the input images and the reconstructed images.}
   \label{fig:VAE_reconstruction}
\end{figure}
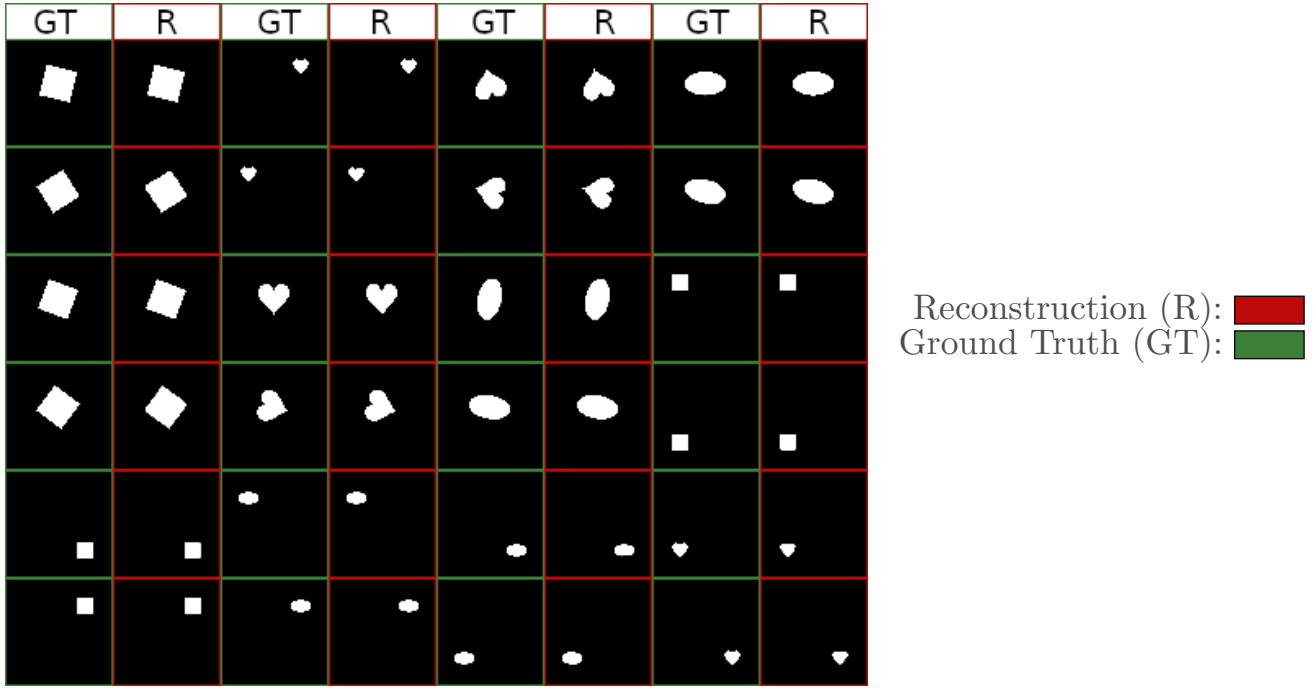
}

\subsection{HMM agent} \label{ssec:hmm_results}

\begin{figure}[H]
	\begin{center}
		\resizebox{0.65\textwidth}{!}{%
			\begin{tikzpicture}[square/.style={regular polygon,regular polygon sides=4}, scale=0.75]
				\node at (0, 0) {\includegraphics[scale=0.56]{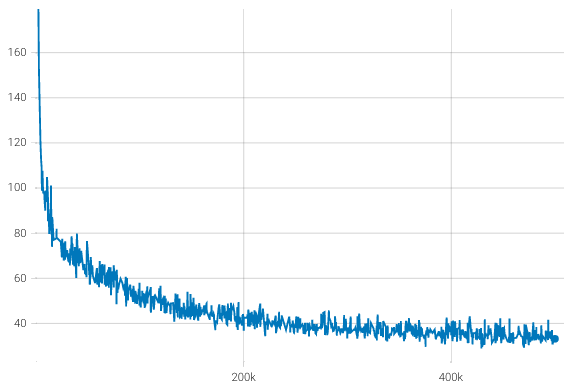}};
				\node[gray!60!black, rotate=90] at (-8.2, 0) {VFE};
				\node[gray!60!black] at (0, -5.8) {Training Iterations};
			\end{tikzpicture}
		} %
	\end{center}
	\caption{This figure illustrates the variational free energy of the HMM agent during the 500K iterations of training.}
	\label{fig:HMM_vfe}
\end{figure}

\begin{figure}[ht!]
	\centering
	\begin{subfigure}{.34\textwidth}
		\centering
		\includegraphics[draft=false,width=\linewidth]{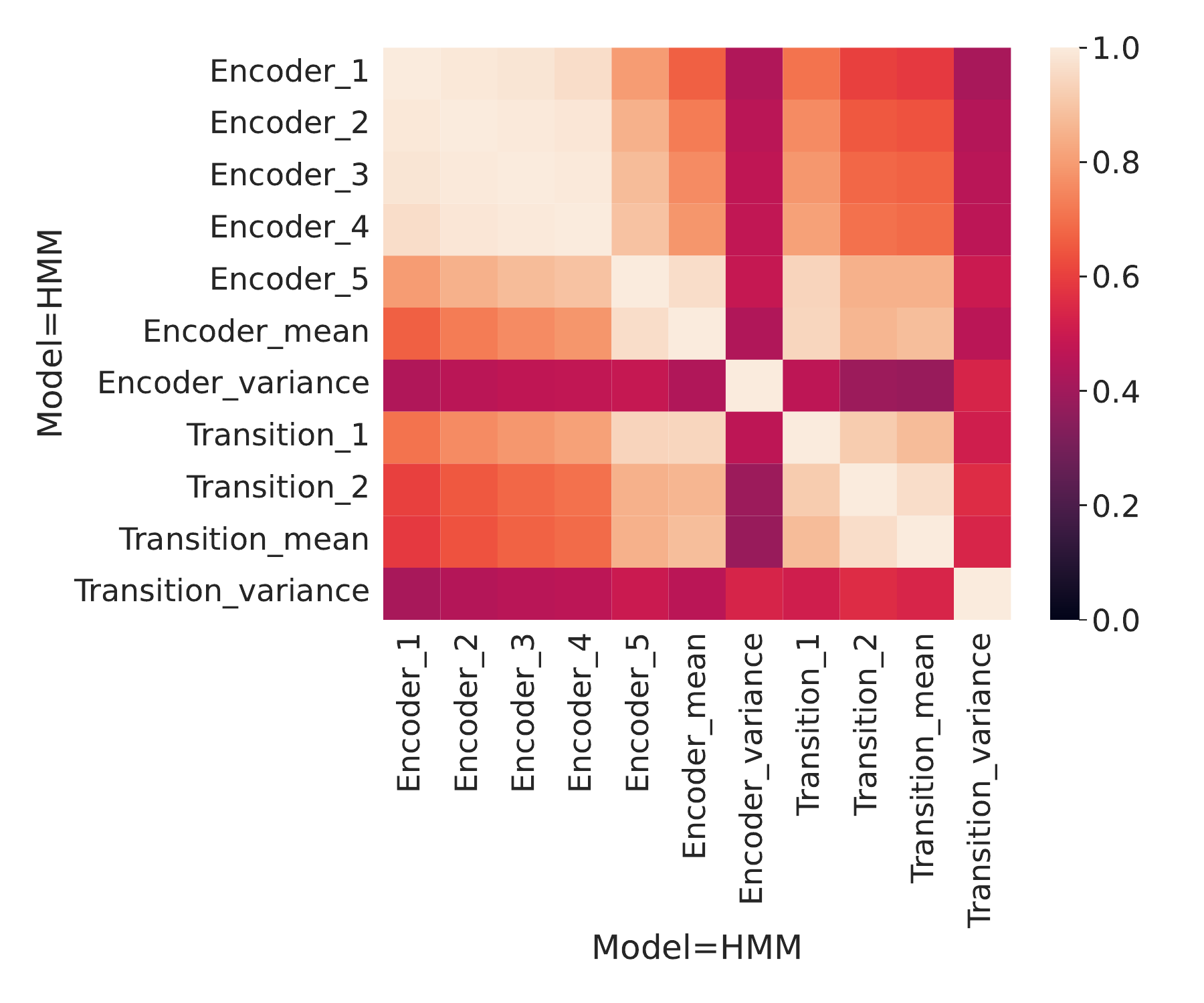}
		\caption{}\label{sfig:cka-hmm-hmm}
	\end{subfigure}%
	\begin{subfigure}{.34\textwidth}
		\centering
		\includegraphics[draft=false,width=\linewidth]{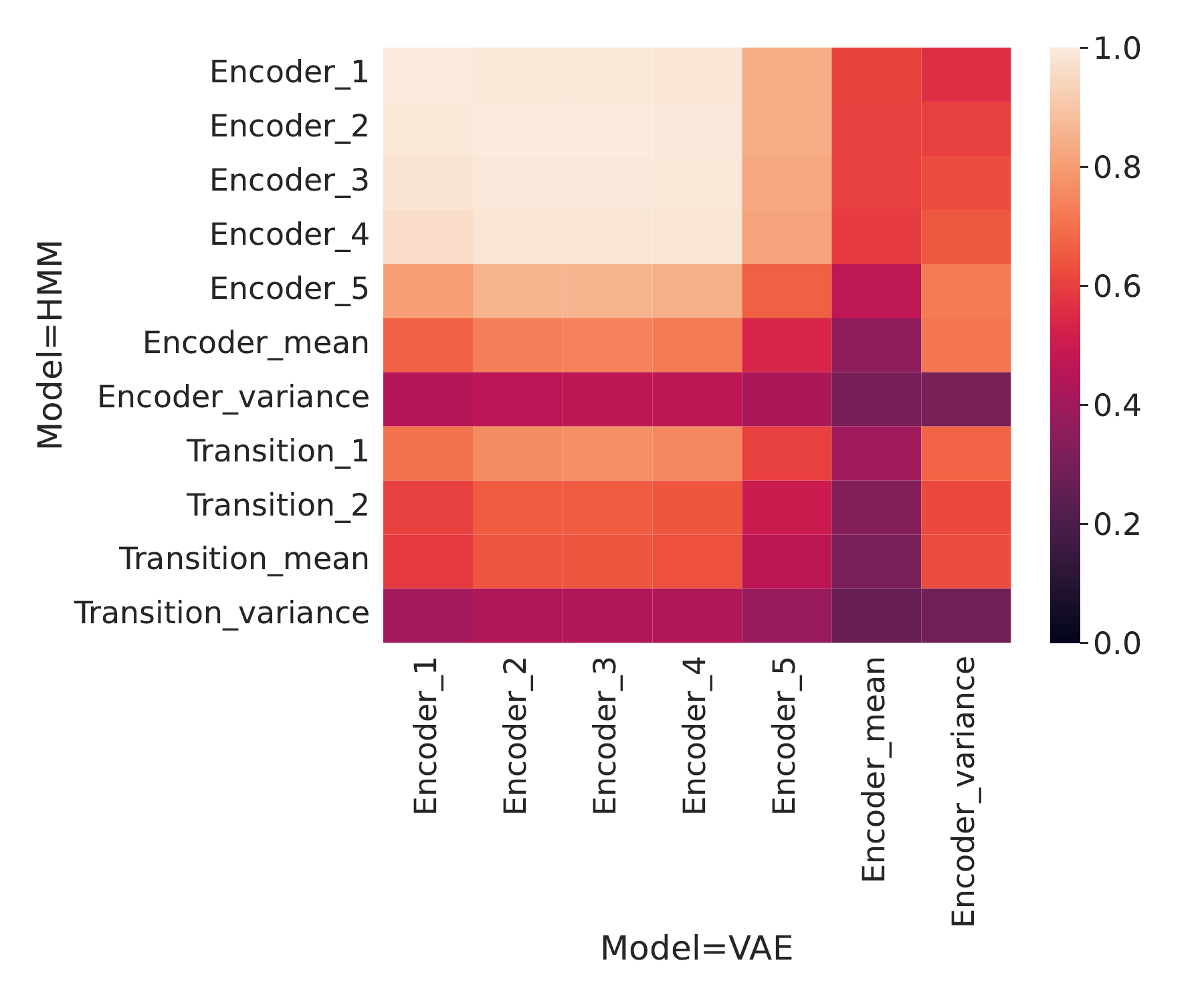}
		\caption{}\label{sfig:cka-hmm-vae}
	\end{subfigure}%
	\begin{subfigure}{.34\textwidth}
		\centering
		\includegraphics[draft=false,width=\linewidth]{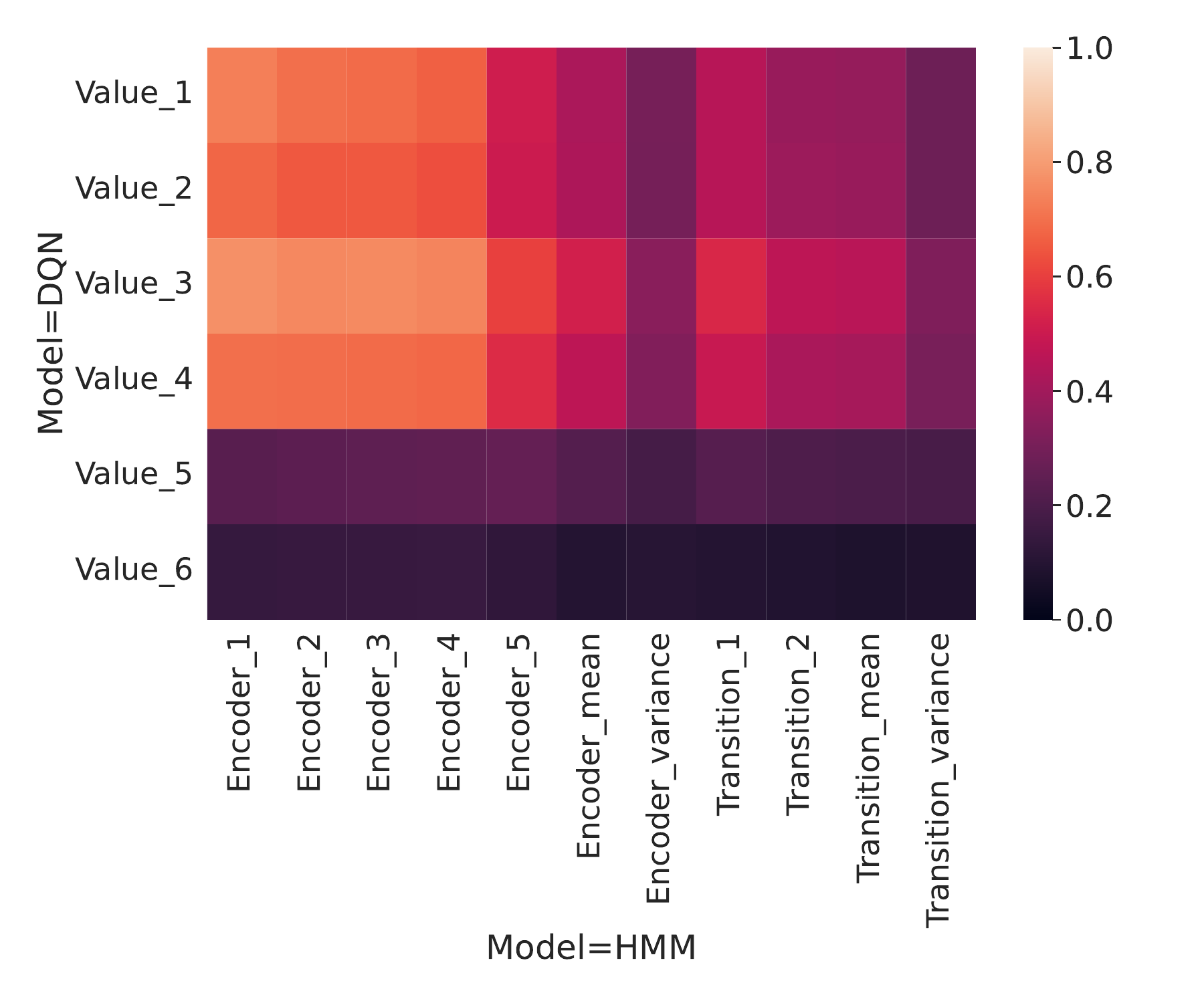}
		\caption{}\label{sfig:cka-hmm-dqn}
	\end{subfigure}
	\caption{Transition\_X is to the X-th layer of the transition network, i.e., Transition\_1 is the closest to the input and Transition\_2 is the one just before the mean and variance layers. Transition\_mean and Transition\_variance are the mean and variance layers of the transition network, respectively. (a) shows the similarity between the representations learned by different layers of the encoder and transition network of an HMM.
		(b) shows the similarity between the representations learned by an HMM and a VAE.
		(c) shows the similarity between the representations learned by a DQN and an HMM.}
	\label{fig:cka-hmm}
\end{figure}

In this section, we report the results obtained by the HMM agent. As shown in Figure \ref{fig:HMM_vfe}, the VFE decreases as training progresses. By comparing Figure \ref{fig:VAE_vfe} and \ref{fig:HMM_vfe}, one can see that the HMM has a lower VFE than the VAE. This is because the agent has more flexibility regarding the prior, i.e., the log-likelihood is the same between the two models but the complexity term is smaller for the HMM than for the VAE. Also, at the end of the 500K training iterations, the HMM is able to properly generate sequences of images, c.f., Figure \ref{fig:HMM_reconstruction}. Additionally, since the HMM takes random actions in the environment, the agent was unable to solve the task and accumulated a total amount of reward of around -7K. These results suggest that our implementation is correct, and comfirm our baseline for the amount of rewards obtained under random play in the dSprites environment.
\newline\\
\noindent Similarly to VAEs, we are interested in observing the representations learned by the encoder of the HMM, and also by its transition network. The representations learned by the encoder of the HMM follow the same trend as those learned by VAEs with an even more marked dissimilarity between the log variance of the HMM and the other representations learned by this model, as illustrated in Figures~\ref{sfig:cka-hmm-hmm} and~\ref{sfig:cka-hmm-vae}. We can further observe in Figure~\ref{sfig:cka-hmm-hmm} that the transition network learns representations similar to the mean representation (Encoder\_mean of HMM) in its first three layers, while the representations learned by the last layer (Transition\_variance) are not similar to any other representations learned by the transition or encoder networks. We can also see in Figure~\ref{sfig:cka-hmm-vae} that the mean and variance representations (Encoder\_mean and Encoder\_variance) learned by HMMs are different from those learned by VAEs, possibly indicating that the transition network influences these two representations. Similarly to VAEs, one can observe in Figure~\ref{sfig:cka-hmm-dqn}, that the representations learned by the variance layers of the encoder and transition networks (Encoder\_variance and Transition\_variance) are very different to the representation learned by the DQN. In contrast, the first four layers of the encoder (Encoder\_1 to Encoder\_4) are similar to the representation learned by the first layers of the DQN (Value\_1 to Value\_4), but are very different from the last two layers (Value\_5 and Value\_6).

{
\definecolor{Red}{RGB}{188, 9, 9}
\definecolor{Green}{RGB}{60, 128, 56}
\begin{figure}[H]
	\begin{center}
    \resizebox{1\textwidth}{!}{%
	\begin{tikzpicture}[square/.style={regular polygon,regular polygon sides=4}, scale=0.75]
	    \node at (0, 0) {\includegraphics[scale=0.3]{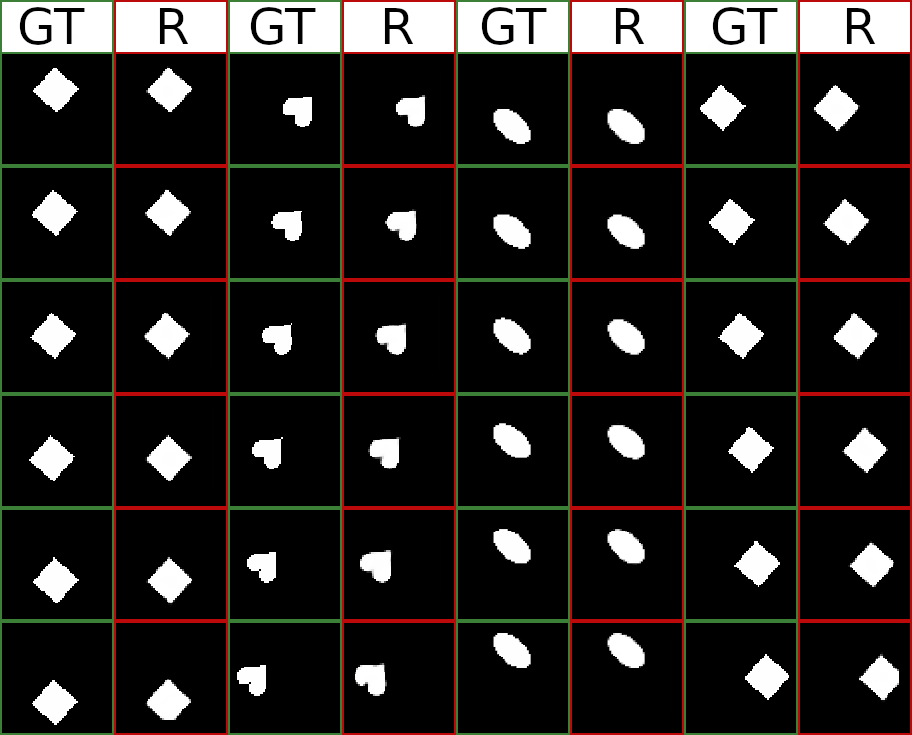}};
	    \node[gray!60!black] at (9, 0) {Ground Truth (GT):};
	    \node[gray!60!black] at (9.1, 0.5) {Reconstruction (R):};
		\draw[fill=Green] (11.5, 0.2) rectangle (12.5, -0.2);
		\draw[fill=Red] (11.5, 0.3) rectangle (12.5, 0.7);
	\end{tikzpicture}
	} %
	\end{center}
   \caption{This figure illustrates the sequences of reconstructed images generated by the HMM after 500K training iterations. The columns alternate between the ground truth images and the reconstructed images. Time passes vertically (from top to bottom), and within each column, the same action is executed repeatedly.}
   \label{fig:HMM_reconstruction}
\end{figure}
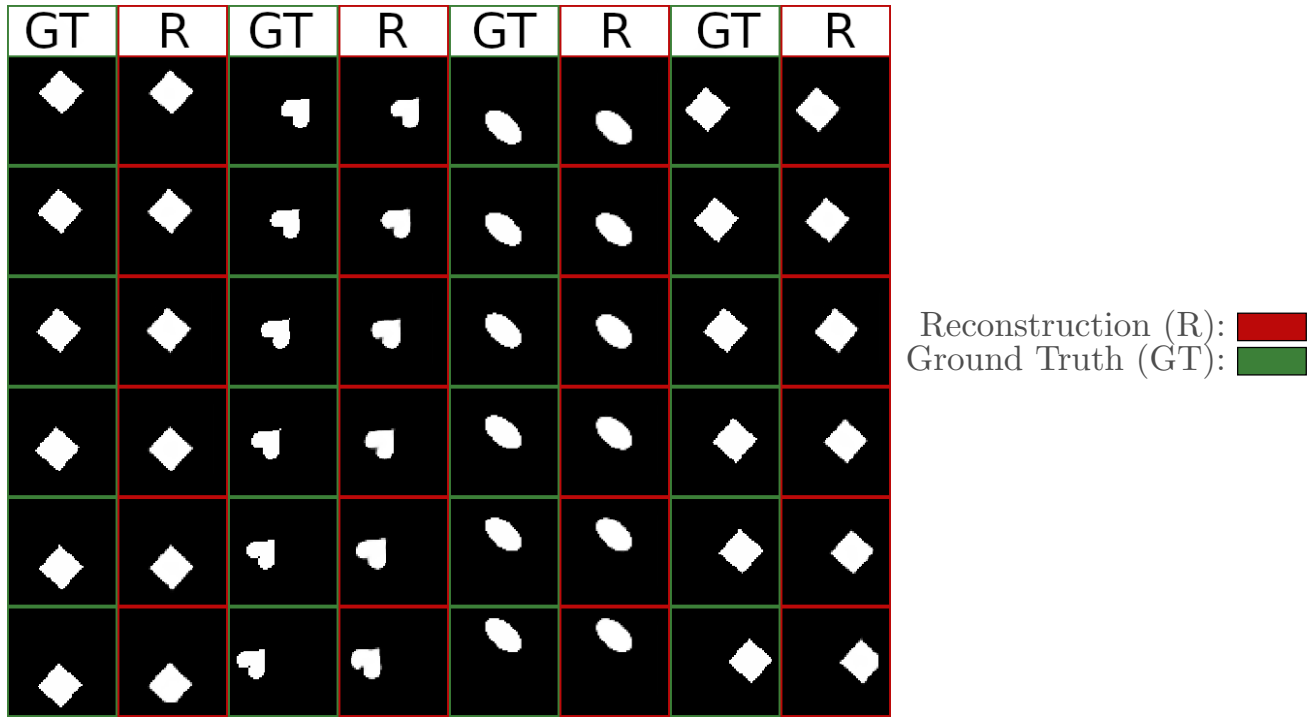
}

\subsection{CHMM agent} \label{ssec:chmm_results}

\begin{figure}[H]
	\begin{center}
		\includegraphics[scale=0.3]{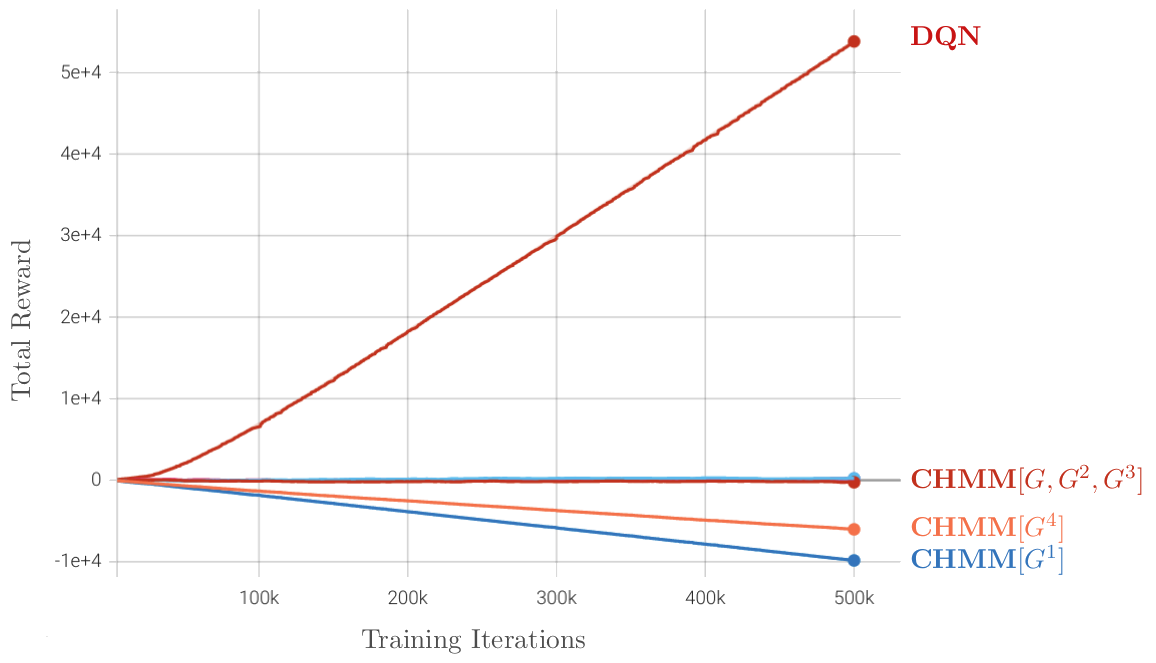}
	\end{center}
   \caption{This figure illustrates the total amount of reward gathered by the CHMM agents (with softmax sampling) during the 500K iterations of training. The only model that was able to solve the task is the DQN agent in red, and all CHMM agents failed.}
   \label{fig:CHMM_rewards_s}
\end{figure}

\begin{figure}[H]
	\begin{center}
	    \includegraphics[scale=0.3]{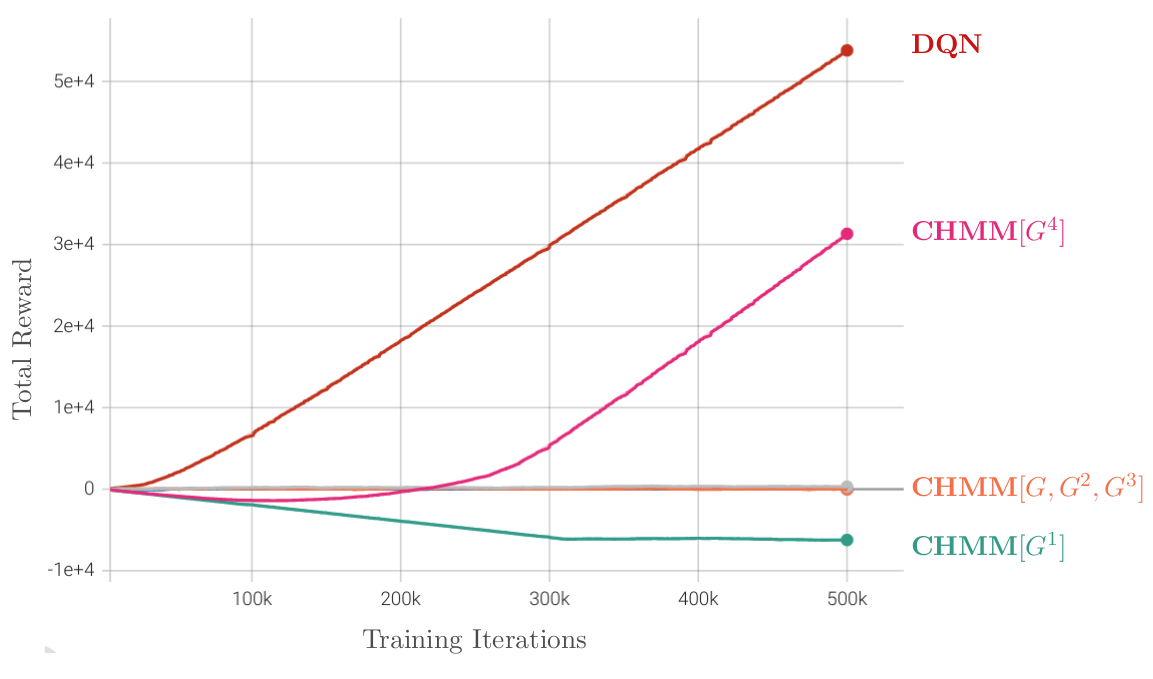}
	\end{center}
   \caption{This figure illustrates the total amount of reward gathered by the CHMM agents (with best action selection) during the 500K iterations of training. The only two models that were able to solve the task are the ones maximising reward (without information gain), i.e., the DQN agent in red and the CHMM whose critic network was predicting only reward in pink.}
   \label{fig:CHMM_rewards_b}
\end{figure}

In this section, we report the results obtained by the CHMM agent, when using different action selection strategies and different definitions of the expected free energy. Figure \ref{fig:CHMM_rewards_eg} presents the cumulated rewards obtained by the CHMM agents using an $\mathring{\epsilon}$-greedy algorithm for action selection, as well as the total rewards obtained by the DQN agent. The critic network of the CHMM agents were trained to predict the five definitions of the expected free energy proposed in Section \ref{ssec:CHMM}. Note, the DQN agent performs better that any of the CHMM agents, and the only agent that solves the task is the CHMM maximising reward only, i.e., without any information gain. Figure \ref{fig:CHMM_rewards_s} presents the same experiement as Figure \ref{fig:CHMM_rewards_eg} except that the CHMM agents were using softmax sampling for action selection. In this setting, none of the CHMM agents were able to solve the task. Finally, Figure \ref{fig:CHMM_rewards_b} presents yet again the same experiements but this time the CHMM agents were selecting the best action according to the critic. In this setting, only the CHMM maximising reward was able to solve the task. Also, by comparing Figures \ref{fig:CHMM_rewards_eg} and \ref{fig:CHMM_rewards_b}, it becomes clear that the CHMM using an $\mathring{\epsilon}$-greedy algorithm performs better than the CHMM selecting the best action according to the critic. Put simply, the latter suffers from a lack of exploration that slows down its learning. 

Additionally, Figure \ref{fig:CHMM_vfe} represents the variational free energy of the CHMM agent using the $\mathring{\epsilon}$-greedy algorithm. All the agents were able to minimise their variational free energy, except the one displayed in orange whose VFE suddenly became equal to NaN (i.e., Not a Number); this agent was minimising the expected free energy as defined by $G^1$. Note, $G^1$ is neither the reward nor the ``principled" expected free energy, $G^1$ is one of definitions that we experimented with to explore alternative definitions. Also, the variational free energy of the CHMM agents using softmax sampling and best action selection are not presented, because their results are very similar to the results shown in Figure \ref{fig:CHMM_vfe}.

\begin{figure}[H]
	\begin{center}
	{
\definecolor{Red}{RGB}{188, 9, 9}
\definecolor{Green}{RGB}{60, 128, 56}
\begin{figure}[H]
	\begin{center}
    \resizebox{0.9\textwidth}{!}{%
	\begin{tikzpicture}[square/.style={regular polygon,regular polygon sides=4}, scale=1]
	    \node at (-0.5, 0) {\includegraphics[scale=0.31]{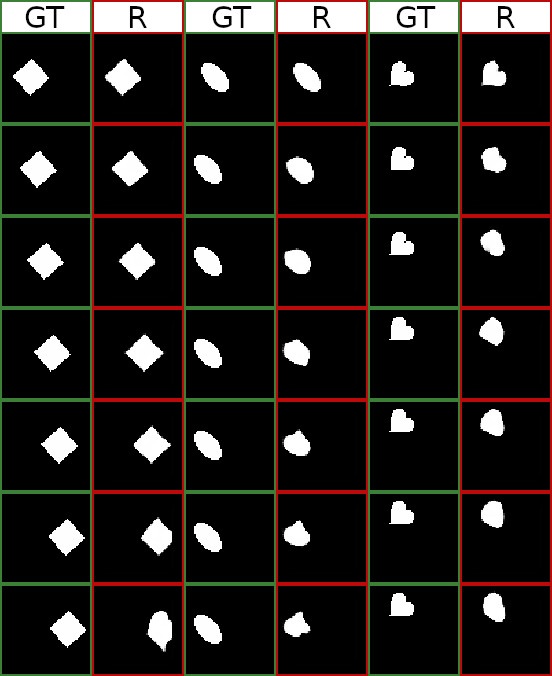}};
	    \node[gray!60!black] at (6, 0) {Ground Truth (GT):};
	    \node[gray!60!black] at (6.1, 0.5) {Reconstruction (R):};
		\draw[fill=Green] (8.5, 0.2) rectangle (9.5, -0.2);
		\draw[fill=Red] (8.5, 0.3) rectangle (9.5, 0.7);
	\end{tikzpicture}
	} %
	\end{center}
\end{figure}
}
	\end{center}
   \caption{This figure illustrates the sequences of reconstructed images generated by a CHMM (maximising reward) after 500K training iterations. The columns alternate between the ground truth images and the reconstructed images. Time passes vertically (from top to bottom), and within each column, the same action is executed repeatedly.}
   \label{fig:CHMM_reconstruction}
\end{figure}

Finally, Figure \ref{fig:CHMM_reconstruction} shows examples of predicted trajectories after a CHMM (maximising reward) was trained. By comparing with Figure \ref{fig:HMM_reconstruction}, we see that the CHMM does not understand the dynamics of the environment as well as the HMM agent. This sugguests a conflict between the two goals of the agent, i.e., maximising reward\footnote{ As shown in Figure \ref{fig:reward_usage}, the reward is used to compute the target values that must be predicted by the critic network.} (or expected free energy) and learning a model of the world. More precisely, Figure \ref{fig:HMM_reconstruction} shows that an HMM agent taking random actions is able to gather a large diversity of training examples and learns the dynamic of the environment beautifully, but does not solve the task. In contrast, the CHMM maximising reward solves the task but learns a poor model of the environment, c.f., Figure \ref{fig:CHMM_reconstruction}.

{
\definecolor{RedCHMM}{RGB}{229, 0, 8}
\definecolor{GrayCHMM}{RGB}{187,187,187}
\definecolor{DarkBlueCHMM}{RGB}{68,116,163}
\definecolor{LightBlueCHMM}{RGB}{92,186,234}
\definecolor{OrangeCHMM}{RGB}{244,112,73}
\begin{figure}[H]
	\begin{center}
		\begin{tikzpicture}[square/.style={regular polygon,regular polygon sides=4}, scale=1]
	    \node at (0, 0) {\includegraphics[scale=0.3]{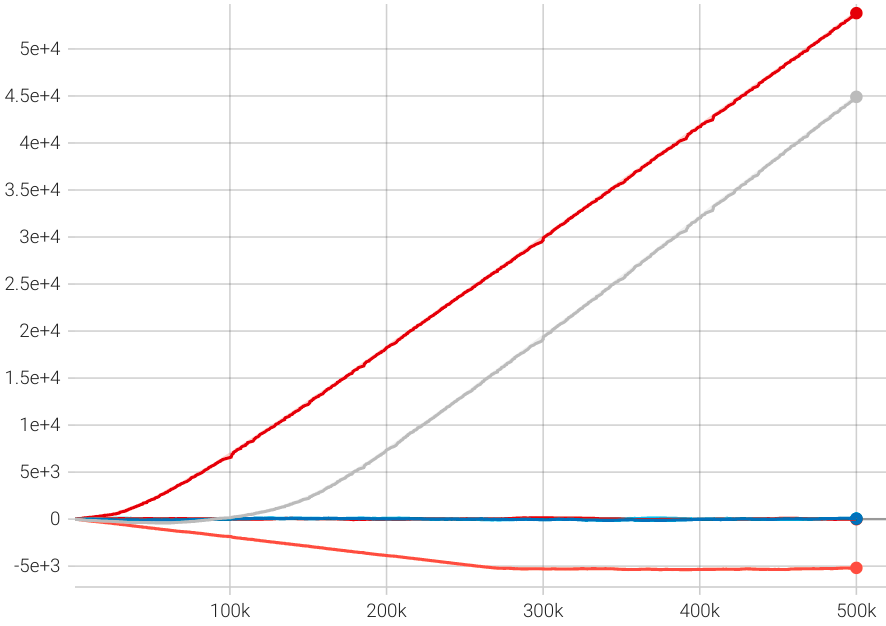}};
	    \node[black] at (0, -3.7) {Training iterations};
	    \node[black, rotate=90] at (-5, 0) {Total Reward};
	    \node[RedCHMM] at (5.5, 3.2) {DQN:};
	    	\node[GrayCHMM] at (6, 2.35) {CHMM[$G^4$]:};
	    \node[DarkBlueCHMM] at (6.7, -2.2) {CHMM[$G$, $G^2$, $G^3$]:};
	    \node[OrangeCHMM] at (6.1, -2.7) {CHMM[$G^1$]:};
	\end{tikzpicture}
	\end{center}
   \caption{This figure illustrates the total amount of reward gathered by the CHMM agents (with $\mathring{\epsilon}$-greedy action selection) during the 500K iterations of training. The only two models that were able to solve the task are the ones maximising reward (without information gain), i.e., the DQN agent in red and the CHMM whose critic network was predicting only reward in gray.}
   \label{fig:CHMM_rewards_eg}
\end{figure}
}

{
\definecolor{RedCHMM}{RGB}{203,61,37}
\definecolor{GrayCHMM}{RGB}{187,187,187}
\definecolor{DarkBlueCHMM}{RGB}{68,116,163}
\definecolor{LightBlueCHMM}{RGB}{92,186,234}
\definecolor{OrangeCHMM}{RGB}{244,112,73}
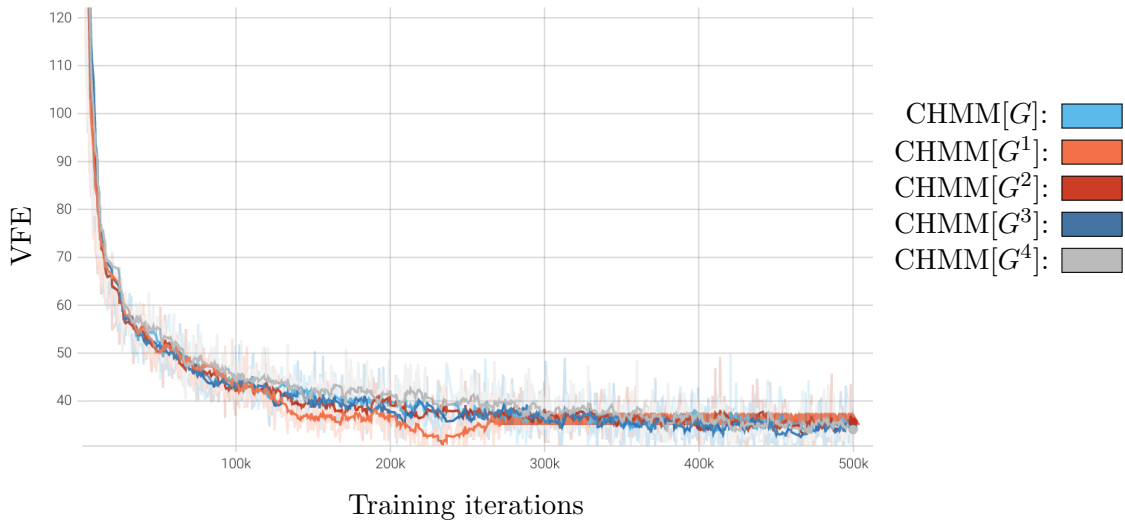
\begin{figure}[H]
	\begin{center}
	\begin{tikzpicture}[square/.style={regular polygon,regular polygon sides=4}, scale=0.8]
	    \node at (-2, 0) {\includegraphics[scale=0.28]{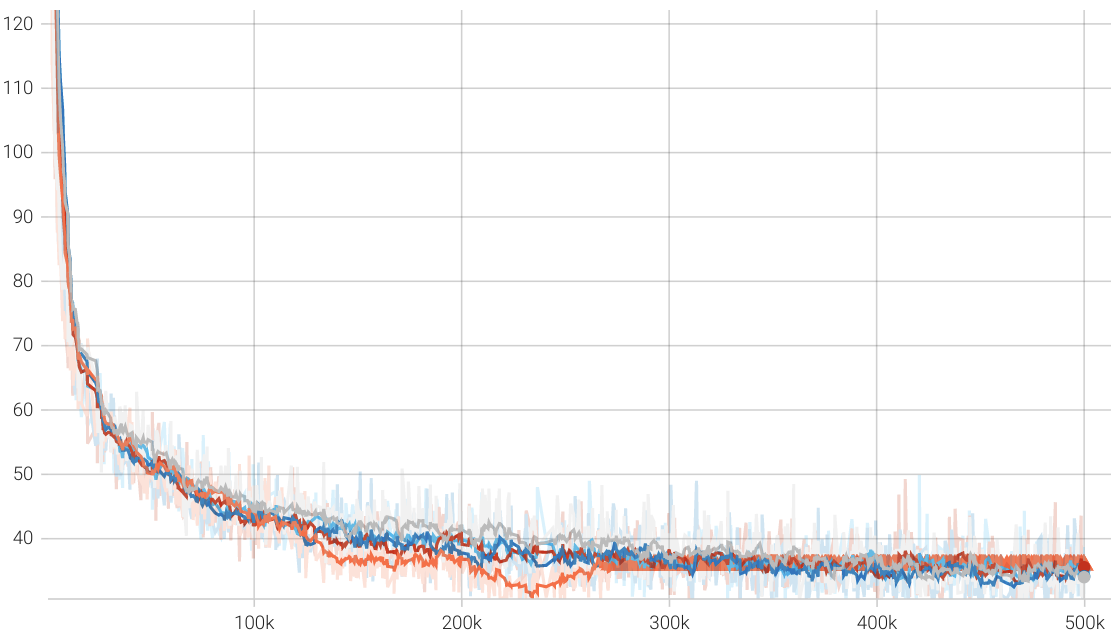}};

	    \node[black] at (6.5, -0.4) {CHMM[$G^4$]:};
		\draw[fill=GrayCHMM] (8, -0.2) rectangle (9, -0.6);

	    \node[black] at (6.5, 0.2) {CHMM[$G^3$]:};
		\draw[fill=DarkBlueCHMM] (8, 0) rectangle (9, 0.4);

	    \node[black] at (6.5, 0.8) {CHMM[$G^2$]:};
		\draw[fill=RedCHMM] (8, 0.6) rectangle (9, 1);

	    \node[black] at (6.5, 1.4) {CHMM[$G^1$]:};
		\draw[fill=OrangeCHMM] (8, 1.2) rectangle (9, 1.6);

	    \node[black] at (6.585, 2) {CHMM[$G$]:};
		\draw[fill=LightBlueCHMM] (8, 1.8) rectangle (9, 2.2);

	    \node[black] at (-1.9, -4.5) {Training iterations};
	    \node[black, rotate=90] at (-9.3, 0) {VFE};
	\end{tikzpicture}
	\end{center}
   \caption{This figure illustrates the variational free energy of the CHMM agents during the 500K iterations of training. All the agents were able to minimise their variational free energy, except the one displayed in orange which crashed; this agent was minimising the expected free energy as defined by $G^1$. More precisely, the variational free energy took the value ``Not a Number" (NaN), which is visible because of the thick horizontal line between 270K and 500K training iterations.}
   \label{fig:CHMM_vfe}
\end{figure}
}

\subsubsection{How do CHMMs learn?}

\begin{figure}[H]
    \centering
    \begin{subfigure}{.33\textwidth}
        \centering
        \includegraphics[draft=false,width=\linewidth]{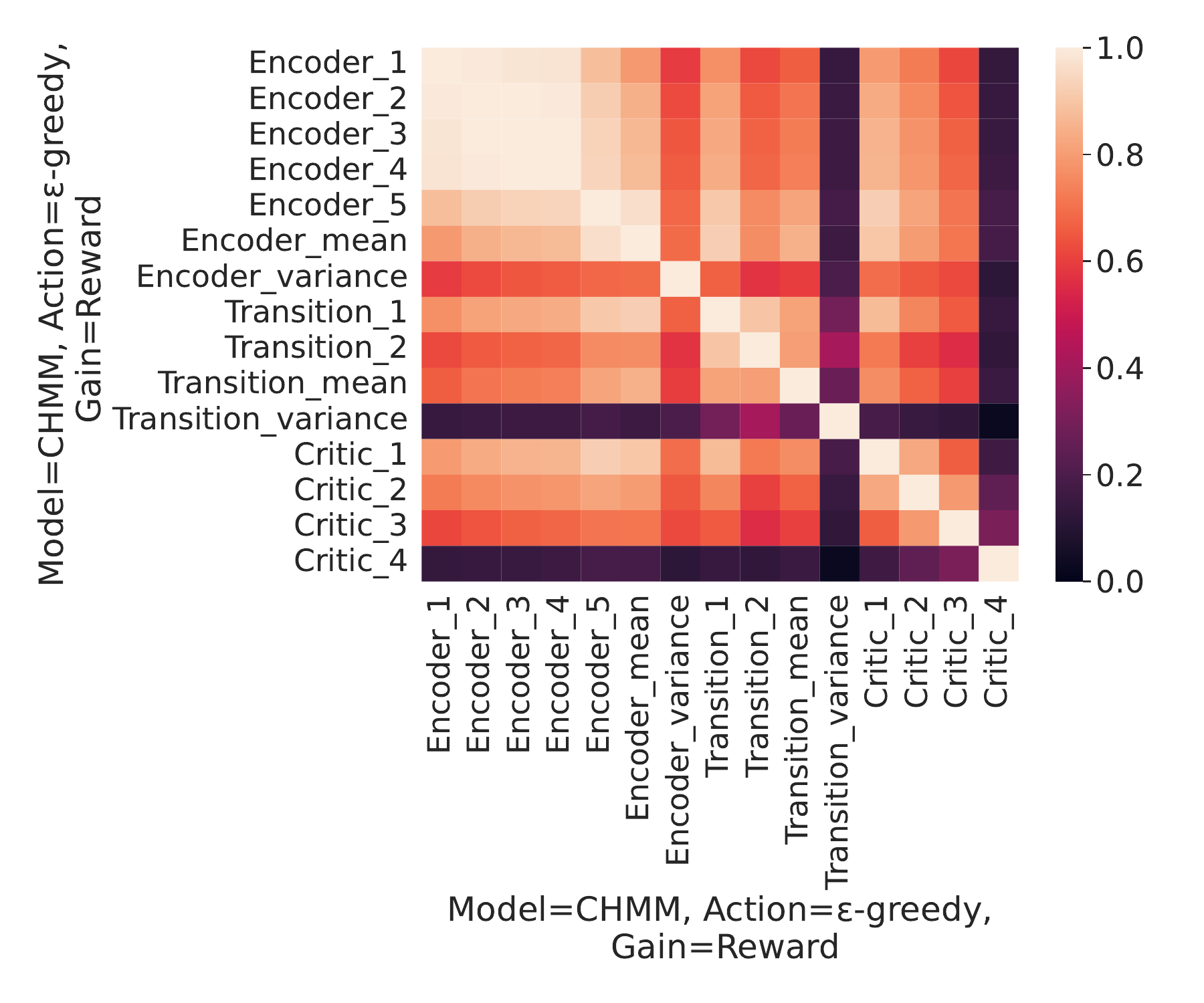}
        \caption{}\label{sfig:cka-chmm-chmm}
    \end{subfigure}%
    \begin{subfigure}{.33\textwidth}
        \centering
        \includegraphics[draft=false,width=\linewidth]{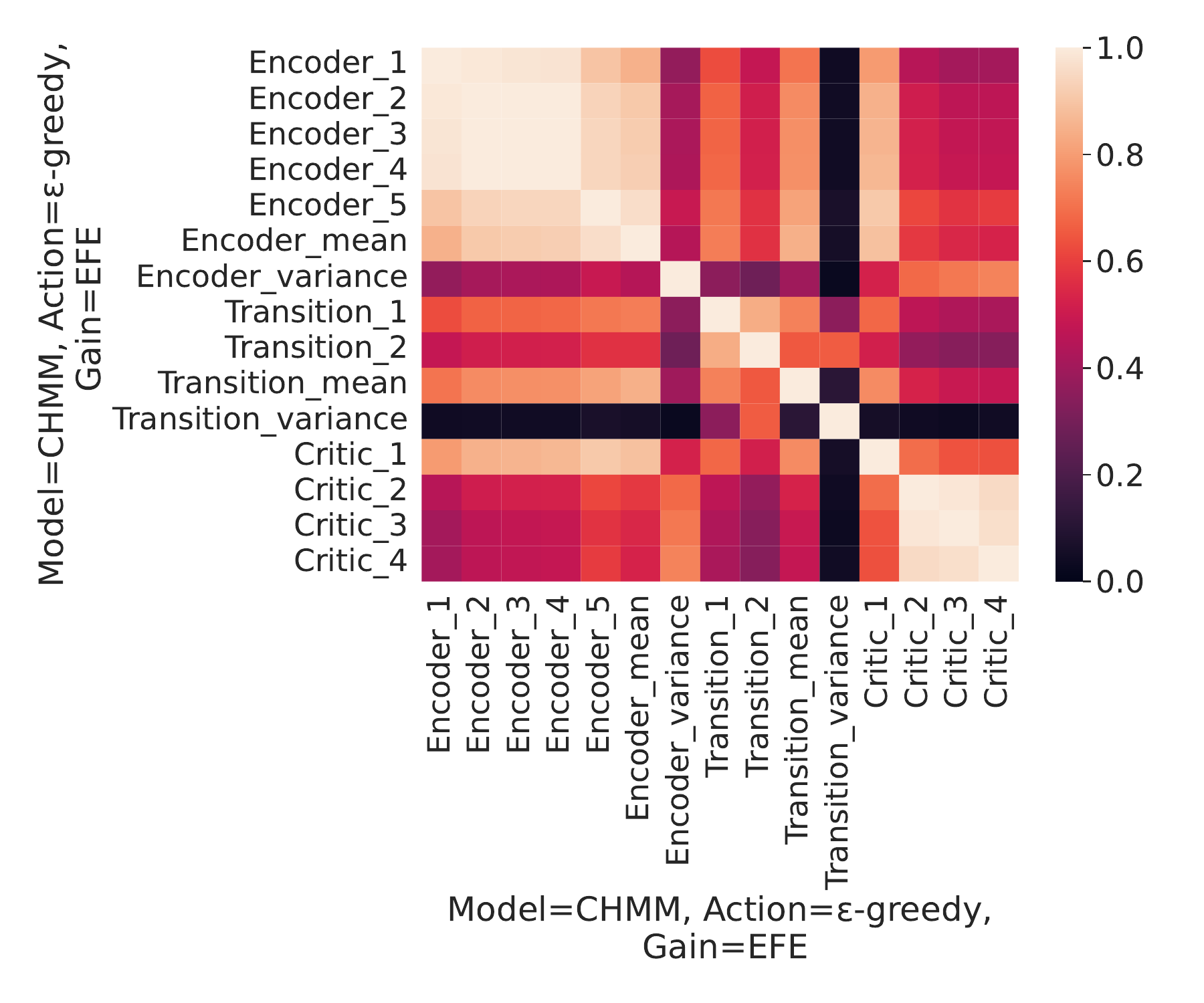}
        \caption{}\label{sfig:cka-chmm2-chmm2}
    \end{subfigure}%
    \begin{subfigure}{.33\textwidth}
        \centering
        \includegraphics[draft=false,width=\linewidth]{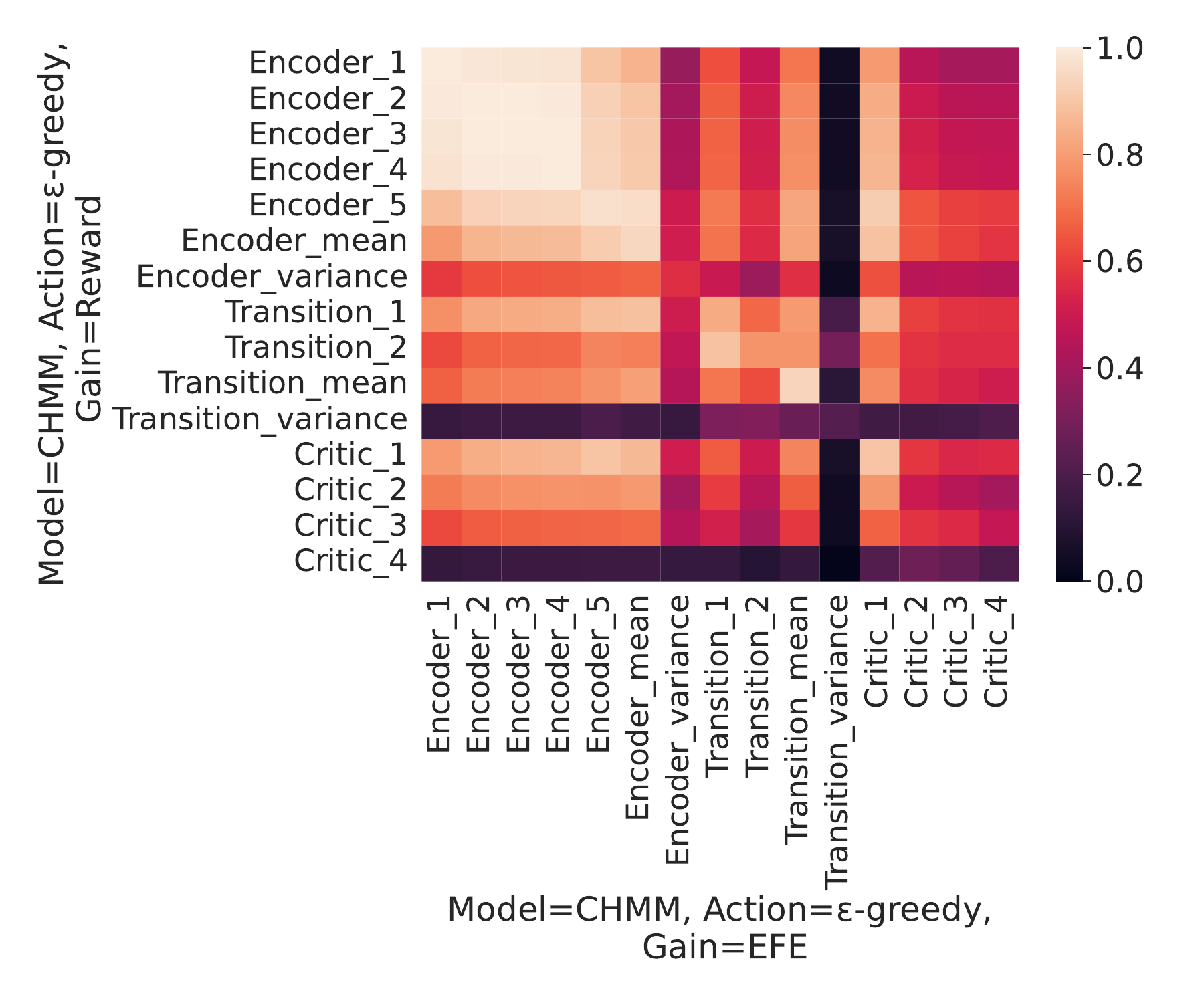}
        \caption{}\label{sfig:cka-chmm-chmm2}
    \end{subfigure}

    \caption{{\small (a) shows the similarity between the representations learned by different layers of the encoder, transition and critic networks of a CHMM whose critic maximises the reward (with $\mathring{\epsilon}$-greedy selection).
        (b) shows the similarity between the representations learned by different layers of the encoder, transition and critic networks of a CHMM whose critic minimises the EFE (with $\mathring{\epsilon}$-greedy selection)
        (c) shows the similarity between the representations learned by two CHMMs, one whose critic optimises EFE and the other optimises reward (both with $\mathring{\epsilon}$-greedy selection).
    }}
    \label{fig:cka-chmm}
\end{figure}

\begin{figure}[H]
	\begin{subfigure}{.3\textwidth}
		\centering
		\includegraphics[draft=false,width=\linewidth]{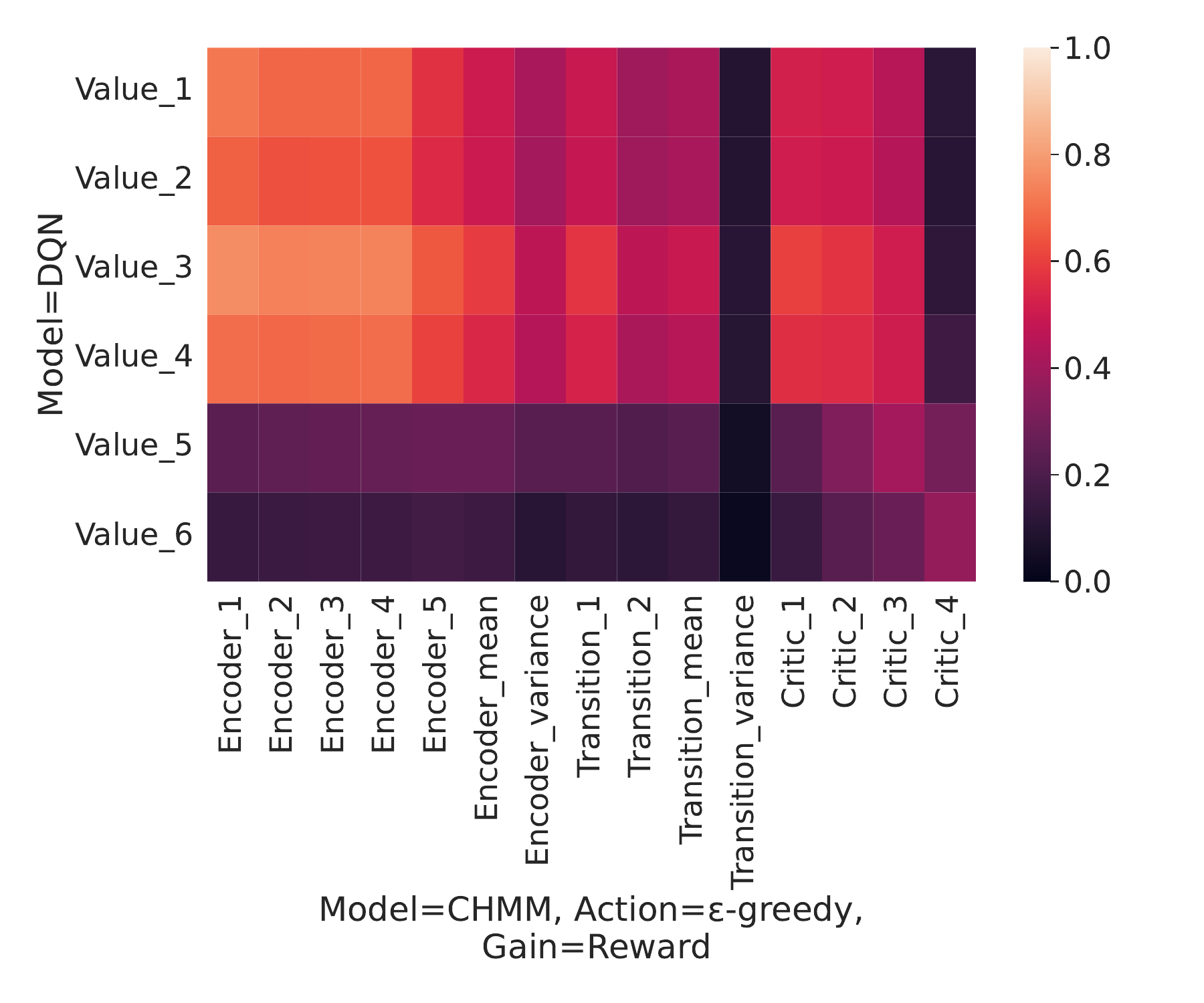}
		\caption{}\label{sfig:cka-dqn-chmm1}
	\end{subfigure}%
	\begin{subfigure}{.3\textwidth}
		\centering
		\includegraphics[draft=false,width=\linewidth]{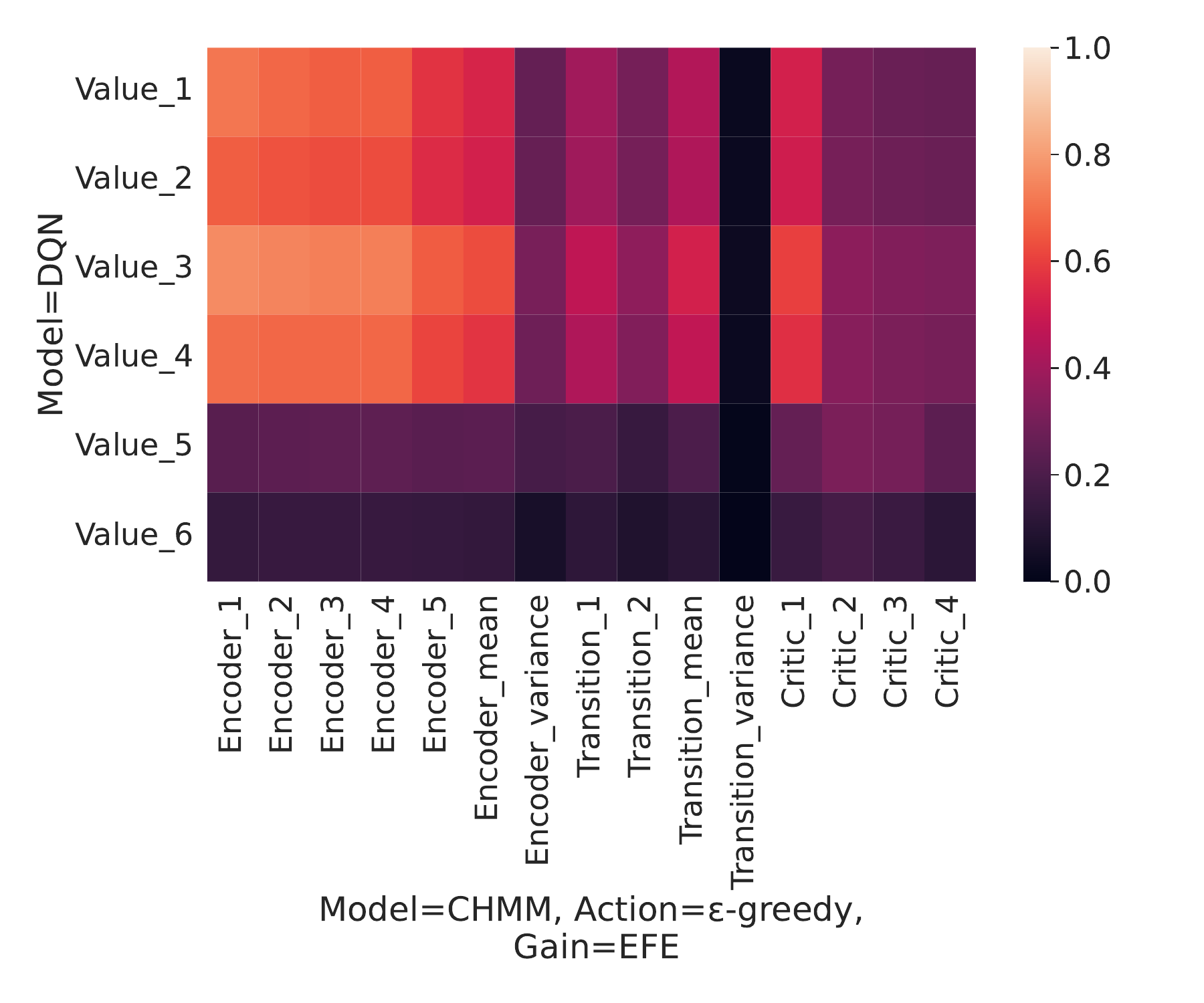}
		\caption{}\label{sfig:cka-dqn-chmm2}
	\end{subfigure}%
	\centering
	\caption{{\small (a) shows the similarity between the representations learned by a CHMM whose critic maximises the reward (with $\mathring{\epsilon}$-greedy selection) and a DQN; (b) shows the similarity between the representations learned by a CHMM whose critic minimises the EFE (with $\mathring{\epsilon}$-greedy selection) and a DQN }}
	\label{fig:cka-dqn-chmm}
\end{figure}

\paragraph{CHMM with $\mathring{\epsilon}$-greedy action selection}
As illustrated in Figure~\ref{fig:CHMM_rewards_eg}, only the CHMM whose critic maximises the reward was able to solve the task. We could thus expect the representations learned by this CHMM to be closer to those learned by the DQN than those learned by the other CHMMs.~We can see in Figure~\ref{fig:cka-dqn-chmm} that the last two layers of the critic network of the CHMM maximising the reward are indeed a bit more similar to the representations of the last two layers of the DQN than the representations learned when the critic is minimising the EFE (see intersection of Value\_5 and Value\_6 with Critic\_3 and Critic\_4, i.e., bottom right corner of the matrix). However, the representations learned by the critic of both CHMMs are still quite different from the last two layers of the DQN (CKA is lower than 0.4, bottom right corner of the matrix, again). Interestingly, the first four layers of the CHMM maximising the reward retain a high similarity with the earlier layers of the DQN ($4\times4$ region at upper left), suggesting some common representations between models. We can further see in Figure~\ref{sfig:cka-chmm-chmm} that the CKA score between the encoder, transition and critic networks is higher or equal to 0.6 (except for the variance layer of the transition and the last layer of the critic), indicating that the transition and critic networks of the CHMM maximising the reward retain some information from the encoder. The information retained by the first three layers of the critic when the CHMM minimises the EFE is much lower, as illustrated in Figures~\ref{sfig:cka-chmm2-chmm2}.
\vspace{-0.5cm}
\vspace{-0.5cm}

\begin{figure}[H]
    \centering
    \begin{subfigure}{.31\textwidth}
        \centering
        \includegraphics[draft=false,width=\linewidth]{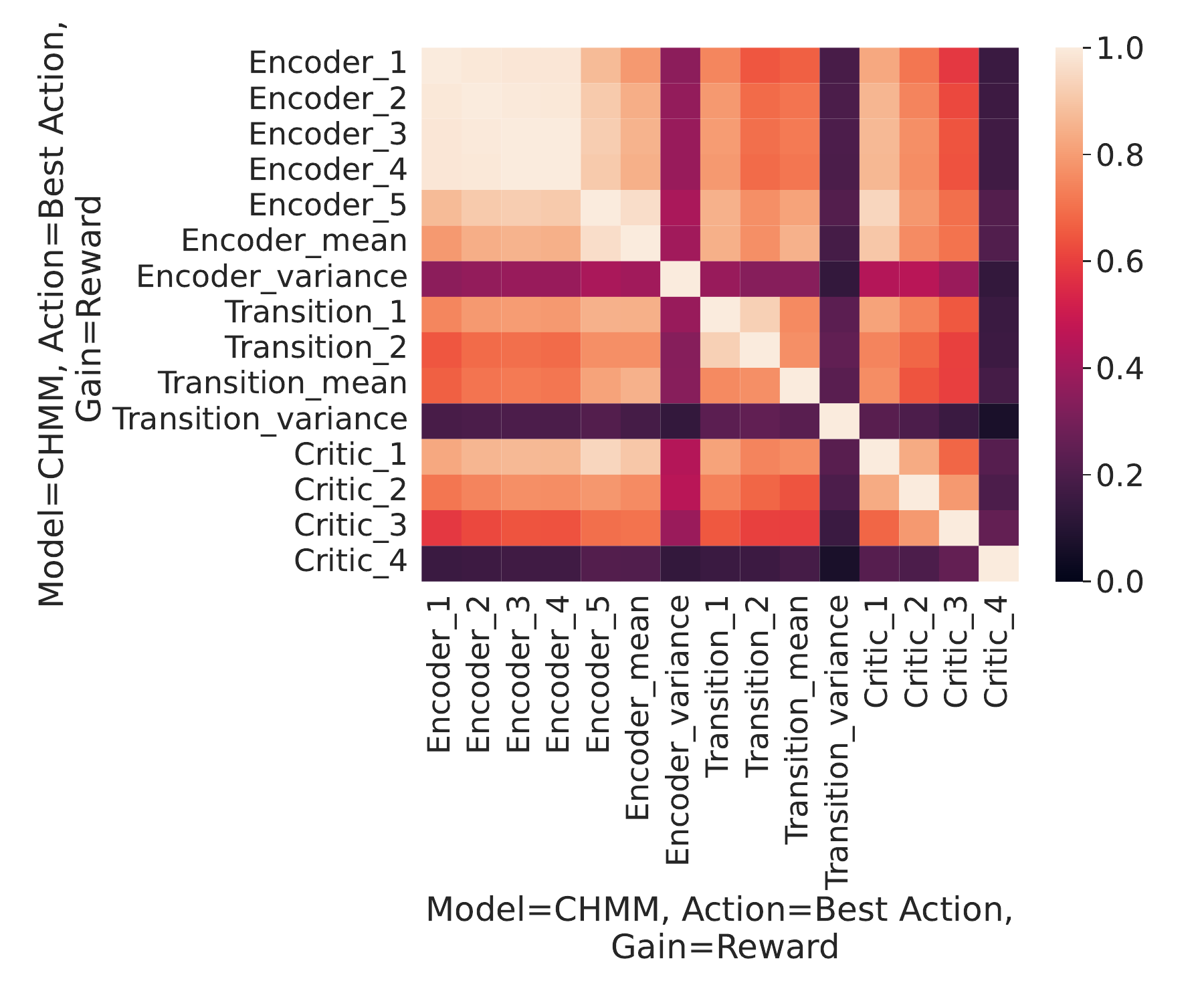}
        \caption{}\label{sfig:cka-chmm-chmm-ba}
    \end{subfigure}%
    \begin{subfigure}{.31\textwidth}
        \centering
        \includegraphics[draft=false,width=\linewidth]{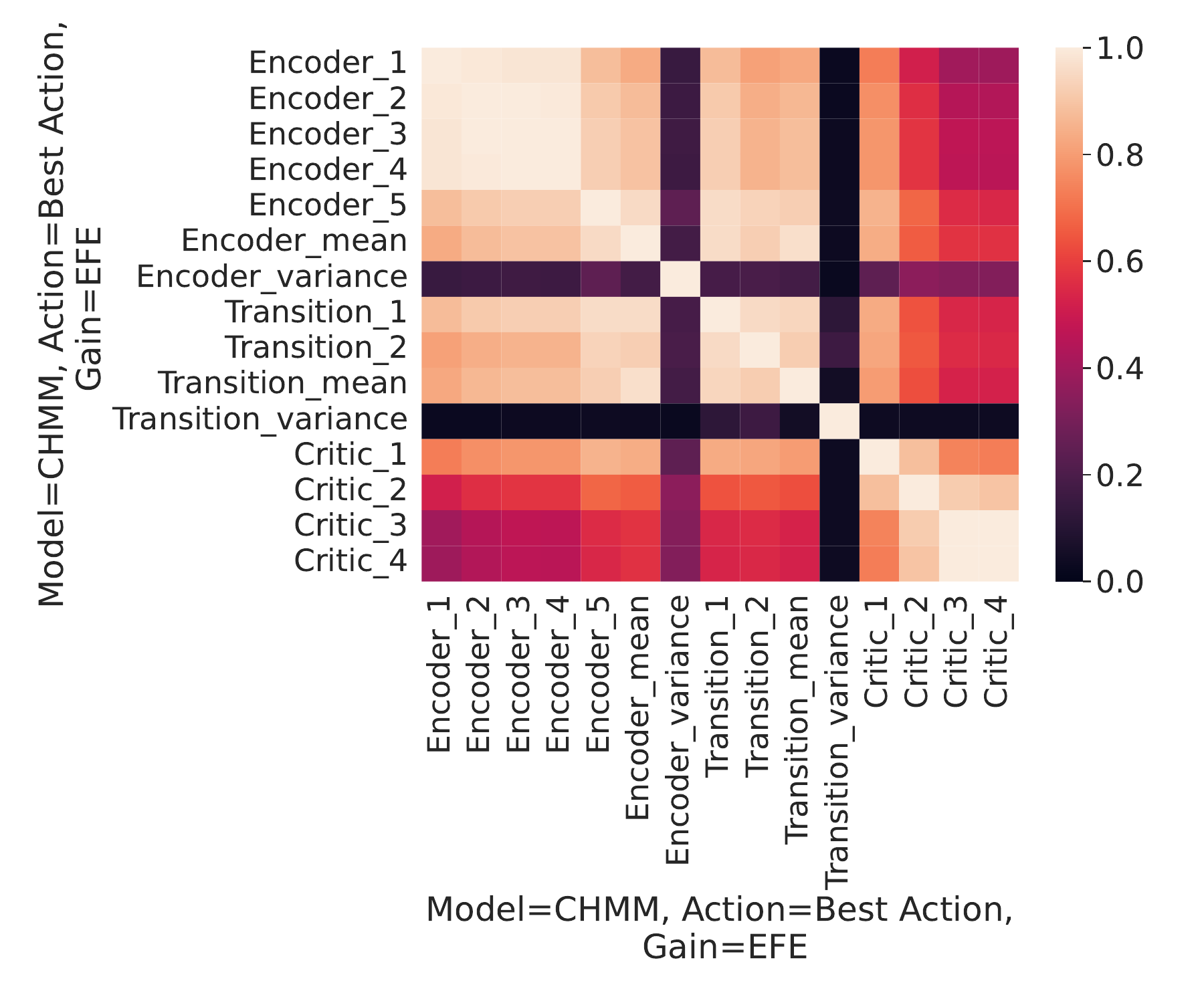}
        \caption{}\label{sfig:cka-chmm2-chmm2-ba}
    \end{subfigure}%
    \begin{subfigure}{.31\textwidth}
        \centering
        \includegraphics[draft=false,width=\linewidth]{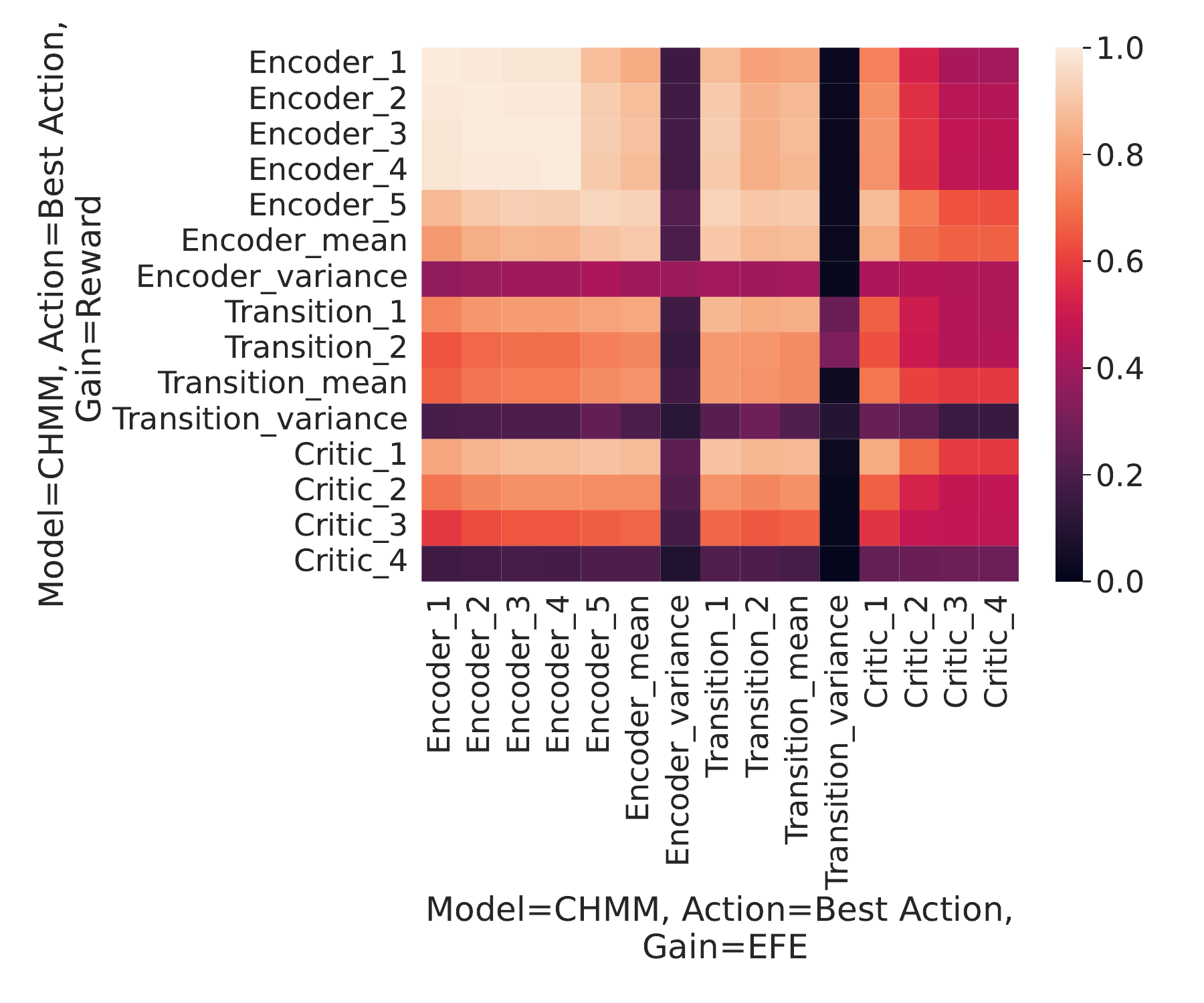}
        \caption{}\label{sfig:cka-chmm-chmm2-ba}
    \end{subfigure}

    \caption{{\small (a) shows the similarity between the representations learned by different layers of the encoder, transition and critic networks of a CHMM whose critic maximises the reward (with best action selection).
        (b) shows the similarity between the representations learned by different layers of the encoder, transition and critic networks of a CHMM whose critic minimises the EFE (with best action selection)
        (c) shows the similarity between the representations learned by two CHMMs, one whose critic optimises EFE and the other that optimises reward (both with best action selection).
    }}
    \label{fig:cka-chmm-ba}
\end{figure}

\begin{figure}[H]
    \centering
    \begin{subfigure}{.305\textwidth}
        \centering
        \includegraphics[draft=false,width=\linewidth]{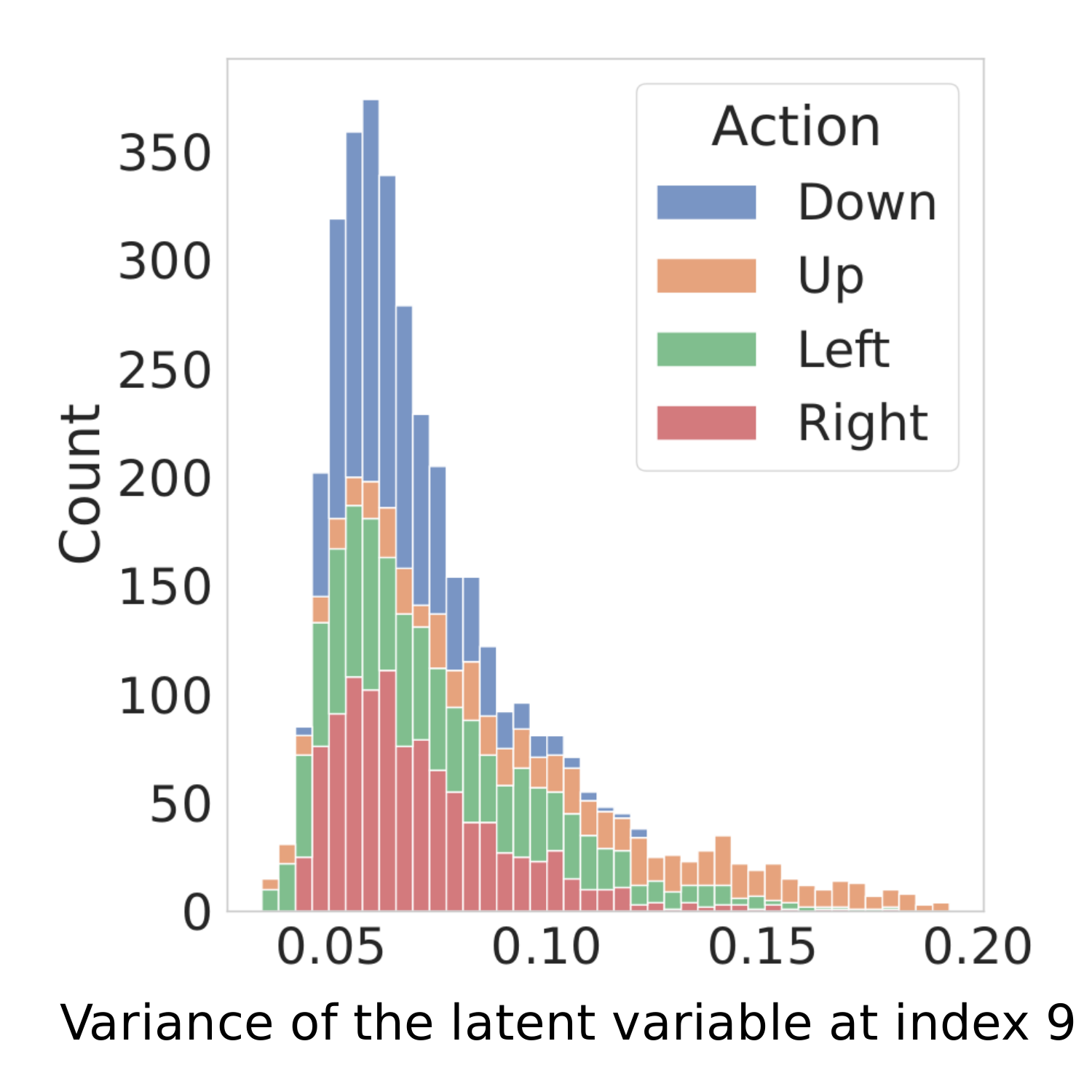}
        \caption{}\label{sfig:cka-trans-chmm-ba}
    \end{subfigure}%
    \begin{subfigure}{.3\textwidth}
        \centering
        \includegraphics[draft=false,width=\linewidth]{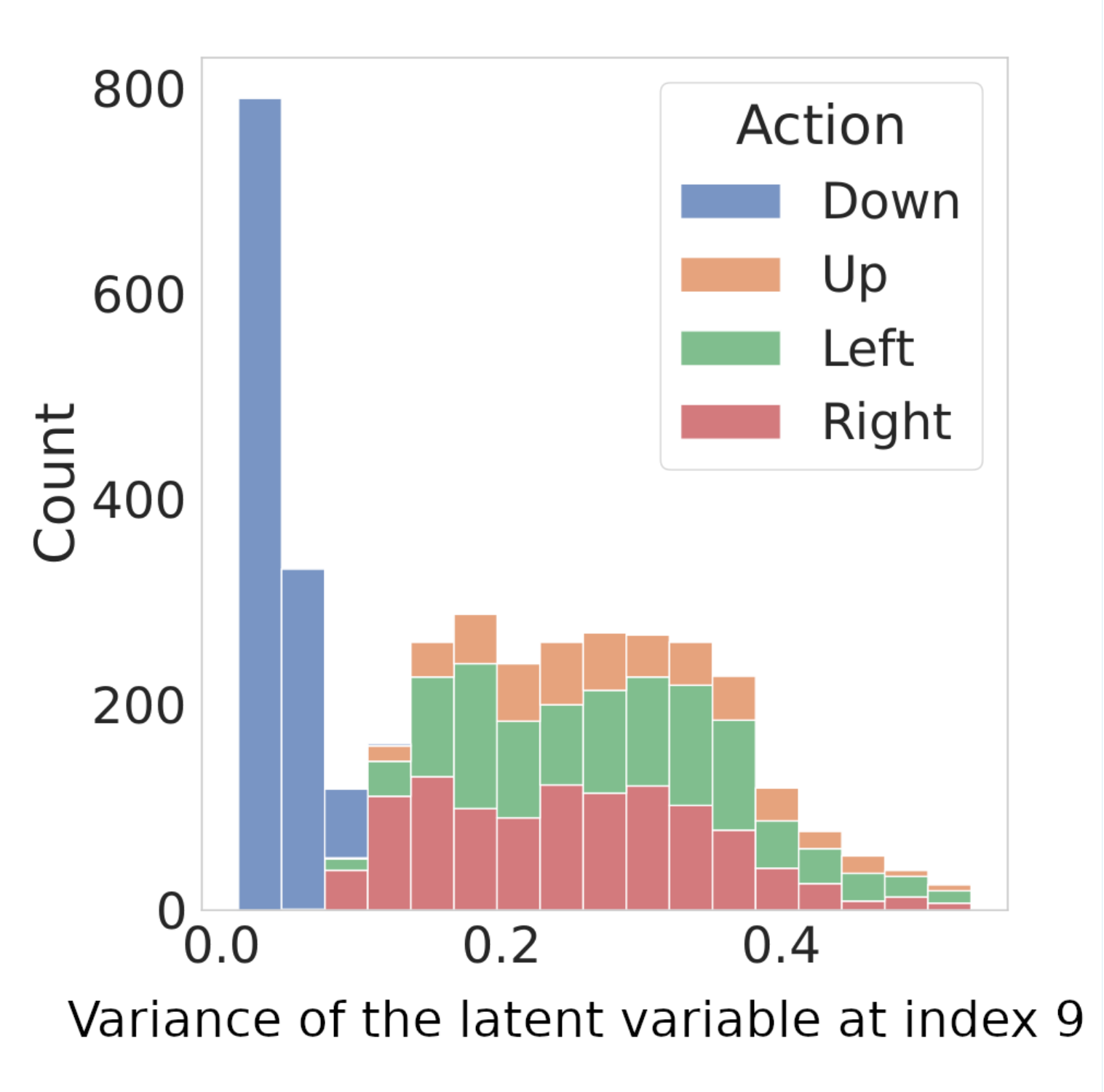}
        \caption{}\label{sfig:cka-trans-chmm-ba}
    \end{subfigure}

    \caption{{\small (a) shows one latent dimension of the variance layer of the transition network for the CHMM maximising the reward. (b) shows one latent dimension of the variance layer of the transition network for the CHMM minimising the EFE. Both figures are typical of the distributions of variance activations in the two models. Note, only the action down has low variance for the CHMM minimising the EFE. This suggests that the CHMM minimising the EFE always picks the action down, and does not gather enough data for the other actions.
    }}
    \label{fig:cka-chmm-trans-ba}
\end{figure}

\paragraph{CHMM with best action selection}
As illustrated in Figure~\ref{fig:CHMM_rewards_b}, only the CHMM whose critic maximises the reward was able to solve the task. However, one can see in Figure~\ref{fig:cka-chmm-ba} that the CHMM whose critic minimises the EFE learns representations similar to those of the CHMM maximising the reward in most layers, with the exception of the variance layer of the encoder and transition network (Encoder\_variance and Transition\_variance). To better understand the differences between the representations learned by the variance layer of both models, we fed 5K state-action pairs through the transition network, and displayed the distribution of the variances outputed by the transition network. This analysis reveals that the variance (of the variance layer) of the transition network is very small and does not change much when maximising the reward but is larger and varies more when minimising the EFE as illustrated in Figure~\ref{fig:cka-chmm-trans-ba}. This reflects a higher uncertainty of the transitions for the CHMM minimising EFE. More specifically, the CHMM minimising EFE seems to be confident for the action down, but is very uncertain for all the other actions, which suggests that the CHMM minimising EFE always picks the action down and does not gather enough data for the other actions.
\vspace{-0.5cm}
\vspace{-1cm}

\begin{figure}[ht!]
    \centering
    \begin{subfigure}{.33\textwidth}
        \centering
        \includegraphics[draft=false,width=\linewidth]{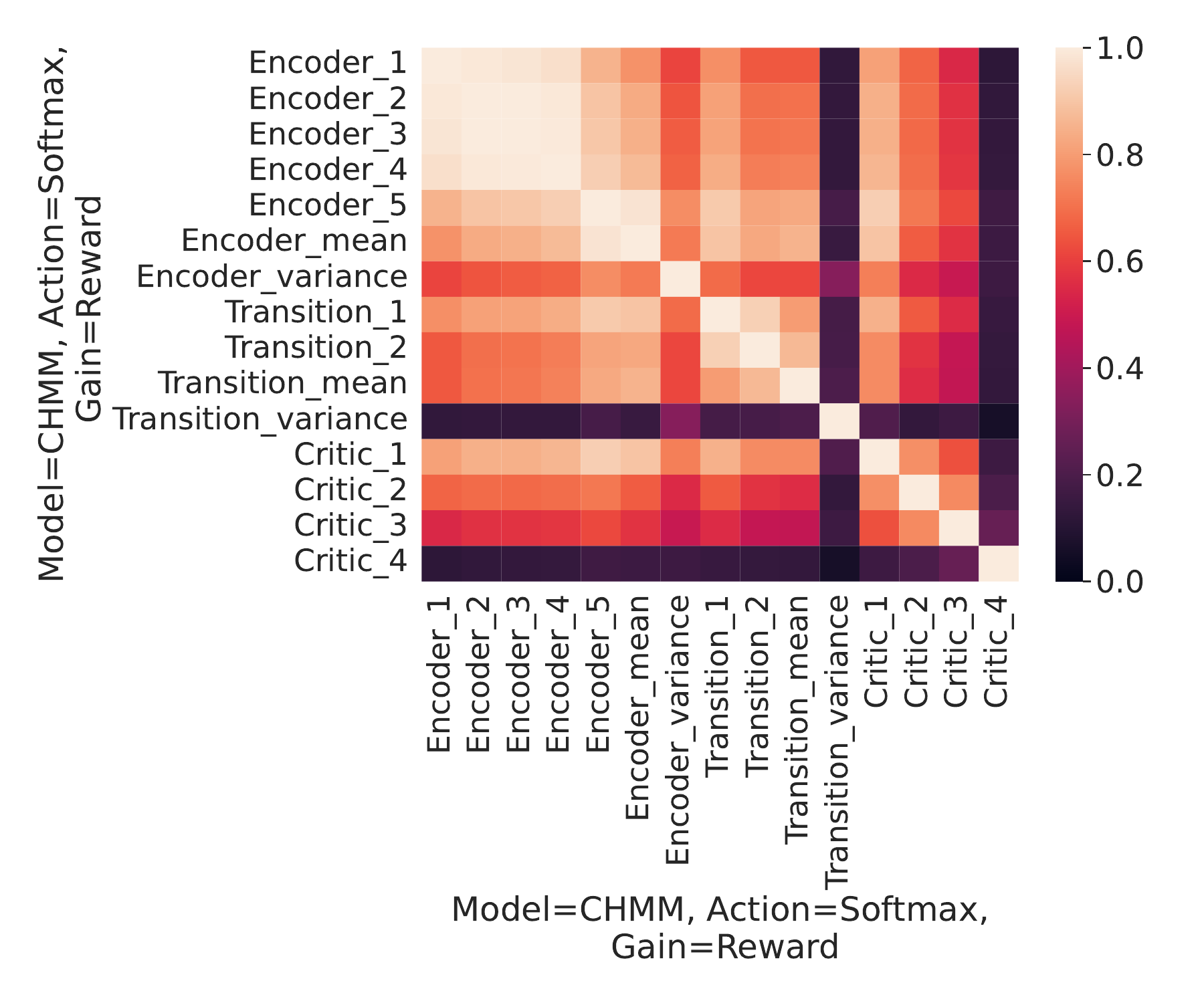}
        \caption{}\label{sfig:cka-chmm-chmm-sm}
    \end{subfigure}%
    \begin{subfigure}{.33\textwidth}
        \centering
        \includegraphics[draft=false,width=\linewidth]{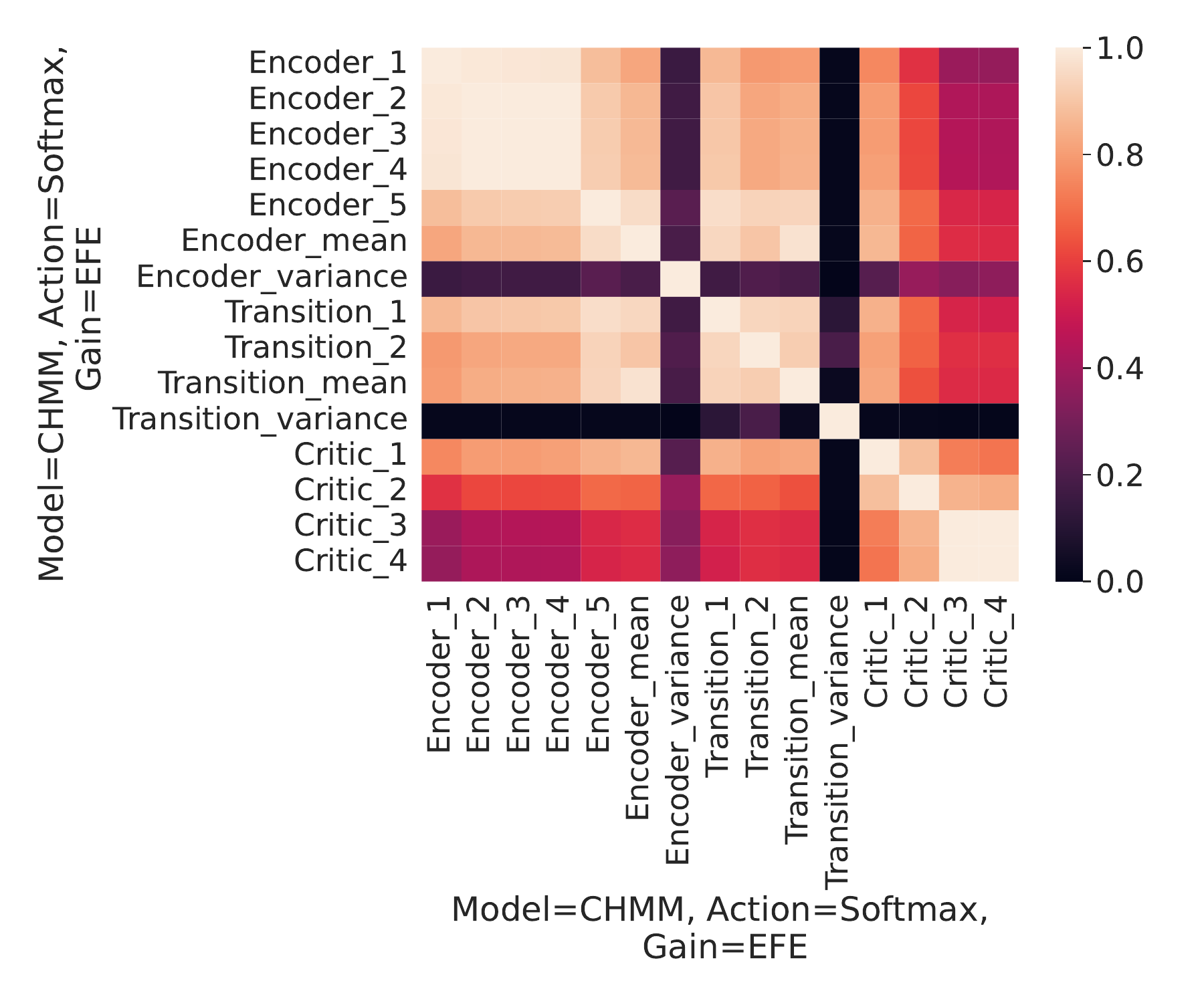}
        \caption{}\label{sfig:cka-chmm2-chmm2-sm}
    \end{subfigure}%
    \begin{subfigure}{.33\textwidth}
        \centering
        \includegraphics[draft=false,width=\linewidth]{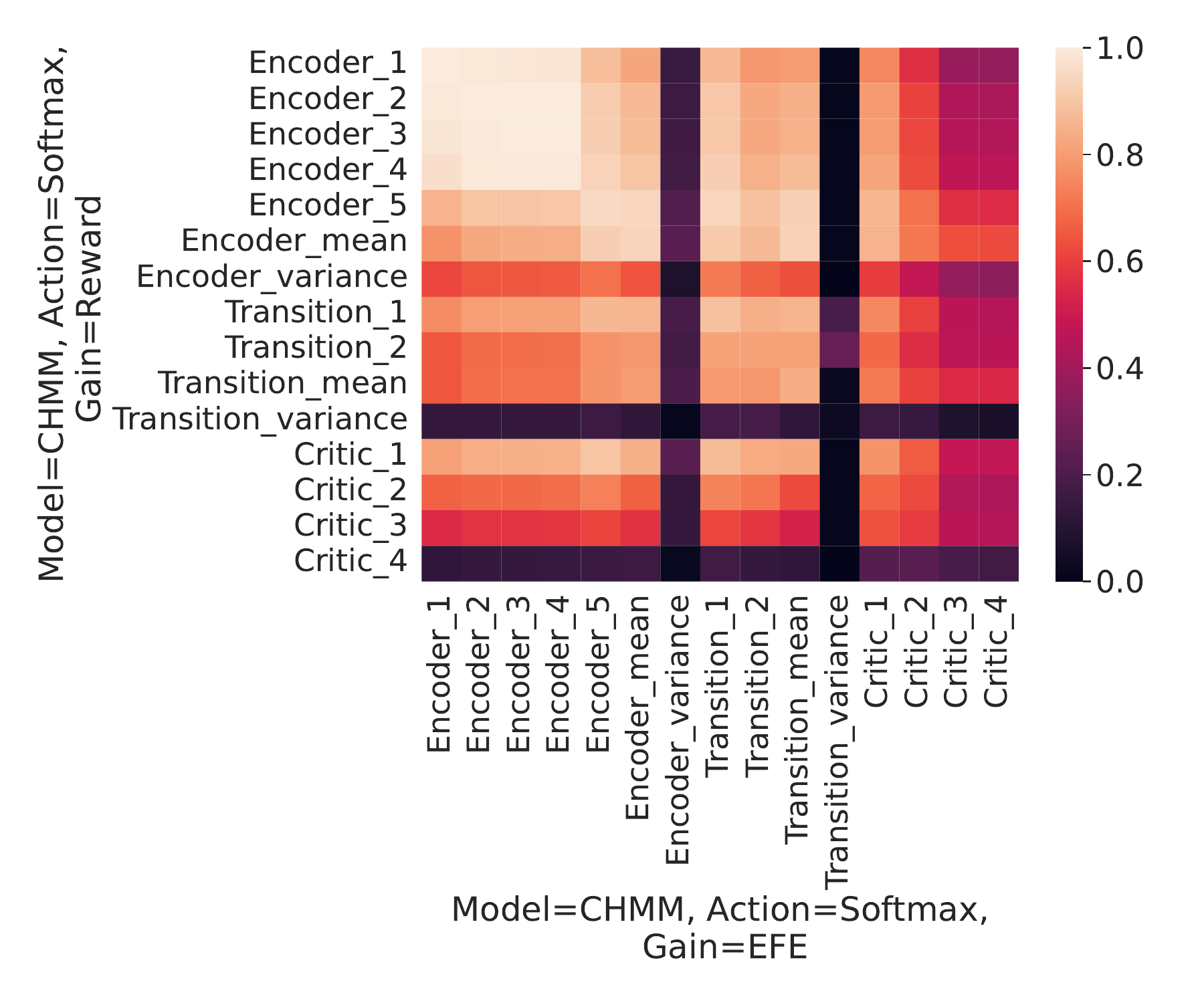}
        \caption{}\label{sfig:cka-chmm-chmm2-sm}
    \end{subfigure}

    \caption{(a) shows the similarity between the representations learned by different layers of the encoder, transition and critic networks of a CHMM whose critic maximises the reward (with softmax action selection). (b) shows the similarity between the representations learned by different layers of the encoder, transition and critic networks of a CHMM whose critic minimises the EFE (with softmax action selection). (c) shows the similarity between the representations learned by two CHMMs, one whose critic optimises EFE and the other that optimises reward (both with softmax action selection).}
    \label{fig:cka-chmm-sm}
\end{figure}

\begin{figure}[ht!]
    \centering
    \vspace{-0.5cm}
    \begin{subfigure}{.42\textwidth}
        \centering
        \includegraphics[draft=false,width=\linewidth]{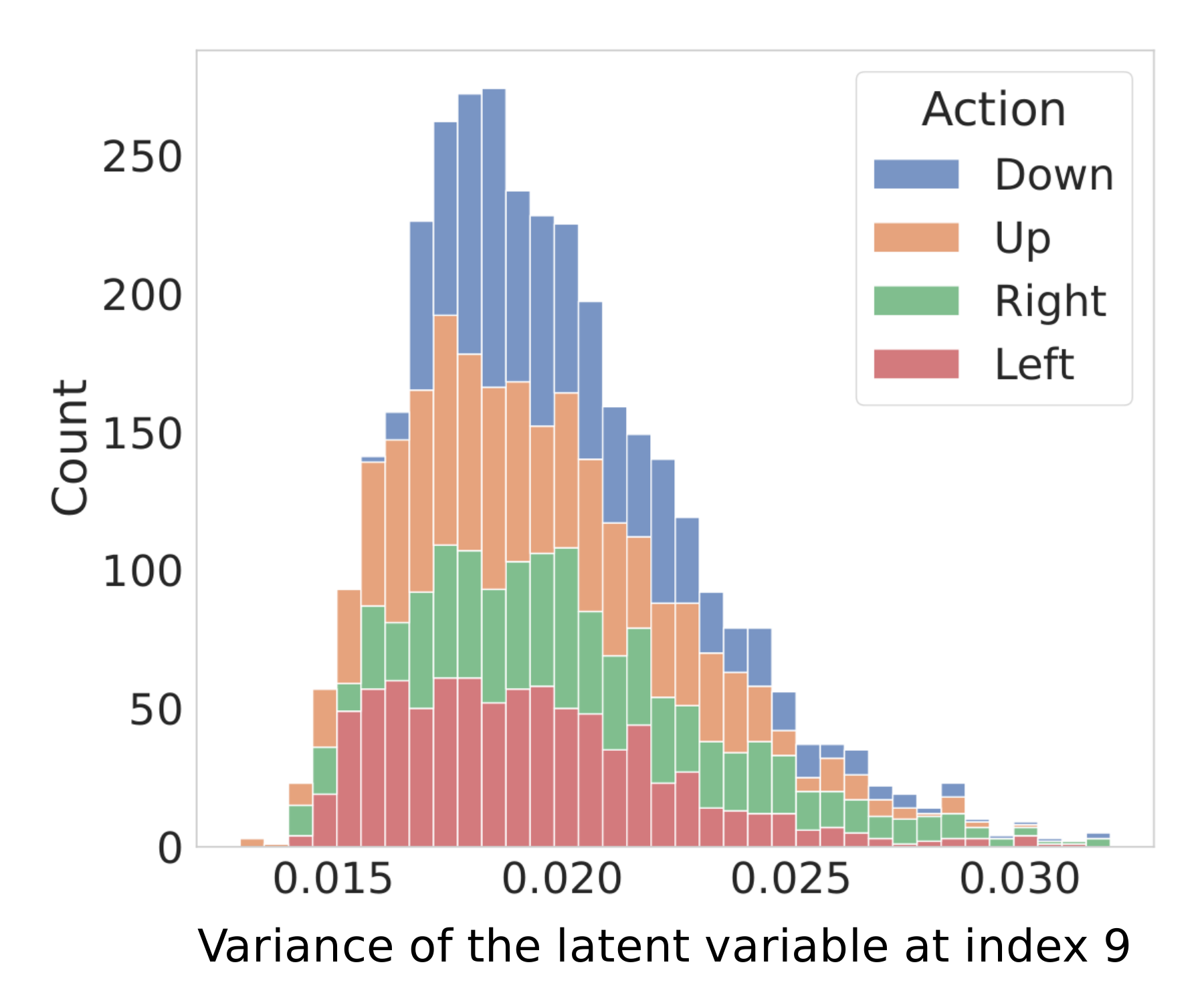}
        \caption{}\label{sfig:cka-trans-chmm-sa}
    \end{subfigure}%
    \begin{subfigure}{.35\textwidth}
        \centering
        \includegraphics[draft=false,width=\linewidth]{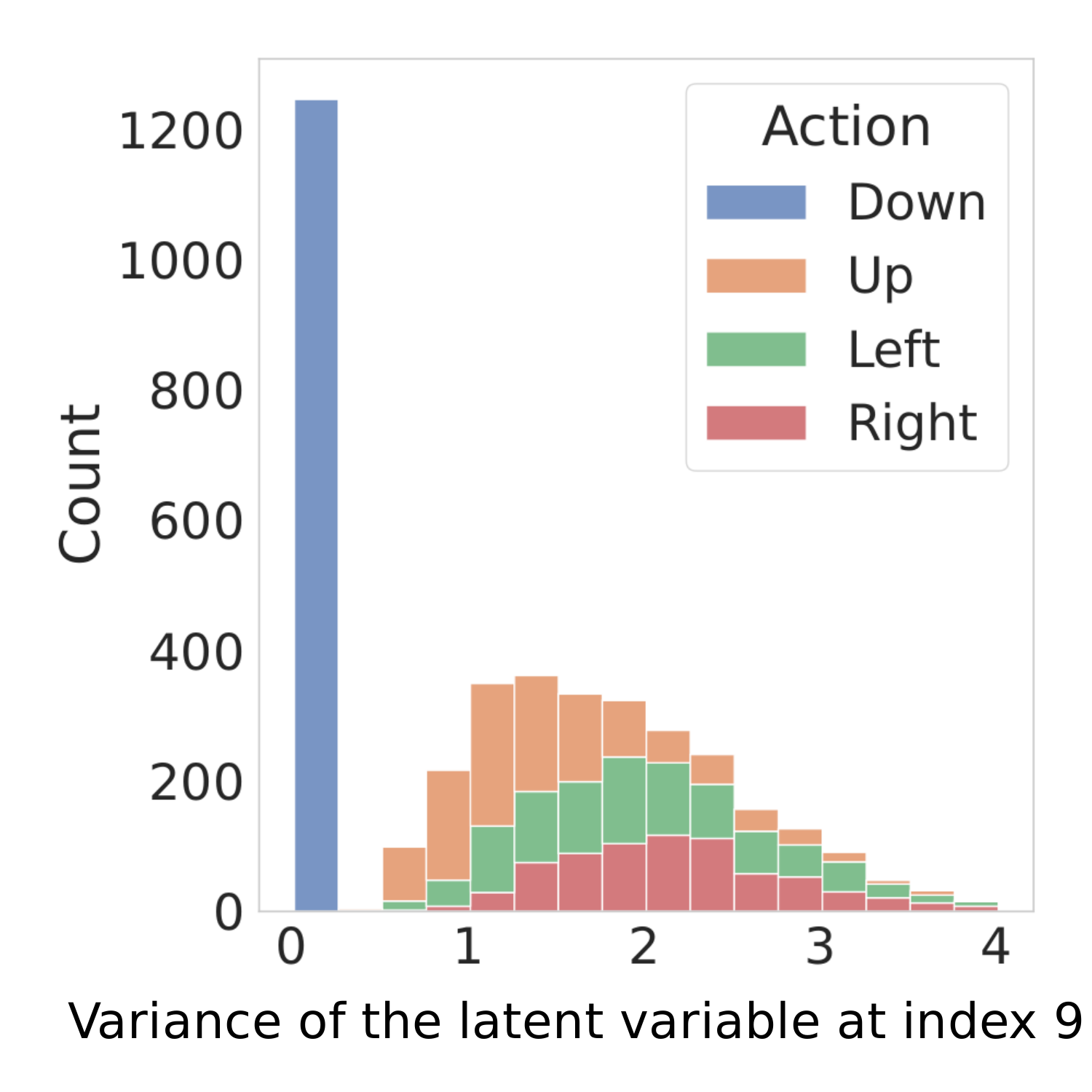}
 in which all our simulations will be         \caption{}\label{sfig:cka-trans-chmm2-sa}
    \end{subfigure}

    \caption{{\small (a) shows one latent dimension of the variance layer of the transition network for the CHMM maximising the reward.
             (b) shows one latent dimension of the variance layer of the transition network for the CHMM minimising the EFE.
             Both figures are representative of the distributions of variance activations in the two models.
             Note, only the action down has low variance for the CHMM minimising the EFE.
             This suggests that the CHMM minimising the EFE always pick the action down, and does not gather enough data for the other actions.
    }}
    \label{fig:cka-chmm-trans-sa}
\end{figure}

\paragraph{CHMM with softmax action selection}
As illustrated in Figure~\ref{fig:CHMM_rewards_s}, none of the CHMMs with softmax action selection were able to solve the task. Once again, the variance (of the variance layer) of the transition network is very different in the CHMM whose critic minimises the EFE compared to the CHMM whose critic maximises the reward (see Figure~\ref{fig:cka-chmm-sm}c at the intersection of the two Transition\_variances). We further observe the same trend regarding the uncertainty of the output of the transition network when optimising the EFE, as shown in Figure~\ref{fig:cka-chmm-trans-sa}. While this may explain why the model optimising the EFE does not solve the task, this does not indicate why the model maximising the reward cannot solve the task, and we can hypothesise that those results may be attributed to the softmax action selection. More precisely, if the values predicted by the critic network are very close to each other, then an agent using softmax sampling may perform random actions.

\subsubsection{Degenerate behaviour with the expected free energy?} \label{ssec:degenerate_behaviour_and_EFE}

\begin{figure}[H]
    \centering
    \begin{subfigure}{.4\textwidth}
        \centering
        \includegraphics[draft=false,width=\linewidth]{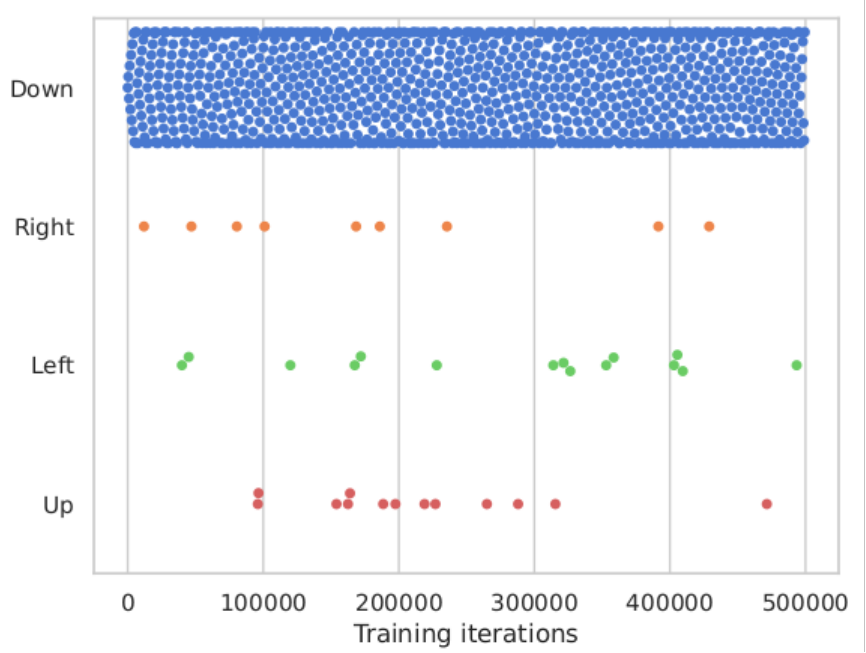}
        \caption{}\label{fig:action_picked_efe}
    \end{subfigure}%
    \begin{subfigure}{.4\textwidth}
        \centering
        \includegraphics[draft=false,width=\linewidth]{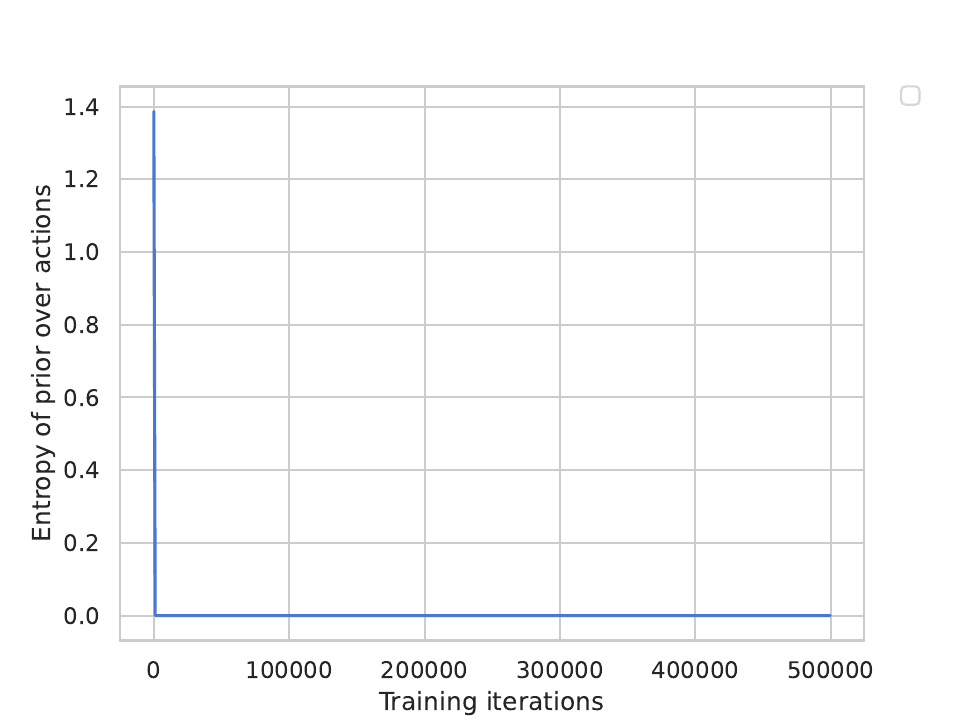}
        \caption{}\label{fig:entropy_action_efe}
    \end{subfigure}
    \begin{subfigure}{.4\textwidth}
        \centering
        \includegraphics[draft=false,width=\linewidth]{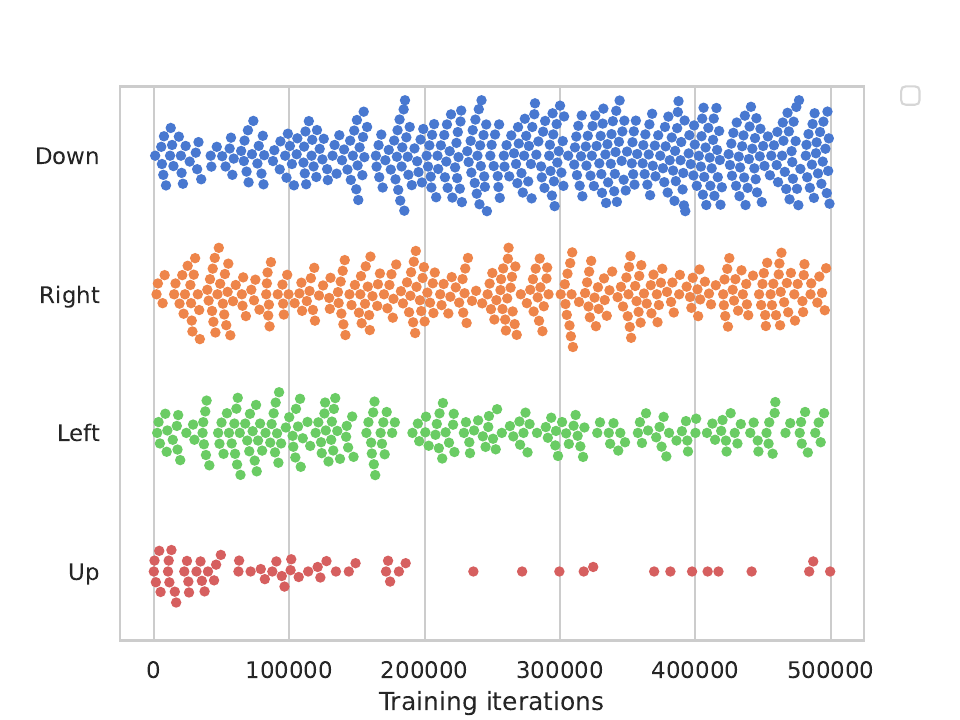}
        \caption{}\label{fig:action_picked_reward}
    \end{subfigure}%
    \begin{subfigure}{.4\textwidth}
        \centering
        \includegraphics[draft=false,width=\linewidth]{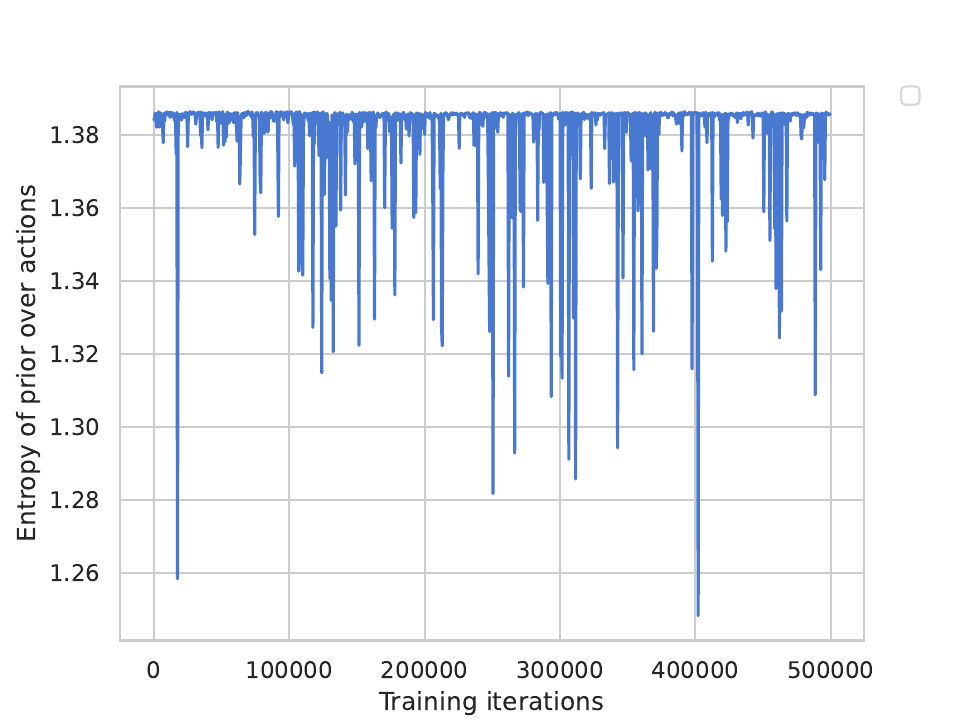}
        \caption{}\label{fig:entropy_action_reward}
    \end{subfigure}

    \caption{{\small (a) shows the action taken for each planning iteration when the CHMM is minimising expected free energy.
             (b) shows the entropy of the prior over actions when the CHMM is minimising the EFE.
             (c) shows the action taken for each planning iteration when the CHMM is maximising reward.
             (d) shows the entropy of the prior over actions when the CHMM is maximising reward.
    }}
    \label{fig:actions_from_critic}
\end{figure}

Up to now, we saw that the CHMM minimising expected free energy (EFE) was not able to solve the task. Also, we discovered that the transition network is uncertain for the actions: up, left, and right, which suggests that the CHMM minimising EFE always takes action down. Figure \ref{fig:actions_from_critic} corroborates this story. Indeed, Figure \ref{fig:action_picked_efe} shows that the agent minimising EFE almost exclusively picked action down, and Figure \ref{fig:entropy_action_efe} shows that the entropy of the prior over actions very quickly converges to zero.

In contrast, the CHMM maximising reward, keeps on selecting the actions right and left, which enables it to drag the shape towards the appropriate corner (see Figure \ref{fig:action_picked_reward}). Also, as shown in Figure \ref{fig:entropy_action_reward}, the entropy of the prior over actions remained a lot higher than zero. Note, the only difference between the CHMM minimising EFE and the one maximising reward is the information gain, which is defined as the KL divergence between the output of the transition and encoder networks. Since the EFE is minimised, the output of those two networks need to be as close as possible to each other.

This suggests that the CHMM minimising EFE is picking a single action (down), and becomes an expert at predicting the future when selecting this action. This effectively makes the KL divergence between the output of the transition and encoder networks small. Additionally, when selecting the action down, the average reward is zero, because (in the dSprites dataset) there are as many shapes on the left of the image as on the right, and when crossing the bottom line, the agent receives a reward which is linearly increasing (or decreasing) as a corner is approached and is zero at the center of the image. For all the other actions, the expected reward will be negative because after 50 action-perception cycles without crossing the bottom line, the trial is interrupted and the agent receive a reward of -1. Thus, if the CHMM has to stick to a single action to keep the KL divergence  small, then the best action it can choose is down, i.e., action down has the highest expected reward.

Also, Figure \ref{fig:IG_progressive} shows the impact of adding X\% of the information gain into the objective function, i.e., the agent starts by only maximising reward (c.f. Equation \ref{eq:reward_maximisation}), and after 200K training iterations minimises reward plus X\% of the information gain. One can see that adding even 1\% of the information gain already dramatically decreases the amount of reward gathered.

To conclude, the same information gain that is intended to give an EFE minimising agent its exploration behaviour, also prevents the agent from solving the dSprites environment. This is because the agent is reduced to picking a single action, leading to a suboptimal policy.

{
\definecolor{LightBlueCHMM}{RGB}{51, 187, 238}
\definecolor{GreenCHMM}{RGB}{0, 153, 136}
\definecolor{GrayCHMM}{RGB}{187,187,187}
\definecolor{OrangeCHMM}{RGB}{255, 112, 67}
\definecolor{PinkCHMM}{RGB}{238, 51, 119}
\definecolor{DarkBlueCHMM}{RGB}{0, 119, 187}
\begin{figure}[H]
	\begin{center}
	\begin{tikzpicture}[square/.style={regular polygon,regular polygon sides=4}, scale=1]
	    \node at (-2, 0) {\includegraphics[draft=false,width=0.5\linewidth]{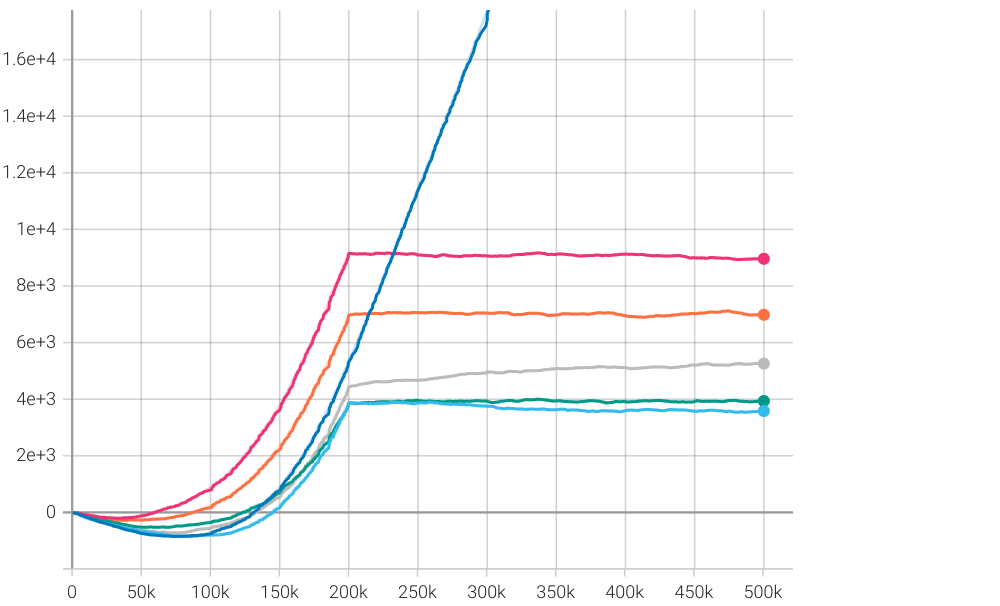}};

	    \node[black] at (4.675, -1) {CHMM[0\%]:};
		\draw[fill=DarkBlueCHMM] (6, -0.8) rectangle (7, -1.2);

	    \node[black] at (4.675, -0.5) {CHMM[1\%]:};
		\draw[fill=GrayCHMM] (6, -0.7) rectangle (7, -0.3);

	    \node[black] at (4.675, 0) {CHMM[5\%]:};
		\draw[fill=OrangeCHMM] (6, -0.2) rectangle (7, 0.2);

	    \node[black] at (4.585, 0.5) {CHMM[15\%]:};
		\draw[fill=GreenCHMM] (6, 0.3) rectangle (7, 0.7);

	    \node[black] at (4.585, 1) {CHMM[25\%]:};
		\draw[fill=PinkCHMM] (6, 0.8) rectangle (7, 1.2);

	    \node[black] at (4.585, 1.5) {CHMM[50\%]:};
		\draw[fill=LightBlueCHMM] (6, 1.3) rectangle (7, 1.7);

	    \node[black] at (-2.8, -3.5) {Training iterations};
	    \node[black, rotate=90] at (-7.3, 0) {Total Reward};
	\end{tikzpicture}
	\end{center}
   \caption{This figure illustrates the total reward aggregated by CHMM agents during the 500K iterations of training. All the agents start by only maximising reward, and after 200K training iterations, X\% of the information gain is added to the objective function. Note, even adding 1\% of the information gain is enough to drastically reduce the total reward aggregated by the agent. The differences in trajectories before 200K are arbitrary, arising from differences in random initializations.}
   \label{fig:IG_progressive}
\end{figure}
}

\subsection{DAI agent} \label{ssec:dai_results}

\begin{figure}[H]
	\begin{center}
	{
\definecolor{Red}{RGB}{188, 9, 9}
\definecolor{Green}{RGB}{60, 128, 56}
\begin{figure}[H]
	\begin{center}
    \resizebox{0.8\textwidth}{!}{%
    \begin{tikzpicture}[square/.style={regular polygon,regular polygon sides=4}, scale=0.75]
	    \node at (-0.5, 0) {\includegraphics[scale=0.22]{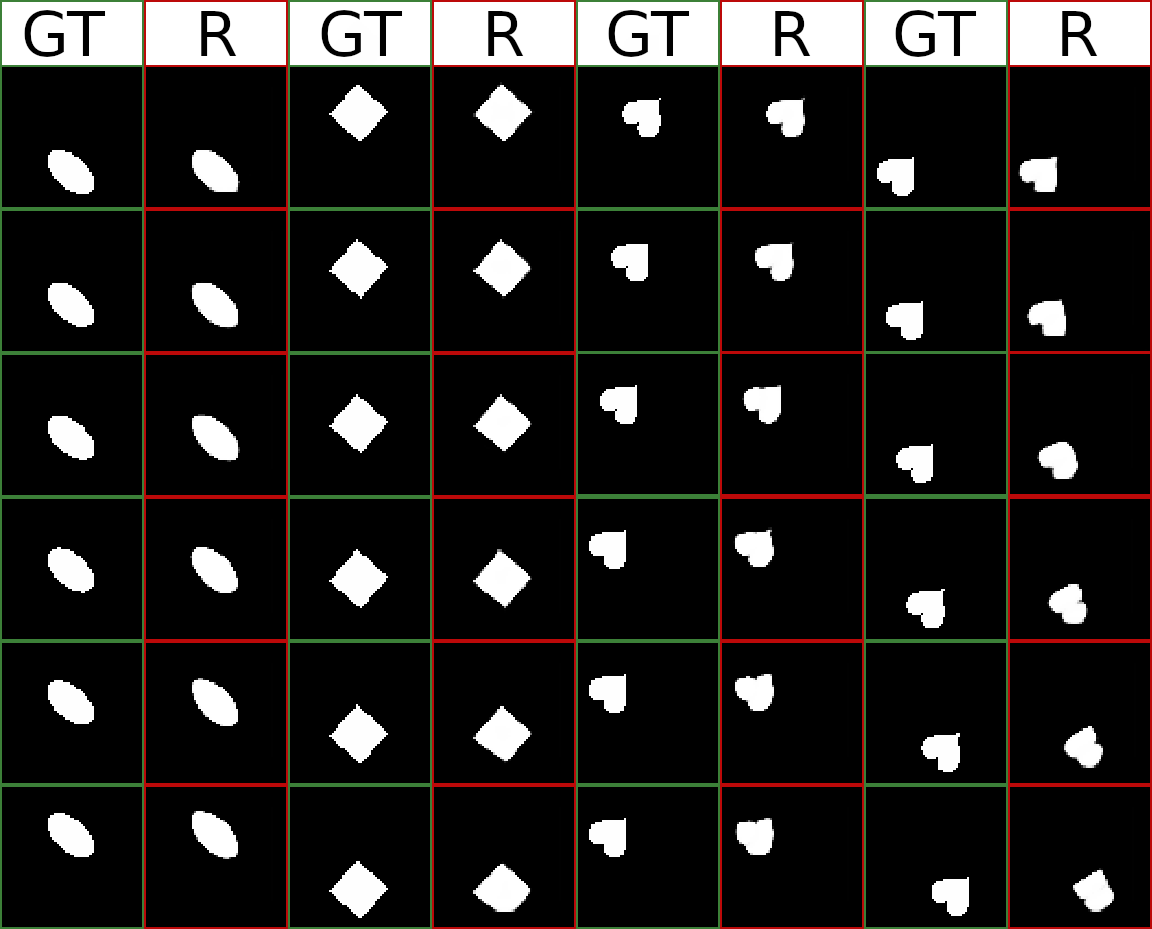}};
	    \node[gray!60!black] at (8, 0) {Ground Truth (GT):};
	    \node[gray!60!black] at (8.1, 0.5) {Reconstruction (R):};
		\draw[fill=Green] (10.5, 0.2) rectangle (11.5, -0.2);
		\draw[fill=Red] (10.5, 0.3) rectangle (11.5, 0.7);
	\end{tikzpicture}
	} %
	\end{center}
\end{figure}
}
	\end{center}
   \caption{This figure illustrates the sequences of reconstructed images generated by the DAI after 500K training iterations. The columns alternate between the ground truth images and the reconstructed images. Time passes vertically (from top to bottom), and within each column, the same action is executed repeatedly.}
   \label{fig:DAI_reconstruction}
\end{figure}

\begin{figure}[H]
	\begin{center}
    \resizebox{0.5\textwidth}{!}{%
	\begin{tikzpicture}[square/.style={regular polygon,regular polygon sides=4}, scale=0.75]
	    \node at (0, 0) {\includegraphics[scale=0.4]{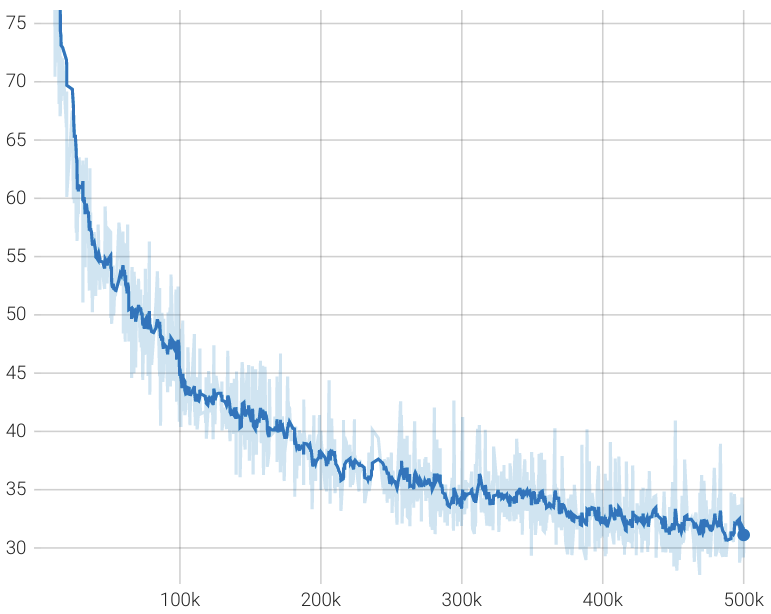}};
	    \node[gray!60!black, rotate=90] at (-8, 0) {VFE};
	    \node[gray!60!black] at (0, -6.7) {Training Iterations};
	\end{tikzpicture}
	} %
	\end{center}
   \caption{{\small This figure illustrates the variational free energy of the DAI agent during the 500K iterations of training. This agent was maximising reward while sampling actions from a softmax function of the policy network output. The agent was able to minimise its variational free energy.}}
   \label{fig:DAI_vfe}
\end{figure}

In this section, we report the results obtained by the DAI agent, when using different action selection strategies and different definitions of the expected free energy. First, most of the fifteen DAI agents crashed because of numerical instability, i.e., the VFE suddenly became ``Not a Number". The only DAI agent that survived (i.e., did not crash) was maximising rewards while performing softmax sampling for action selection. Figure \ref{fig:DAI_vfe} shows that the DAI agent successfully minimises its variational free energy, but as shown in Figure \ref{fig:DAI_rewards}, the DAI agent does not solve the task and performs as well as a random agent. Finally, Figure \ref{fig:DAI_reconstruction} shows sequences of images produced by the DAI agent after 500K training iterations. Note, while the agent does not solve that task, it understands the dynamics of the environment pretty well. However, the agent struggles with images representing hearts.

By comparing Figures \ref{fig:HMM_reconstruction}, \ref{fig:CHMM_reconstruction} and \ref{fig:DAI_reconstruction}, we see that the DAI agent with softmax sampling has a better reconstruction than the CHMM agent with the $\mathring{\epsilon}$-greedy algorithm (which is presented in Figure \ref{fig:CHMM_reconstruction}). In contrast, the DAI agent does not reconstruct the sequences of images as well as the HMM agent performing random actions (which is presented in Figure \ref{fig:HMM_reconstruction}).

\begin{figure}[ht!]
    \centering
    \begin{subfigure}{.3\textwidth}
        \centering
        \includegraphics[draft=false,width=\linewidth]{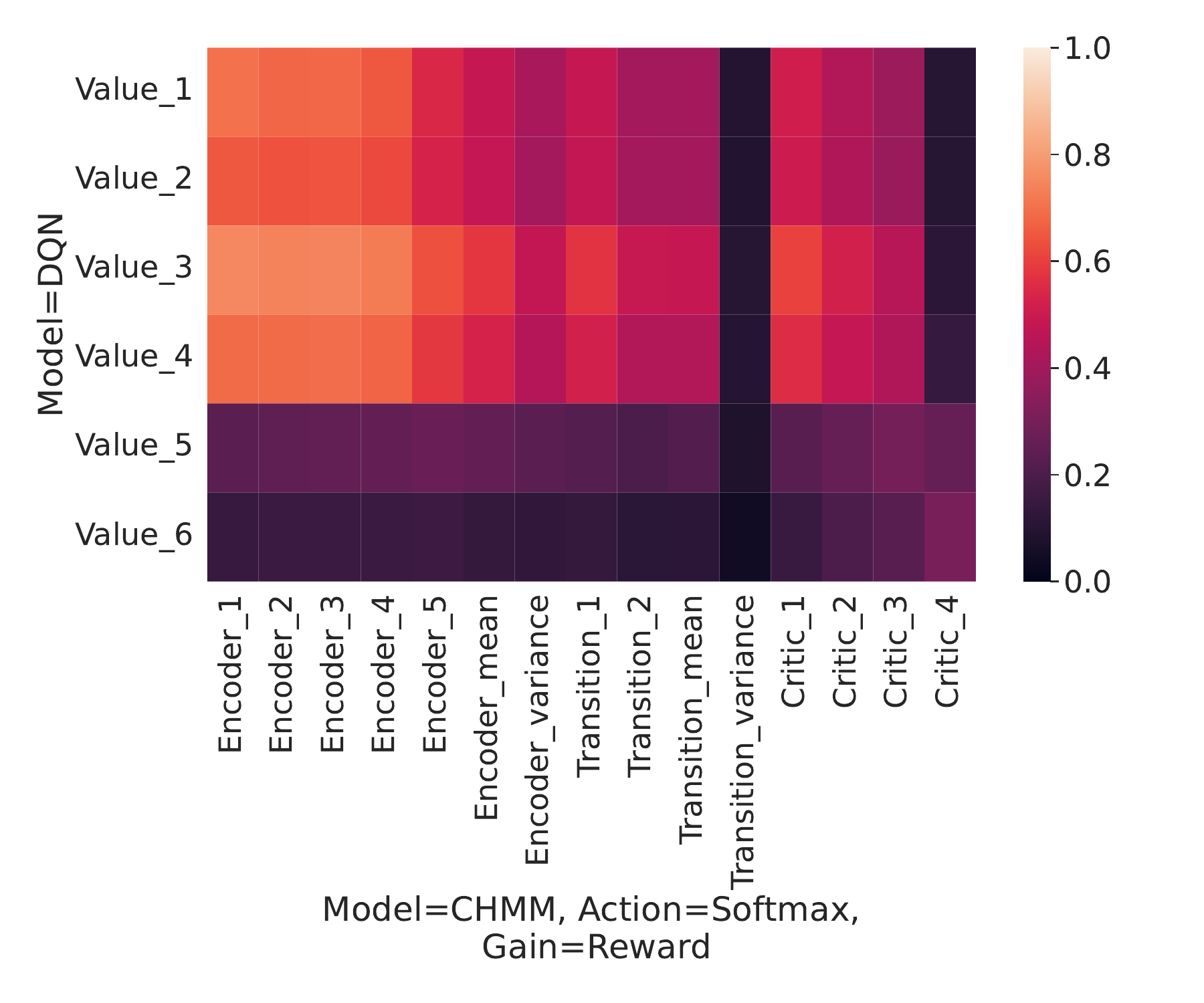}
        \caption{}\label{sfig:cka-dqn-chmm}
    \end{subfigure}%
    \begin{subfigure}{.3\textwidth}
        \centering
        \includegraphics[draft=false,width=\linewidth]{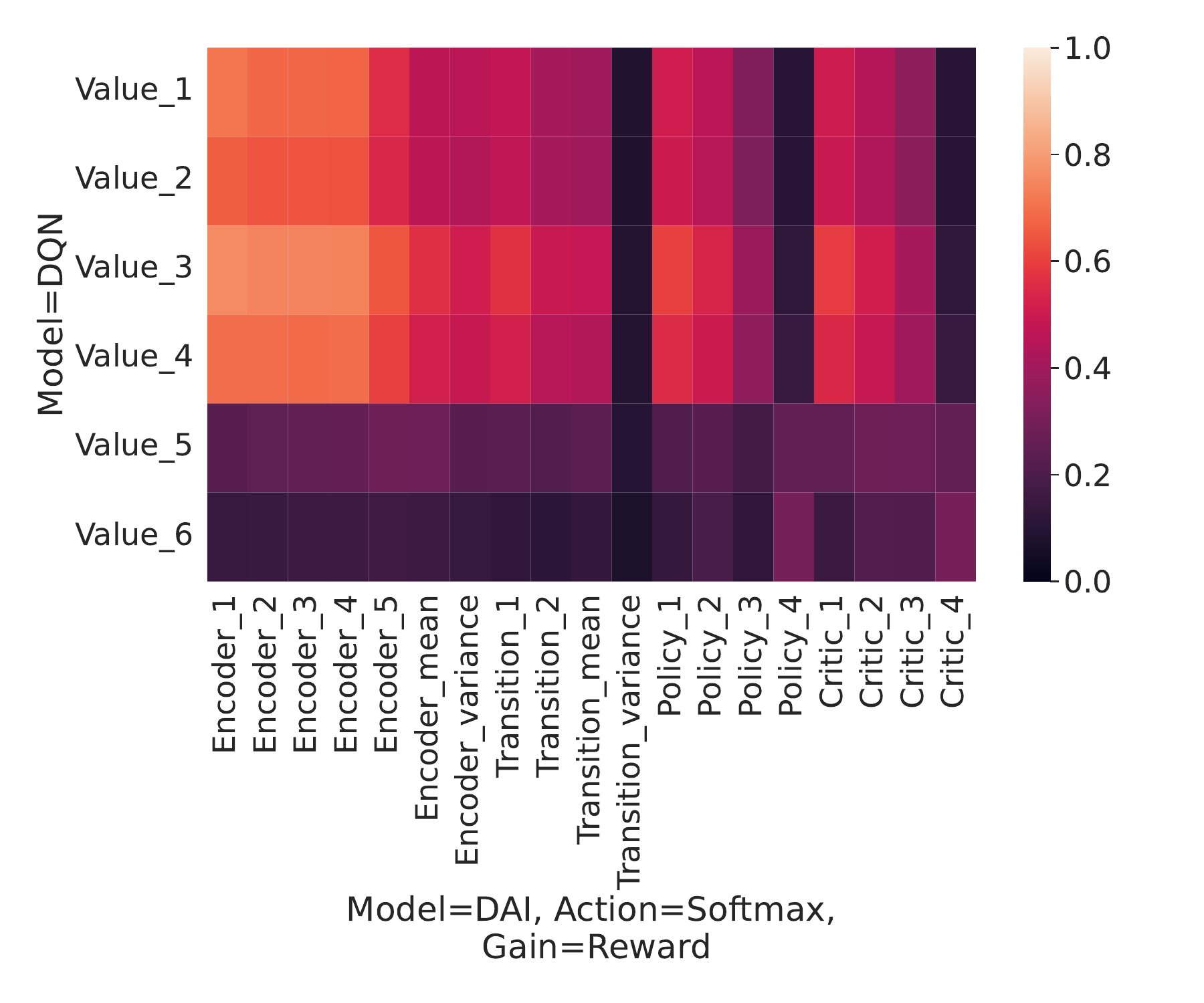}
        \caption{}\label{sfig:cka-dqn-dai}
    \end{subfigure}%
    \begin{subfigure}{.3\textwidth}
        \centering
        \includegraphics[draft=false,width=\linewidth]{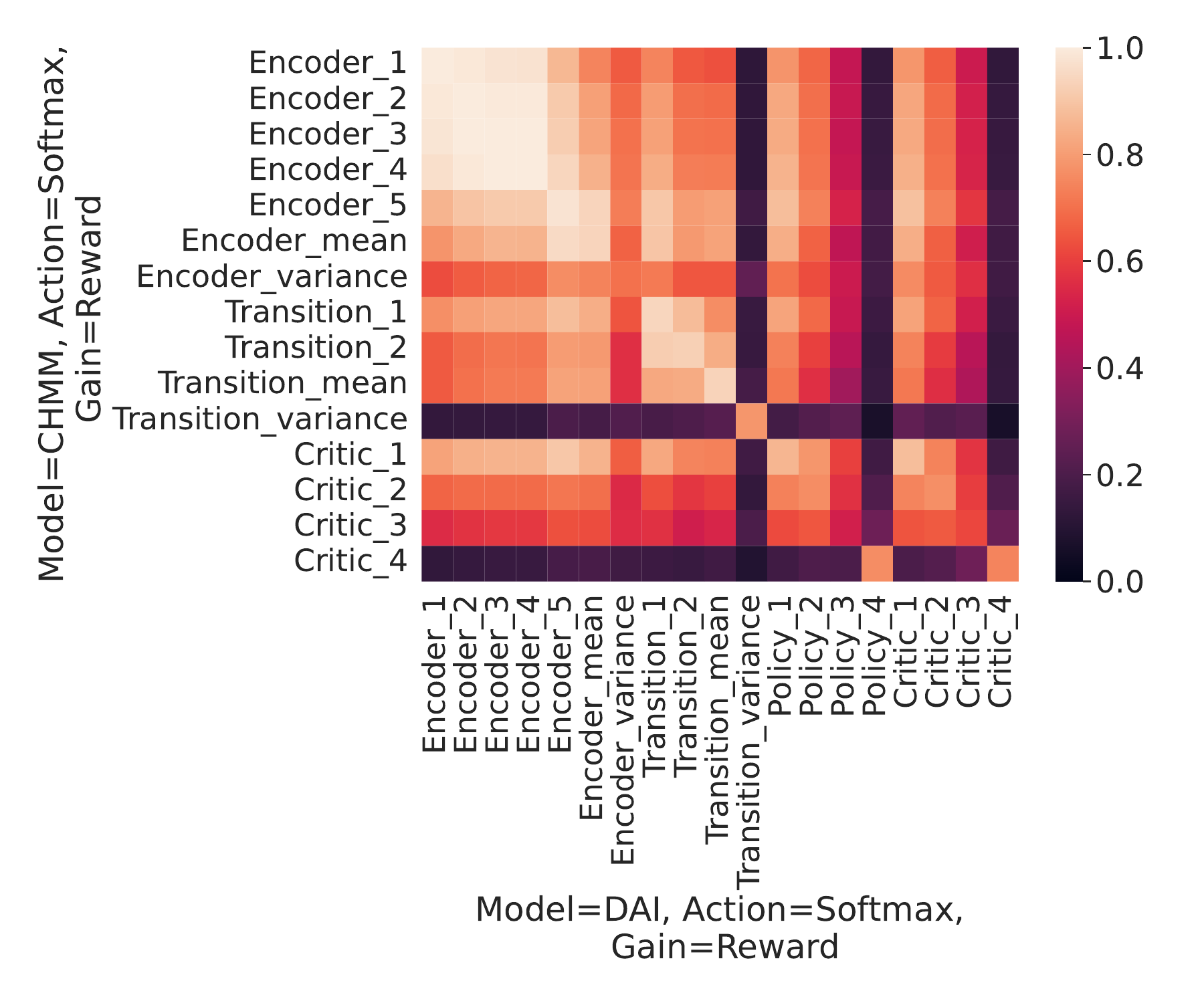}
        \caption{}\label{sfig:cka-chmm-dai}
    \end{subfigure}%

    \caption{(a) shows the similarity between the representations learned by a DQN and a CHMM maximising the reward and using softmax action selection. (b) shows the similarity between the representations learned by a DQN and a DAI maximising the reward and using softmax action selection. (c) shows the similarity between the representations learned by a CHMM and a DAI. Both maximise the reward and use softmax action selection.
    }
    \label{fig:cka-dai}
\end{figure}

As previously mentioned, the only DAI that did not crash during training used softmax action selection and had a critic maximising reward. We can see in Figures~\ref{sfig:cka-dqn-dai} and~\ref{sfig:cka-dqn-chmm} that the representational similarity between this DAI and DQN is very close to the representational similarity between a DQN and a CHMM using the same action selection and maximising reward. This is further confirmed by a comparison between the CHMM and the DAI model in Figure~\ref{sfig:cka-chmm-dai}. Interestingly, we can see that the policy and critic network learn similar representations, indicating that the policy network is learning correctly. However, we previously inferred that the softmax action selection may be suboptimal and this seems to hold true for the DAI as well, given that it is unable to solve the task.
\vspace{-0.75cm}
{
\definecolor{RedDQN}{RGB}{194,49,30}
\definecolor{BlueCHMM}{RGB}{51,116,187}
\begin{figure}[H]
	\begin{center}
	\begin{tikzpicture}[square/.style={regular polygon,regular polygon sides=4}, scale=0.75]
	    \node at (0, 0) {\includegraphics[scale=0.4]{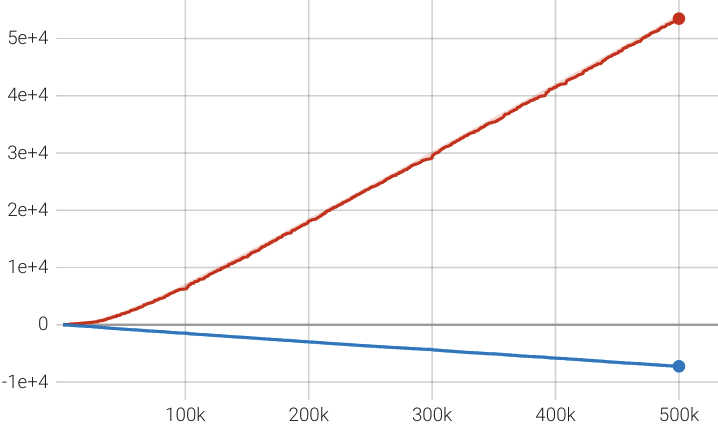}};
	    \node[gray!60!black, rotate=90] at (-7.7, 0) {Total rewards};
	    \node[gray!60!black] at (0, -4.8) {Training Iterations};
	    \node[RedDQN, anchor=west] at (7.2, 4) {DQN};
	    \node[BlueCHMM, anchor=west] at (7.2, -2.6) {$DAI[G^4]$};
	\end{tikzpicture}
	\end{center}
   \caption{{\small This figure illustrates the total amount of reward gathered by a DAI agent during the 500K iterations of training. This agent was maximising rewards while sampling actions from a softmax function of the policy network output. Put simply, the DAI agent does not solve the task and performs at the level of a random agent.}}
   \label{fig:DAI_rewards}
\end{figure}
}
\vspace{-0.75cm}

\section{Discussion of epistemic value} \label{sec:discussion}

In this section, we discuss the decomposition of the expected free energy into epistemic and extrinsic value. More precisely, intrigued by the poor results of deep active inference agents (e.g., CHMM and DAI), we seek to understand the damaging impact of the epistemic value on performance. Assuming the following definition for the epistemic value:
\begin{align*}
\bm{EV} &= \mathbb{E}_{P(o_\tau|s_\tau)Q(s_\tau | \pi)}\big[\ln P(s_\tau|o_\tau) - \ln Q(s_\tau | \pi) \big],
\end{align*}
we set up two experiments (c.f., Appendix C for details) in which all the disributions are stored in tables. In other words, the following categorical distributions are represented using matrices: $P(o_\tau|s_\tau)$, $Q(s_\tau | \pi)$, and $P(s_\tau|o_\tau)$.

\begin{figure}
     \centering
     \begin{subfigure}[b]{0.45\textwidth}
     	\begin{center}
		\begin{tikzpicture}[square/.style={regular polygon,regular polygon sides=4}]
		\node at (0,0) {\includegraphics[scale=0.5]{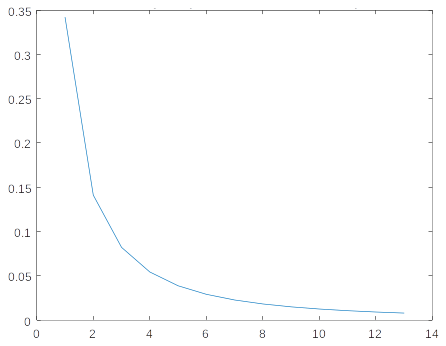}};
		\node[rotate=90] at (-4.3,0) {\small Epistemic Value};
		\node at (0.2,-3.5) {\small Similarity of posteriors: $P(s_\tau|o_\tau)$ and $Q(s_\tau|\pi)$};
    	\end{tikzpicture}
		\end{center}
     \end{subfigure}
     \begin{subfigure}[b]{0.45\textwidth}
     	\begin{center}
		\begin{tikzpicture}[square/.style={regular polygon,regular polygon sides=4}]
		\node at (0,0) {\includegraphics[scale=0.5]{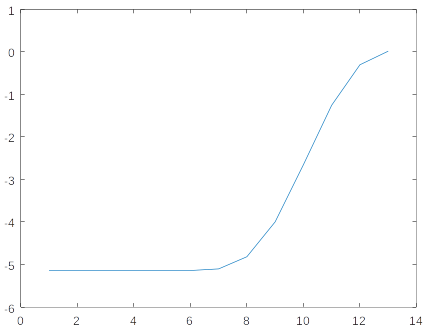}};
		\node[rotate=90] at (-4.3,0) {\small Epistemic Value};
		\node at (0.05,-3.5) {\small Similarity of posteriors: $P(s_\tau|o_\tau)$ and $Q(s_\tau|\pi)$};
    	\end{tikzpicture}
		\end{center}
     \end{subfigure}
        \caption{The left-most figure shows the result of the first experiment where the likelihood was becoming more and more uniform. In this setting, the epistemic value encourages exploration, as maximising the epistemic value makes the true posterior $P(s_\tau | o_\tau)$ as different as possible to the approximate posterior $Q(s_\tau | \pi)$, leftward on x-axis. However, the right-most figure illustrates the result of the second experiment where the prior over states was changing. In this case, the epistemic value does not promote exploration, as maximising the epistemic value makes the true posterior $P(s_\tau | o_\tau)$ as similar as possible to the approximate posterior $Q(s_\tau | \pi)$, rightward on x-axis. Indeed, in this case, maximising the epistemic value causes information to be lost.}
        \label{fig:epistemic_value_experiments}
\end{figure}

In the first experiement, the prior over states $P(s_\tau)$ is fixed, and the likelihood $P(o_\tau|s_\tau)$ is becoming more and more uniform. While this is happening, the true posterior $P(s_\tau|o_\tau)$ becomes more similar to the approximate posterior $Q(s_\tau | \pi)$, and the epistemic value decreases; see left panel of Figure \ref{fig:epistemic_value_experiments}. This is the expected behaviour, and in this setting, the epistemic value encourages exploration.

In the second experiment, the likelihood $P(o_\tau|s_\tau)$ is fixed and has rather high entropy, while the prior over states $P(s_\tau)$ is shifted in one direction on the state axis. As a result of this, the true posterior $P(s_\tau|o_\tau)$ becomes more different to the approximate posterior $Q(s_\tau | \pi)$, but this causes the epistemic value to decrease; see Figure \ref{fig:epistemic_value_experiments}, right panel. This is a degenerate behaviour, as in this setting, the epistemic value discourages exploration.

To sum up, the expected free energy decomposition into epistemic and extrinsic value --- as presented in Equation (10) of \citet{Parr304782} --- seems to exhibit two very different behaviours depending on how the distributions are defined, c.f., Figure \ref{fig:epistemic_value_experiments}. This is particularly important in the deep active inference literature, which builds on this equation. For example, the graphs presented in Section \ref{ssec:degenerate_behaviour_and_EFE} indicate that the CHMM agent minimising expected free energy is focusing almost exclusively on the action ``down", i.e., it is not exploring, which leads to poor performance.

\section{Conclusion} \label{sec:conclusion}

In this paper, we challenged the common assumption that deep active inference is a solved problem, by highlighting the challenges that need to be resolved for the field to move forward. We reviewed eight approaches implementing deep active inference: (a) $DAI_{MC}$ by \citet{DeepAIwithMCMC}, (b) $DAI_{VPG}$ by \citet{DeepAI}, (c) $DAI_{RHI}$ by \citet{rood2020deep}, (d) $DAI_{HR}$ by \citet{sancaktar2020endtoend}, (e) $DAI_{FA}$ by \citet{DAI_Kai}, (f) $DAI_{POMDP}$ by \citet{DAI_POMDP}, (g) $DAI_{SSM}$ by \citet{ccatal2020learning}, and (h) the approach by \citet{schneider2022active} for which the code was not available online.

Overall, those approaches brought interesting ideas such as: using deep neural networks to predict the parameters of the distributions of interest, using Monte-Carlo tree search for planning in active inference, and using a bootstrap estimate of the expected free energy to train a critic network. Yet, we struggled to replicate some of the results claimed, e.g. training $DAI_{MC}$ on the animal AI environment, and we were unable to access code in some instances. Ideally, future research should draw inspiration from the open science framework, by making the code that produced the claimed results open source. Also, the definition of the expected free energy varied between papers. This suggests that additional research is required to clarify the definition and justification of the expected free energy both in tabular and deep active inference (c.f., Section \ref{sec:discussion} and Appendix C). To sum up, recent research on deep active inference \citep{DeepAIwithMCMC,DeepAI,rood2020deep,sancaktar2020endtoend,DAI_Kai,DAI_POMDP,ccatal2020learning} has made an important first step towards a complete deep active inference agent, but a number of details still need to be honed, e.g., the definition and derivation of the expected free energy, and reproducibilty of research. For more details, the reader is referred to Section \ref{sec:existing_research}.

After reviewing existing research, we tried to progressively implement a deep active inference agent. First, we produced a variational auto-encoder agent that takes random actions. This agent was able to learn a useful (latent) representation of the task. However, since the agent takes random actions, it was unable to solve the task. Then, we added the transition network to create the hidden Markov model (HMM) agent, which also takes random actions. The agent was able to learn a good represenation of the task and of its dynamics, but was not able to solve the task.

Next, we tried to incorporate a critic network into the approach leading to the critical hidden Markov model (CHMM). In this context, we experimented with several possible implementations of the expected free energy. We also tried to remove the information gain and simply predict the reward. Additionally, we implemented three types of action selection strategies, namely: best action according to the expected free energy, softmax sampling, and epsilon-greedy.

When the epsilon-greedy algorithm was used, only the agent maximising reward was able to solve the task.
However, the agent requires more training iterations than a simple deep Q-network to learn the right behaviour.
This may be explained by the fact that the CHMM not only has to learn to solve the task, but also learn the dynamics of the environment.
When softmax sampling was used, all the agents failed to solve the task. One of them perfoming even worse than an agent selecting random actions.
Lastly, when selecting the best action, only the reward maximising agent was able to solve the task.
Importantly, according to our experiments, the agent using the epsilon-greedy algorithm received the highest amount of cumulated rewards and learned to solve the task the fastest.
Additionally, the reward maximising agents properly solve the task, but the quality of their latent representation is not as good as that of an HMM agent.
This may be due to the fact that when performing reward maximising actions, the data available to learn the model of the environment lacks diversity, i.e., not enough exploration.

Next, we tried to incorporate a policy network into the approach leading to a complete deep active inference agent (DAI). As for the CHMM agent, we experiemented with several possible implementations of the expected free energy, tried to remove the information gain and simply predict the reward, and implemented three types of action selection strategies, namely: best action according to the expected free energy, softmax sampling, and epsilon-greedy algorithm. When the epsilon-greedy algorithm was used or the best action was selected, all agents failed to solve the task. When using softmax sampling, most of the agents were numerically unstable and crashed, and the remaining agents failed to solve the task. 

Finally, we compared the similarity of the representation learned by the layers of various models (e.g., deep Q-network, CHMM, DAI, etc...) using centered kernel alignment. This reveals that the DQN learns general features in its first few layers, and very specialised features in its last two layers. The VAE learns similar features to the DQN in the first layers, but differs from the DQN in the last two layers, reflecting the difference in learning objectives. Similarly, the HMM learns similar features to the DQN in the first layers, but differs from the representation learned by the DQN in the last two layers. Also, the mean and variance representations learned by the HMM are different from their VAE counterparts, which suggests that the transition network influences the latent representation of the model.

Additionally, when using the $\mathring{\epsilon}$-greedy algorithm for action selection, the representations learned by the CHMM maximising reward is closer to the DQN than the CHMM minimising expected free energy is to the DQN. Importantly, the critic network of the reward maximising CHMM retains more information from the encoder than the CHMM minimising expected free energy. When the best action (according to the critic network) is selected, the CHMM maximising reward and the CHMM minimising expected free energy learn very similar representations except for the variance layers of the transition and encoder network. While performing further inspection of those (variance) layers, we found that the transition network of the reward maximising CHMM is a lot more certain than the transition network of the CHMM minimising expected free energy. More precisely, the CHMM minimising expected free energy is only confident about the world transition when performing action down. This suggests that the CHMM minimising expected free energy always picks the action down, and does not gather enough data for the other actions. Visualising the distribution of actions selected as training progresses corroborates this story by showing that the agent minimising EFE almost exclusively picks action down. In contrast, the CHMM maximising reward, keeps on selecting the actions left and right, which enables it to successfully solve the task. The only difference between those two CHMMs is the epistemic value, which aims to make the outputs of the transition and encoder network as close as possible. Thus, the CHMM minimising expected free energy is picking a single action (down), and becomes an expert at predicting the future when selecting this action. This effectively makes the KL divergence between the output of the transition and encoder networks small. Additionally, when selecting the action down, the average reward is zero, while for all the other actions, the expected reward will be
negative. Therefore, if the CHMM has to stick to a single action to keep the KL divergence small, then the action down
is the most rewarding.

The same observations about the variance layers also applies to CHMMs using softmax sampling for action selection. While this may explain why the model optimising the expected free energy does not solve the task, it does not explain why the model maximising the reward cannot solve the task, and we can hypothesise that those results may be due to the softmax action selection. More precisely, if the values predicted by the critic network are very close to each other, then an agent using softmax sampling may perform random actions. Note, increasing the gain parameter may help the agent to differentiate between values close to each other.

In addition, the representational similarity between the DAI (maximising reward using softmax sampling) and DQN is very close to the representational similarity between a DQN with a CHMM (maximising reward using softmax sampling). Also, the DAI's policy and critic network learn similar representations, which indicates that the policy network is learning correctly. Thus, the fact that the DAI (maximising reward using softmax sampling) fails in the dSprites environment is likely due to the softmax action selection and not to the representation learned by the model.

Lastly, our investigations of the expected free energy often used in deep active inference \citep{DeepAIwithMCMC}, suggests that degenerate behaviour can arise from it, in certain situations. This could explain why adding epistemic value to our planning objective seems to have such a damaging impact on the agent's capacity to explore its environment and gain information. The use of this definition of the expected free energy in the deep learning context may be at the core of our difficulty getting deep active inference to work and may also explain some of the presentational uncertainties (e.g., additions of minus signs) found in the deep active inference literature.

To conclude, the field of deep active inference has benefited from a large variety of ideas from the reinforcement and deep learning literature. In the future, it would be valuable to provide an approach that satisfies the following five desirata: (i) the approach is complete, i.e., it is composed of an encoder, a decoder, a transition network, a policy network and (optionally) a critic network, (ii) the mathematics underlying the approach is errorless and consistent with the free energy principle, (iii) the implementation is consistent with the mathematics, (iv) the code is publicly available so that the correctness of the implementation can be verified and the results reproduced, and (v) the approach is able to solve tasks with a large input space, e.g. image-based tasks. We believe that such an approach will benefit the field of deep active inference by providing a strong and reproducible baseline against which future research could benchmark.

\acks{We thank Karl J. Friston, Thomas Parr, Lancelot Da Costa, and Zafeirios Fountas for useful discussions and feedback surronding this paper, as well as for pointing us towards important resources such as their new book and several papers.}

\vskip 0.2in
\bibliography{Challenges}

\appendix

\afterpage{%
\newgeometry{left=0.8cm, bottom=2cm, top=1cm}
\thispagestyle{empty}
\vspace{-0.5cm}
\section*{Appendix A: Notation}
\begin{table}[H]
\hspace{-1.4cm}
\begin{tabular}{ |c|c|  }
 \hline
 Symbol & Meaning\\
 \hline
 \hline
 $s_\tau$, $o_\tau$, $r_\tau$, $a_\tau$ & State, observation, reward and action at time step $\tau$, respectively.\\
 \hline
 \multirow{3}{*}{$o_\tau^r$, $\hat{s}_\tau^r$, $\mathring{o}^r_\tau$} & An observation in which a reward has been encoded as explained in Figure \ref{fig:encoding_rewards_in_image}, the hidden \\
 &state sampled from the encoder when feeding $o_\tau^r$ as input, and the observation reconstructed \\
 &by the decoder from the output of the transition network, respectively.\\
 \hline
 \multirow{2}{*}{$\hat{s}_\tau$, $\hat{o}_\tau$} & A state sampled from the encoder at time step $\tau$ when $o_\tau$ is provided as input, and the image \\
 &reconstructed by the decoder at time $\tau$ when using $\hat{s}_\tau$ as input, respectively.\\
 \hline
  $\mathring{o}_\tau$ & The observations at time step $\tau$ predicted by the transition network.\\
 \hline
 $s_{i:j}$, $o_{i:j}$, $a_{i:j}$ & Respectively, the set of states, observations, and actions between time step $i$ and $j$ (included).\\
 \hline
 \multirow{2}{*}{$\pi$, $\pi_\tau$, $\pi'$} & A policy, i.e. a sequence of actions, the action prescribed by the policy at time step $\tau$, and \\
 & another policy whose size is smaller or equal than the size of $\pi$, respectively.\\
 \hline
 $\mathcal{A}$, $\Pi$ & The set of possible actions, and the set of possible policies, respectively.\\
 \hline
 $\mathbb{A}_s$ & The set of all anscestors of a node $s$.\\
 \hline
 $\#A, \#C, \#S, \#O$ & The number of actions, channels, states and observations, respectively.\\
 \hline
 $\mathcal{Q}_{\theta_a}$, $\hat{\mathcal{Q}}_{\hat{\theta}_a}$ & The Q-network parameterised by $\theta_a$, and the target network parameterised by $\hat{\theta}_a$\\
 \hline
 $\mathcal{G}_{\theta_a}$, $\hat{\mathcal{G}}_{\hat{\theta}_a}$ & The critic network parameterised by $\theta_a$, and the target network parameterised by $\hat{\theta}_a$.\\
 \hline
 $\mathcal{E}_{\phi_s}$, $\mathcal{D}_{\theta_o}$ & The encoder network parameterised by $\phi_s$, and the decoder network parameterised by $\theta_o$.\\
 \hline
 $\mathcal{P}_{\phi_a}$ & The policy network parameterised by $\phi_a$.\\
 \hline
 $\mathcal{T}_{\theta_s}$, $\mathcal{T}_{\theta_o}$ & The transition network parameterised by $\theta_s$ or $\theta_o$, respectively.\\
 \hline
 $\theta$, $\phi$ & All the parameters of the generative model, and the variational distribution, respectively.\\
 \hline
 $a$, $b$, $c$, $d$ & Four hyperparameters involved in the computation of $\omega_t$.\\
 \hline
 $N(s_\tau, a_\tau)$ & The number of times action $a_\tau$ was explored in state $s_\tau$.\\
 \hline
 $t$, $\gamma$ & The present time step, and the discount factor, respectively.\\
 \hline
 $\zeta$, $\psi$ & The precision of the prior over actions, and the precision of the prior preferences.\\
 \hline
 $\epsilon$, $\hat{\epsilon}$ & The random variable used in the re-parameterisation trick, and a sample of epsilon.\\
 \hline
 $\mathring{\epsilon}$ & The probability of selecting a random actions when using the $\mathring{\epsilon}$-greedy algorithm.\\
 \hline
 $T_{dec}$ & A hyperparameter defining the threshold value corresponding to a clear winner during MCTS.\\
 \hline
 \multirow{2}{*}{$G(\pi)$, $G_\tau(\pi)$} & The expected free energy (EFE) of policy $\pi$, and the EFE received at time \\
 & step $\tau$ when following policy $\pi$, respectively.\\
 \hline
 \multirow{2}{*}{$\bar{\bm{G}}_s$, $\bm{G}_s^{\text{aggr}}$, $\bm{G}_s$, $\bm{N}_s$} & The average EFE, the aggregated EFE, the EFE, and the number of visits \\
 & of a node $s$, respectively.\\
 \hline
 $\mu_o, \sigma_o, \mu_a, \sigma_a$ & The mean and variance vectors predicted by the encoder and policy networks of the $DAI_{FA}$.\\
 \hline
 $\mu$, $\sigma$ & The mean and variance of the Gaussian distribution over $s_t$ predicted by the encoder.\\
 \hline
 $\mathring{\mu}$, $\mathring{\sigma}$ & The mean and variance of the Gaussian distribution over $s_{t+1}$ predicted by the transition.\\
 \hline
 $\hat{\mu}$, $\hat{\sigma}$ & The mean and variance of the Gaussian distribution over $s_{t+1}$ predicted by the encoder.\\
 \hline
 $\hat{\pi}$ & The parameters of the categorical distribution over $a_t$ predicted by the policy network.\\
 \hline
 $\omega_t$ & The top-down attention parameter modulating the precision of the transition mapping.\\
 \hline
 $[\,\text{condition}\,]$ & An indicator function that equals one if the condition is satisfied and zero otherwise.\\
 \hline
 $\sigma[\,\bigcdot\,]$ & The softmax function.\\
 \hline
 $\text{Cat}(x; \phi_x)$ & A categorical distribution over $x$ parameterised by $\phi_x$.\\
 \hline
 $\text{Bernoulli}(x; \phi_x)$ & A Bernoulli distribution over $x$ parameterised by $\phi_x$.\\
 \hline
 $\MultiBernoulli(x; \phi_x)$ & A product of Bernoulli distributions over $x$ parameterised by $\phi_x$.\\
 \hline
 \multirow{2}{*}{$\mathcal{N}(x; \mu_x, \sigma_x)$} & A multivariate Gaussian over $x$ parameterised by a mean vector $\mu_x$, and \\
 &a diagonal covariance matrix whose diagonal elements are $\sigma_x$.\\
 \hline
 $X \overset{i}{\rightarrow} Y$, $X \overset{m}{\rightarrow} Y$ & $X$ is fed as \textit{input} to $Y$, and the \textit{mean} of the distribution predicted by $X$ is fed as input to $Y$.\\
 \hline
 $X \overset{s}{\rightarrow} Y$, $X \rightarrow Y$ & a \textit{sample} from the distribution predicted by $X$ is fed as input to $Y$, and $X$ outputs $Y$.\\
 \hline
\end{tabular}
\caption{Notation of Sections \ref{sec:existing_research} and \ref{sec:build_dai}.}
\label{tab:notation}
\end{table}
}
\newpage

\section*{Appendix B: $DAI_{MC}$ discrepancies between the paper and the code}

In this section, we focus on the authors' implementation of $DAI_{MC}$ available on GitHub: \url{https://github.com/zfountas/deep-active-inference-mc/}. First, according to a personal communication with one of the authors, the code available on GitHub (on the 6th of June 2022) is not the same as the one used to run the experiments of the paper. Below, we describe the discrepancies between the paper and the code. For example, the computation of $\omega_t$ in the paper is as follows:
\begin{align*}
\omega_t = \frac{a}{1 + \exp(-\frac{b - D_t}{c})} + d,
\end{align*}
while the code uses the following formula:
\begin{align*}
\omega_t = a\times \Bigg(1 - \frac{1}{1 + \exp\big(-\frac{D_t - b}{c}\big)}\Bigg) + d.
\end{align*}
Also, the paper states that MCTS is perfomed to compute the prior over policies during training. However, in the code, MCTS is only used when testing the model, i.e., no MCTS when training the agent. Additionally, the paper states that actions are selected by sampling from:
\begin{align*}
\tilde{P}(a_t) = \frac{N(\hat{s}_t, a_t)}{\sum_{\hat{a}_t}N(\hat{s}_t, \hat{a}_t)}.
\end{align*}
However, the code selects an entire sequence of actions $\pi = (\mathring{a}_t, \mathring{a}_{t+1}, ..., \mathring{a}_{t+n})$ recursively from the root node in the tree. At each step in the recursion, the node with the highest number of visits $\mathring{a}_\tau$ is selected. Then, actions cancelling each other are removed from the sequence, e.g., if $a_\tau = LEFT$ and $a_{\tau+1} = RIGHT$ then both actions are removed from the sequence. This procedure generates a new sequence of actions $\pi'$ of equal or smaller length. Finally, the entire sequence of actions $\pi'$  is performed in the environment. This avoids the repetition of the planning process for each action-perception cycle (saving computational time), however, this also requires domain knowledge (to remove actions that cancel each other out).

\begin{figure}
	\begin{center}
	\begin{tikzpicture}[square/.style={regular polygon,regular polygon sides=4}]
	\node at (0,0) {\includegraphics[scale=0.4]{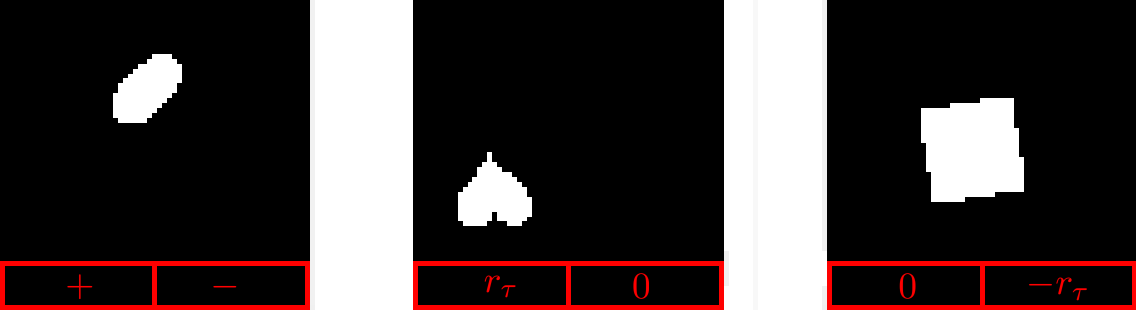}};
    \end{tikzpicture}
	\end{center}
  \caption{This figure illustrates how the reward $r_\tau \in [-1, 1]$ is encoded in image $o_\tau$. On the left, the plus and minus signs shows where the reward will be encoded in the image if the reward is positive or negative, respectively. In the middle, a positive reward is being encoded on the left side of the image. On the right, a negative reward is being encoded on the right of the image.}
   \label{fig:encoding_rewards_in_image}
\end{figure}

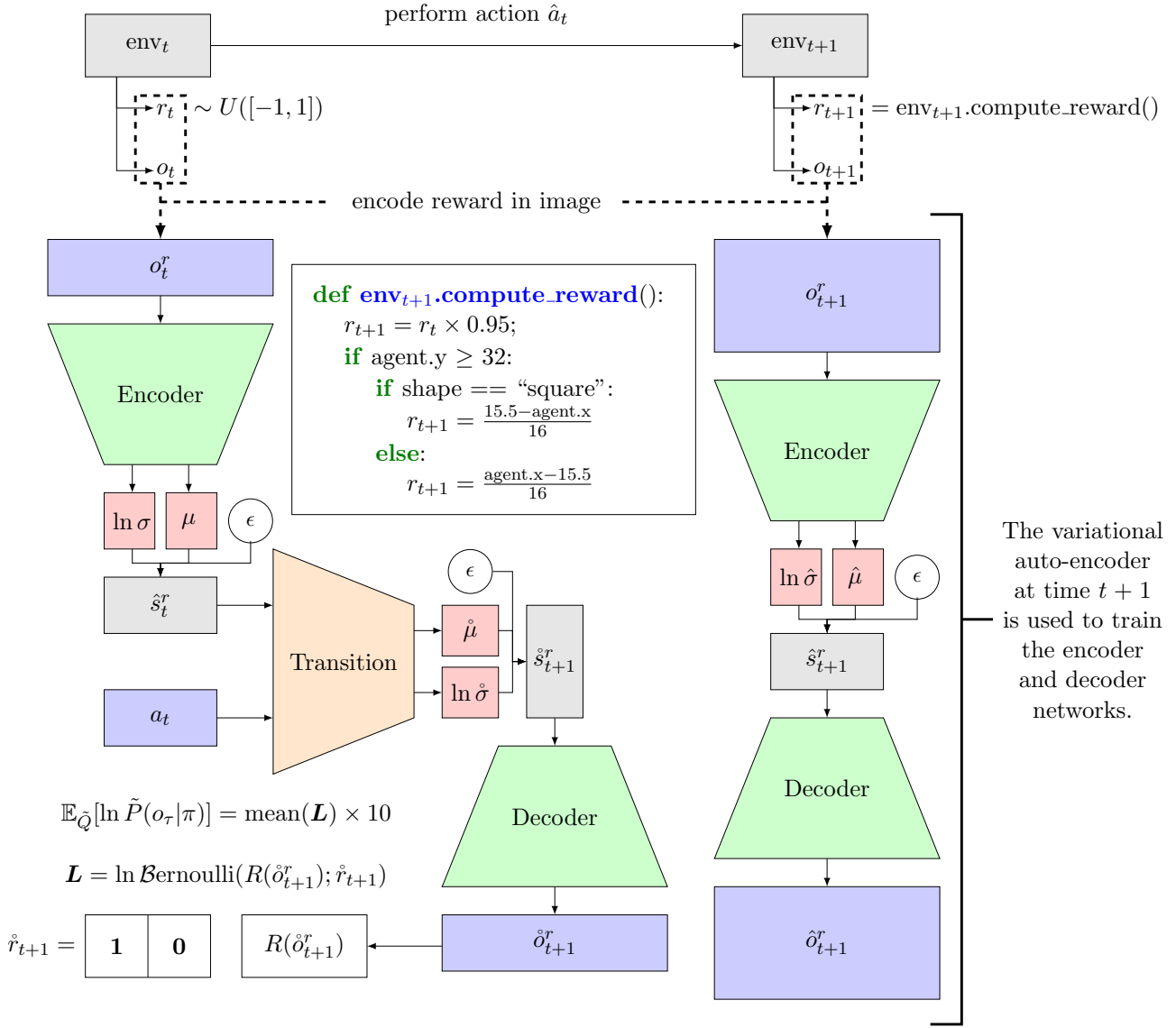
\begin{figure}
	\begin{center}\vspace{-1cm}
	\resizebox{1\textwidth}{!}{%
	\begin{tikzpicture}[square/.style={regular polygon,regular polygon sides=4}]
		\draw[fill=gray!20!white] (-1,-0.5) rectangle (1,0.5);
		\node at (0, 0) {$\text{env}_t$};
		\draw[fill=gray!20!white] (9.5,-0.5) rectangle (11.5,0.5);
		\node at (10.5, 0) {$\text{env}_{t+1}$};
		\node at (5.25, 0.5) {$\text{perform action } \hat{a}_t$};
		\draw[-latex] (1,0) -- (9.5,0);
		\draw[-latex] (-0.5,-0.5) -- (-0.5,-1) -- (0.1,-1);
		\draw[-latex] (-0.5,-0.5) -- (-0.5,-2) -- (0.1,-2);

		\draw[-latex] (10,-0.5) -- (10,-1) -- (10.6,-1);
		\draw[-latex] (10,-0.5) -- (10,-2) -- (10.6,-2);

		\node[anchor=west] at (0, -1) {$r_t \,\,\, \sim U([-1, 1])$};
		\node[anchor=west] at (0, -2) {$o_t$};
		\draw[very thick, dashed] (-0.2,-0.8) rectangle (0.6,-2.2);
		\draw[very thick, dashed, -latex] (0.2,-2.2) -- (0.2,-3.1);

		\node[anchor=west] at (10.5, -1) {$r_{t+1} \,\, = \text{env}_{t+1}.\text{compute\_reward}()$};
		\node[anchor=west] at (10.5, -2) {$o_{t+1}$};
		\draw[very thick, dashed] (10.3,-0.8) rectangle (11.4,-2.2);
		\draw[very thick, dashed, -latex] (10.85,-2.2) -- (10.85,-3.1);

		\draw[very thick, dashed] (10.85,-2.5) -- (7.5,-2.5);
		\draw[very thick, dashed] (0.2,-2.5) -- (2.9,-2.5);
		\node at (5.25,-2.5) {encode reward in image};

		\pic[rotate=-90,xshift=4cm,yshift=19.85cm, scale=0.9]{vae=$o_{t+1}^r$/$\hat{\mu}$/$\ln \hat{\sigma}$/$\hat{s}_{t+1}^r$/$\hat{o}_{t+1}^r$};

		\draw[very thick] (12.5,-2.7) -- (13,-2.7) -- (13,-15.65) -- (12.5,-15.65);
		\draw[very thick] (13,-9.175) -- (13.5,-9.175);
		\node[anchor=west,text width=3cm,align=center] at (13.3,-9.175) {The variational auto-encoder at time $t+1$ is used to train the encoder and decoder networks.};
		
		\pic[rotate=-90, xshift=4cm,yshift=9.2cm, scale=0.9]{fountas=$o_t^r$/$\mu$/$\ln \sigma$/$\hat{s}_t^r$/$\mathring{o}_{t+1}^r$};

		\draw (1.5,-13.9) rectangle (3.5,-14.9);
		\draw (-1,-13.9) rectangle (1,-14.9);
		\draw (0,-13.9) -- (0,-14.9);
		\draw[-latex] (4.7,-14.4) -- (3.5,-14.4);
		\node at (-1.7,-14.4) {$\mathring{r}_{t+1} =$};
		\node at (0.5,-14.4) {$\bm{0}$};
		\node at (-0.5,-14.4) {$\bm{1}$};
		\node at (2.5,-14.4) {$R(\mathring{o}_{t+1}^r)$};

		\node at (1.25,-13.3) {$\bm{L} = \ln \mathcal{B}\text{ernoulli}(R(\mathring{o}_{t+1}^r); \mathring{r}_{t+1})$};
		\node at (1.25,-12.3) {$\mathbb{E}_{\tilde{Q}}[\ln \tilde{P}(o_\tau|\pi)] = \text{mean}(\bm{L}) \times 10$};

		\draw (2.3,-3.5) rectangle (8.75,-7.5);
		\node[anchor=west] at (2.5,-4) {{\color{GreenCode}\textbf{def}} {\color{BlueCode}\textbf{env$_{t+1}$.compute\_reward}}():};
		\node[anchor=west] at (3,-4.5) {$r_{t+1} = r_t \times 0.95;$};
		\node[anchor=west] at (3,-5) {{\color{GreenCode}\textbf{if}} agent.y $\geq$ 32:};
		\node[anchor=west] at (3.5,-5.5) {{\color{GreenCode}\textbf{if}} shape == ``square":};
		\node[anchor=west] at (4,-6) {$r_{t+1} = \frac{15.5 - \text{agent.x}}{16}$};
  		\node[anchor=west] at (3.5,-6.5) {{\color{GreenCode}\textbf{else}}:};
		\node[anchor=west] at (4,-7) {$r_{t+1} = \frac{\text{agent.x} - 15.5}{16}$};
    \end{tikzpicture}
    } %
	\end{center}
  \caption{This figure illustrates the computation of $\mathbb{E}_{\tilde{Q}}[\ln \tilde{P}(o_\tau|\pi)]$ in (the code of) \citet{DeepAIwithMCMC}. The environment at time $t$ provides the agent with an image $o_t$ and a reward $r_t$ randomly sampled from the interval $[-1;1]$. Then, action $\hat{a}_t$ is performed in the environment and the agent observes an image $o_{t+1}$ and a reward $r_{t+1}$, where $r_{t+1}$ is computed according to the function presented in the center of the image. Next, the reward at time $t$ and $t+1$ are encoded in the images received at time $t$ and $t+1$, respectively, c.f., Figure \ref{fig:encoding_rewards_in_image} for details about the encoding. The encoded image at time $t+1$ (i.e., $o^r_{t+1}$) is then fed into the encoder, the re-parameterisation trick is then used to sample a state from the variational posterior. This state is fed into the decoder which tries to reconstruct the image inputed into the encoder. Once $\hat{o}^r_{t+1}$ has been computed, the weights of the encoder and decoder are learned using back-propagation. On the other hand, the encoded image at time $t$ (i.e., $o^r_t$) is used to compute $\mathbb{E}_{\tilde{Q}}[\ln \tilde{P}(o_\tau|\pi)]$. More precisely, the $o^r_t$ is fed into the encoder, and a state is sampled from the variational posterior $Q_{\phi_s}(s_t)$. This state is then fed as input into the transition network along with the action prescribed by $\pi$ at time $t$, i.e., $a_t$. A state at time $t+1$ can then be sampled from the distribution predicted by the transition network. This state is then inputed into the decoder, which outputs $\mathring{o}^r_{t+1}$. Next, a matrix (i.e., $\mathring{r}_{t+1}$) encoding the maximum reward that the agent can gather is used as a parameter of a Bernoulli distribution to compute the logarithm of the probability (i.e., $\bm{L}$) of the first three rows of $\hat{o}^r_{t+1}$, i.e., $R(\mathring{o}^r_{t+1})$. The mean of $\bm{L}$ is then computed and is multiplied by ten to obtain $\mathbb{E}_{\tilde{Q}}[\ln \tilde{P}(o_\tau|\pi)]$.}
   \label{fig:computation_of_extrinsic_value}
\end{figure}

Additionally, in the paper, experiments are run on both the dSprites environment and the animal AI environment. However, the code does not allow the replication of the results on the animal AI environment, i.e., the code handling the animal AI environment has been removed. In addition, the evaluation of the expected free energy is non trivial (see below) and the details are not discussed in the paper. Before explaining how the terms of the EFE are computed, we introduce notation that allows us to express those computational steps concisely. For example, we note: 
$$o^r_t \overset{i}{\rightarrow} \text{Encoder} \overset{s}{\rightarrow} \text{Transition} \overset{m}{\rightarrow} \mathring{s}^r_{t+1},$$
meaning that $o^r_t$ is used as \textit{input} ($\overset{i}{\rightarrow}$) for the encoder, then a state is \textit{sampled} ($\overset{s}{\rightarrow}$) from the distribution predicted by the encoder and used as input for the transition network, finally, the \textit{mean} ($\overset{m}{\rightarrow}$) of the distribution predicted by the transition network is used as a maximum aposteriori estimate of $\mathring{s}^r_{t+1}$. Note, the transition network takes two inputs (i.e., a state and an action), when using our concise notation we implicitly assume that the actions prescribed by the policy\footnote{$\pi$ is the policy for which the expected free energy is being computed.} $\pi$ are provided as input to the transition network. Also, for each time step $\tau$, the reward $r_\tau$ collected by the agent is encoded in the pixels of the image $o_\tau$ as explained in Figure \ref{fig:encoding_rewards_in_image}, leading to a new image $o_\tau^r$. As illustrated on the right of Figure \ref{fig:computation_of_extrinsic_value}, the encoder/decoder networks are trained to predict the resulting images $o_\tau^r$. The computation of the first term in equation \eqref{eq:efe_rearranged_fountas} is illustrated on the left of Figure \ref{fig:computation_of_extrinsic_value}. Concisely, we have:
$$o_t^r \overset{i}{\rightarrow} \text{Encoder} \overset{s}{\rightarrow} \text{Transition} \overset{s}{\rightarrow} \text{Decoder} \overset{m}{\rightarrow} \mathring{o}_{t+1}^r.$$
Next, a matrix ($\mathring{r}_{t+1}$) encoding the maximum reward that the agent can gather is used as parameter of Bernoulli distributions to compute the logarithm of the probability (i.e., $\bm{L}$) of the three first rows of the reconstructed image $\mathring{o}_{t+1}^r$. Note, as explained in Figure \ref{fig:encoding_rewards_in_image}, the first three rows contain the predicted reward obtained at time $t+1$. Finally, the mean of $\bm{L}$ is then computed and is multiplied by ten to get $\mathbb{E}_{\tilde{Q}}[\ln \tilde{P}(o_\tau|\pi)]$. Similarly, the computation of $H[Q(s_\tau|\pi)]$ proceeds as follows:
$$o_t^r \overset{i}{\rightarrow} \text{Encoder} \overset{s}{\rightarrow} \text{Transition} \rightarrow \mathring{\mu}, \ln \mathring{\sigma},$$
where $Q(s_\tau|\pi)$ is equated with $\mathcal{N}(s_\tau;\mathring{\mu}, \mathring{\sigma})$, and an analytical solution is used to compute the entropy of $Q(s_\tau|\pi)$. Next, the computation of $H[Q(o_\tau|s_\tau, \pi)]$ proceeds  as follows:
$$o_t^r \overset{i}{\rightarrow} \text{Encoder} \overset{s}{\rightarrow} \text{Transition} \overset{s}{\rightarrow} \text{Decoder} \overset{m}{\rightarrow} \mathring{o}^r_{t+1},$$
where observation $\mathring{o}^r_{t+1}$ is equated to the parameters of the Bernoulli distribution $Q(o_\tau|s_\tau, \pi)$, and an analytical solution is used to compute $H[Q(o_\tau|s_\tau, \pi)]$. Surprisingly, another observation $\mathring{o}^r_{t+1}$ sampled exactly as before is equated to the parameters of the Bernoulli distribution $Q(o_\tau|s_\tau,\theta,\pi)$, and the same analytical solution is used to compute $H[Q(o_\tau|s_\tau,\theta,\pi)]$. Finally, $H[Q(s_\tau|o_\tau, \pi)]$ is computed by feeding $\mathring{o}^r_{t+1}$ back into the encoder to obtain the mean and log-variance of the Gaussian distribution $Q(s_\tau|o_\tau, \pi)$, and the analytical solution for the entropy of a Gaussian is used to compute $H[Q(s_\tau|o_\tau, \pi)]$.

In summary, two samples of $\mathring{o}^r_{t+1}$ (sampled as described in Figure \ref{fig:encoding_rewards_in_image}) have been equated to the parameters of two diferent distributions, i.e., $Q(o_\tau|s_\tau,\theta,\pi)$, and $Q(o_\tau|s_\tau, \pi)$. Additionally, a third sample of $\mathring{o}^r_{t+1}$ (sampled in the same way) has also been used as input to the distribution $\tilde{P}(o_\tau|\pi)$. Lastly, while \citet{DeepAIwithMCMC} defines the EFE as in \eqref{eq:efe_rearranged_fountas}, the code turns a plus into a minus, leading to the following definition of the EFE:
\begin{align*}
G_{\tau}(\pi) = &- \mathbb{E}_{\tilde{Q}}\Big[ \ln \tilde{P}(o_\tau|\pi)\Big]\nonumber\\
&\,{\color{red}-} \,\, \mathbb{E}_{Q(\theta|\pi)}\Big[ \mathbb{E}_{Q(o_\tau|\theta,\pi)}\big[H[Q(s_\tau|o_\tau, \pi)]\big] - H[Q(s_\tau|\pi)] \Big]\nonumber\\
&+ \mathbb{E}_{Q(\theta|\pi)Q(s_\tau|\theta,\pi)}\Big[H[Q(o_\tau|s_\tau,\theta,\pi)] \Big] - \mathbb{E}_{Q(s_\tau|\pi)}\Big[ H\big[ Q(o_\tau|s_\tau, \pi) \big] \Big],
\end{align*}
where the red minus was a plus.

\section*{Appendix C: Analysis of the epistemic value}

\begin{figure}
	\begin{center}
		\includegraphics[scale=0.4]{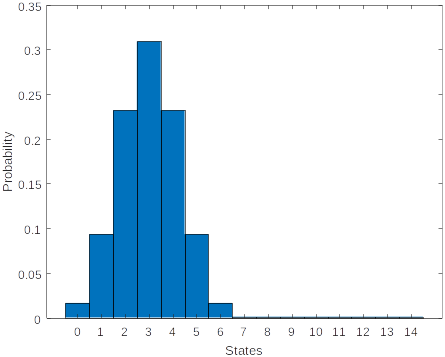}
	\end{center}
  \caption{This figure illustrates the approximate posterior $Q(s_\tau | \pi)$, which is distributed according to a binomial distribution corresponding to 6 trials with a probability of success of 0.5.}
   \label{fig:approximate_posterior_fig}
\end{figure}

\begin{figure}
     \centering
     \begin{subfigure}[b]{0.4\textwidth}
         \centering
			\includegraphics[scale=0.3]{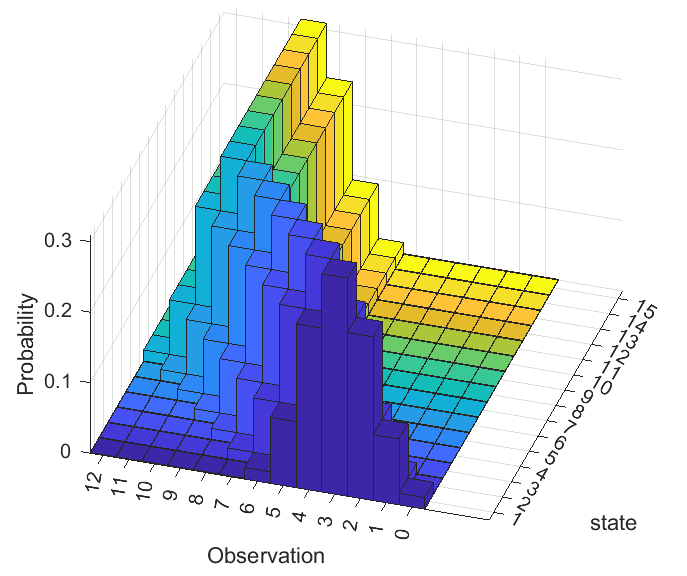}
     \end{subfigure}
     \begin{subfigure}[b]{0.4\textwidth}
         \centering
			\includegraphics[scale=0.3]{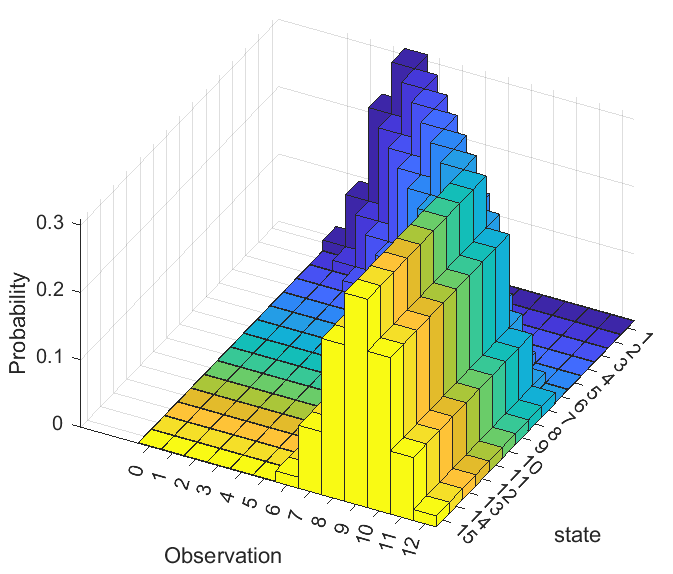}
     \end{subfigure}
        \caption{This figure provides two views of the same likelihood mapping $P(o_\tau | s_\tau)$, i.e., one from the front and one from behind. The likelihood was created by sliding a binomial distribution (corresponding to 6 trials with a probability of success of 0.5) across the observation axis, each time the state increases by one. Finally, when the binomial reached its most extreme position, it stays the same for the remaining states.}
        \label{fig:likelihood_distribution}
\end{figure}

\begin{figure}
     \centering
     \begin{subfigure}[b]{0.3\textwidth}
         \centering
			\includegraphics[scale=0.23]{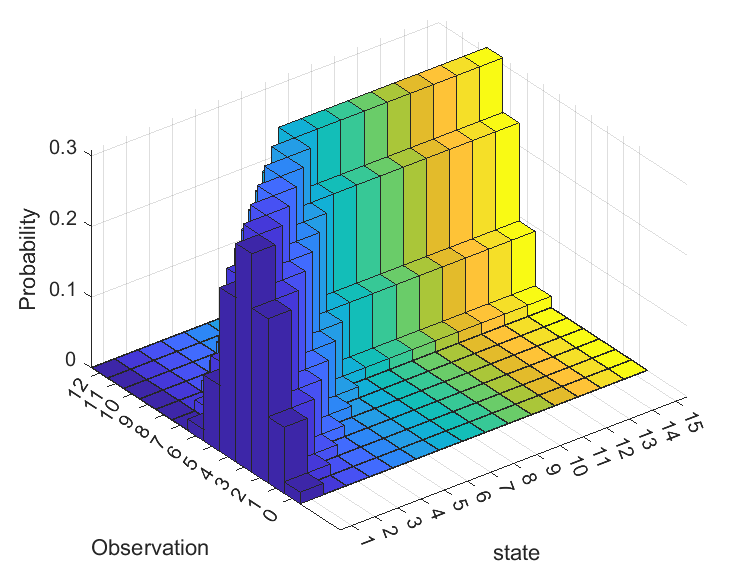}
     \end{subfigure}
     \begin{subfigure}[b]{0.3\textwidth}
         \centering
			\includegraphics[scale=0.23]{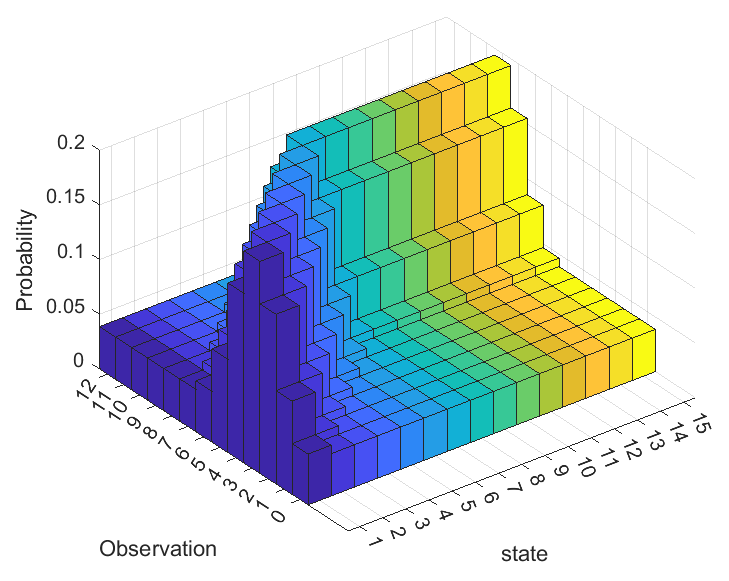}
     \end{subfigure}
     \begin{subfigure}[b]{0.3\textwidth}
         \centering
			\includegraphics[scale=0.23]{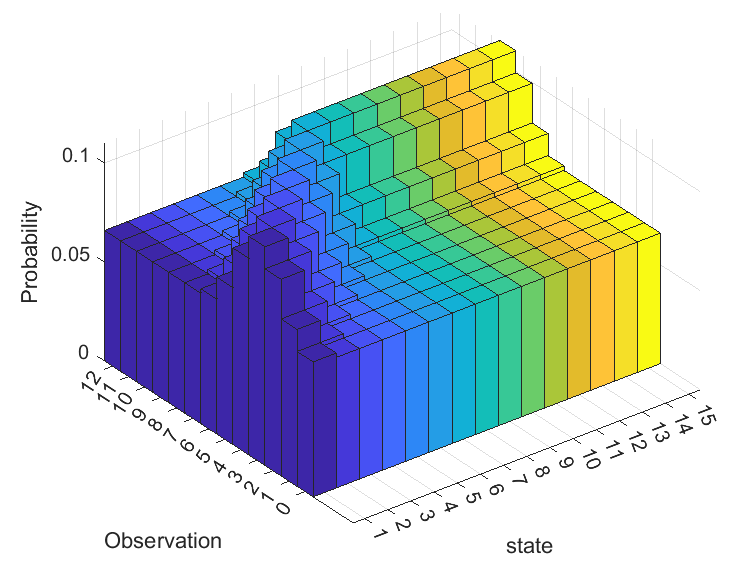}
     \end{subfigure}
        \caption{During the first experiement, we changed the likelihood mapping $P(o_\tau | s_\tau)$ by making it more and more uniform. This change is shown in the figure from left to right.}
        \label{fig:likelihood_change}
\end{figure}

\begin{figure}
     \centering
     \begin{subfigure}[b]{0.3\textwidth}
         \centering
			\includegraphics[scale=0.35]{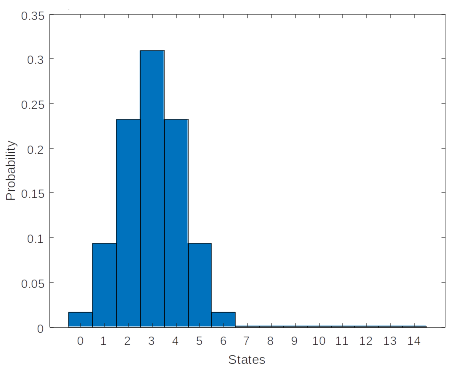}
     \end{subfigure}
     \begin{subfigure}[b]{0.3\textwidth}
         \centering
			\includegraphics[scale=0.35]{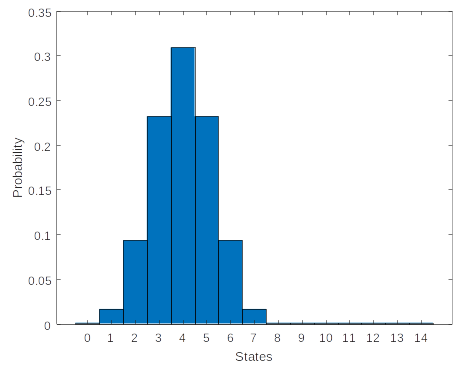}
     \end{subfigure}
     \begin{subfigure}[b]{0.3\textwidth}
         \centering
			\includegraphics[scale=0.35]{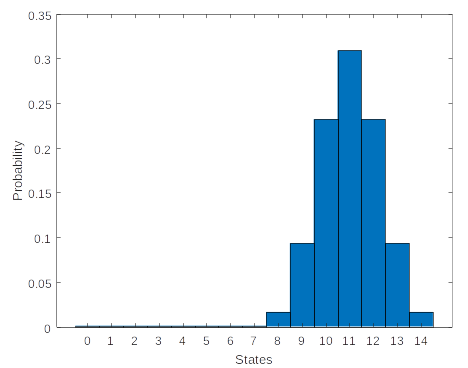}
     \end{subfigure}
        \caption{During the second experiement, we changed the prior over states $P(s_\tau)$ by making it more and more different to the approximate posterior $Q(s_\tau | \pi)$. This change is shown in the figure from left to right.}
        \label{fig:prior_state_change}
\end{figure}

In this appendix, we study the expected free energy decomposition into epistemic and extrinsic value as given in Equation (10) of \cite{Parr304782}. A particular reason for being interrested in this formulation is that it is the version of the expected free energy that has the most influenced the deep active inference approaches, such as \citet{DeepAIwithMCMC}. Starting with the definition of the expected free energy, see Equation \eqref{eq:efe_practice} in the main-body:
\begin{align*}
G_\tau(\pi) &= \mathbb{E}_{P(o_\tau|s_\tau)Q(s_\tau | \pi)}\big[\ln Q(s_\tau | \pi) - \ln P(o_\tau, s_\tau)\big]\nonumber\\
&= \mathbb{E}_{P(o_\tau|s_\tau)Q(s_\tau | \pi)}\big[\ln Q(s_\tau | \pi) - \ln P(s_\tau|o_\tau)- \ln P(o_\tau)\big]\nonumber\\
&= - \underbrace{\mathbb{E}_{P(o_\tau|s_\tau)Q(s_\tau | \pi)}\big[\ln P(s_\tau|o_\tau) - \ln Q(s_\tau | \pi) \big]}_{\text{Epistemic value}} - \underbrace{\mathbb{E}_{P(o_\tau|s_\tau)Q(s_\tau | \pi)}\big[\ln P(o_\tau)\big]}_{\text{Extrinsic value}}\nonumber
\end{align*}
In the rest of this appendix, we will focus on the epistemic value and report two experiments. In the first experiment, the prior over states (c.f., left-most graph in Figure \ref{fig:prior_state_change}) is equal to the approximate posterior (c.f., Figure \ref{fig:approximate_posterior_fig}), and the likelihood (c.f., Figure \ref{fig:likelihood_distribution}) is becoming more and more uniform (c.f., Figure \ref{fig:likelihood_change}). Note, if the likelihood becomes uniform, then the true posterior $P(s_\tau | o_\tau)$ becomes equal to the prior over states $P(s_\tau)$, i.e.,
$$P(s_\tau | o_\tau) \propto P(o_\tau | s_\tau ) P(s_\tau) =  \frac{1}{|o_\tau|} P(s_\tau),$$
where $|o_\tau|$ is the number of observations at time step $\tau$, and after renormalisation $P(s_\tau | o_\tau) = P(s_\tau)$. The left-most graph of Figure \ref{fig:epistemic_value_experiments} shows that the epistemic value decreases as the likelihood becomes more uniform, i.e., the epistemic value decreases as the true posterior $P(s_\tau | o_\tau)$ becomes more similar to the approximate posterior $Q(s_\tau | \pi)$. This behaviour is to be expected, as the epistemic value is bigger when the true posterior $P(s_\tau | o_\tau)$ and the approximate posterior $Q(s_\tau | \pi)$ are more different. Thus, maximising epistemic value will promote exploration and information gain.

\begin{figure}
     \centering
     \begin{subfigure}[b]{0.3\textwidth}
         \centering
			\includegraphics[scale=0.30]{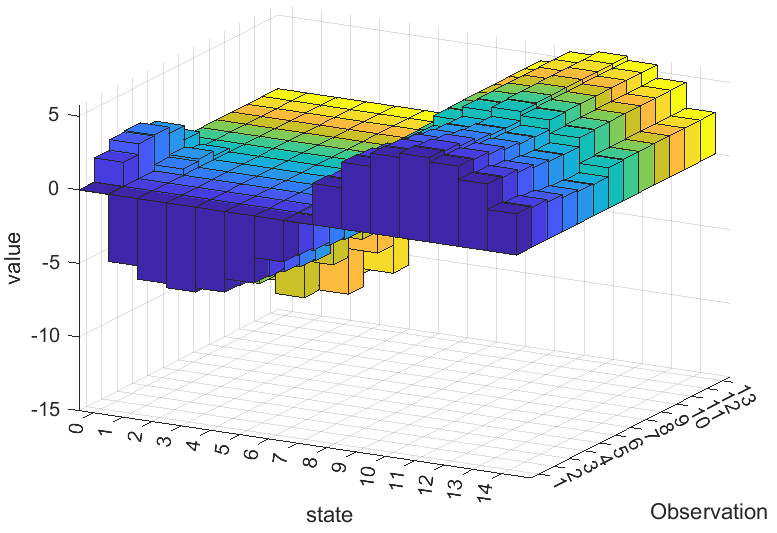}
     \end{subfigure}
     \begin{subfigure}[b]{0.3\textwidth}
         \centering
			\includegraphics[scale=0.23]{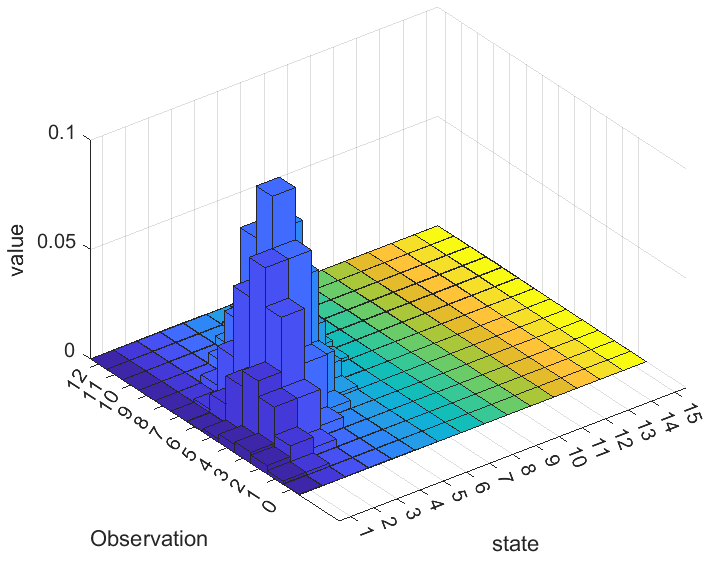}
     \end{subfigure}
     \begin{subfigure}[b]{0.3\textwidth}
         \centering
			\includegraphics[scale=0.23]{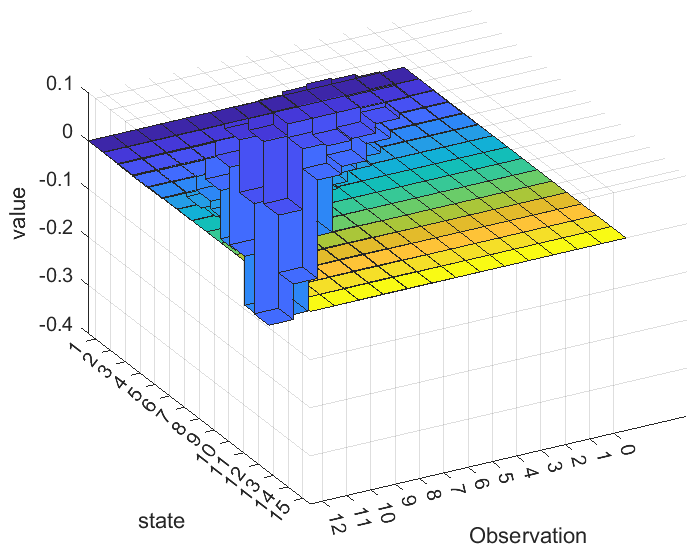}
     \end{subfigure}
        \caption{To compute the epistemic value, we first computed the difference between the logarithm of the true posterior $\ln P(s_\tau | o_\tau)$ and the logarithm of the approximate posterior $\ln Q(s_\tau | \pi)$ for all the values taken by the observation $o_\tau$ (c.f., left-most figure). Then, we computed the joint distribution $P(o_\tau | s_\tau)Q(s_\tau | \pi)$ used in the expectation (c.f., middle figure). Next, we compute the element-wise product between the matrices illustrated in the left-most and middle figures (c.f., right-most figure). Finally, the epistemic value is obtained by summing up all the elememts of the element-wise product. In this instance, the epistemic value will be strongly negative.}
        \label{fig:epistemic_value_computation}
\end{figure}

In the second experiment, the likelihood has high entropy (c.f., right-most graph in Figure \ref{fig:likelihood_change}), and the prior over states is shifting from left to right (c.f., Figure \ref{fig:prior_state_change}). When the prior over states $P(s_\tau)$ and the approximate posterior $Q(s_\tau | \pi)$ are different, the joint distribution $P(o_\tau | s_\tau)Q(s_\tau | \pi)$ will be more similar to the approximate posterior $Q(s_\tau | \pi)$ than it is to the true posterior $P(s_\tau | o_\tau)$. Indeed, as the likelihood is almost uniform, the true posterior $P(s_\tau | o_\tau)$ will almost be equal to the prior over states $P(s_\tau)$. At the same time, as the likelihood is almost uniform, it will not have much impact on the joint distribution $P(o_\tau | s_\tau)Q(s_\tau | \pi)$ and the approximate posterior $Q(s_\tau | \pi)$ will dominate. To sum up, when the prior over states $P(s_\tau)$ and the approximate posterior $Q(s_\tau | \pi)$ are different:
\begin{itemize}
\item the true posterior will almost be equal to the prior over states $P(s_\tau | o_\tau) \approx P(s_\tau)$
\item the joint distribution $P(o_\tau | s_\tau)Q(s_\tau | \pi)$ will be more similar to $Q(s_\tau | \pi)$ than it is to $P(s_\tau | o_\tau)$
\end{itemize}
Therefore, the joint distribution $P(o_\tau | s_\tau)Q(s_\tau | \pi)$ will tend to be large when the difference within the expectation is negative (c.f., Figure \ref{fig:epistemic_value_computation}). Indeed, as the joint is similar to the approximate posterior, it means that the joint distribution is large when the approximate posterior is large and the true posterior is smaller. This implies that the logarithm of true posterior will be very negative, while the logarithm of the approximate posterior will be less negative, i.e., closer to zero. Then, the logarithm of the approximate posterior will be substrated from the logarithm of the true posterior, i.e., a small positive number will be added to a very negative number. Therefore, the result will be negative.

The right-most graph of Figure \ref{fig:epistemic_value_experiments} shows that the epistemic value increases as the prior $P(s_\tau)$ and therefore the true posterior $P(o_\tau | s_\tau)$ becomes more similar to the approximate posterior $Q(s_\tau | \pi)$. This behaviour should not be observed, as the epistemic value is bigger when the true posterior $P(s_\tau | o_\tau)$ and the approximate posterior $Q(s_\tau | \pi)$ are more similar. Thus, maximising epistemic value will not promote exploration. In fact, it will recommend a strong focus on a single action as was observed in Figure \ref{fig:actions_from_critic}.

\end{document}